\pdfoutput=1

\documentclass[11pt]{article}

\usepackage[final]{acl}

\usepackage{times}
\usepackage{latexsym}

\usepackage{hyperref}
\usepackage{url}
\usepackage{multirow}
\usepackage{graphicx}
\usepackage{comment}
\usepackage{amssymb}
\usepackage{amsmath}
\usepackage{url}

\usepackage[T1]{fontenc}

\usepackage[utf8]{inputenc}

\usepackage{microtype}

\usepackage{inconsolata}

\usepackage{graphicx}

\title{Can We Trust the Performance Evaluation of Uncertainty Estimation Methods in Text Summarization?}

\author{
Jianfeng He{$^\dag$}\thanks{This work was done before joining Amazon.}, Runing Yang{$^\dag$}, Linlin Yu{$^\ddag$}, Changbin Li{$^\ddag$}, \\
\textbf{Ruoxi Jia{$^\dag$}, Feng Chen{$^\ddag$}, Ming Jin{$^\dag$}, Chang-Tien Lu{$^\dag$}}\thanks{Ming Jin and Chang-Tien Lu are co-corresponding authors.}~
\\ 
 {$^\dag$}Department of Computer Science, Virginia Tech, Falls Church, VA, USA\\
{$^\ddag$}Department of Computer Science, The University of Texas at Dallas, Richardson, TX, USA\\
 {$^\dag$}\{jianfenghe, yruning, ruoxijia, jinming, ctlu\}@vt.edu, \\{$^\ddag$}\{Linlin.Yu, Changbin.Li, Feng.Chen\}@utdallas.edu
 }

\begin{document}
\maketitle

\begin{abstract}

Text summarization, a key natural language generation (NLG) task, is vital in various domains. However, the high cost of inaccurate summaries in risk-critical applications, particularly those involving human-in-the-loop decision-making, raises concerns about the reliability of uncertainty estimation on text summarization (UE-TS) evaluation methods. This concern stems from the dependency of uncertainty model metrics on diverse and potentially conflicting NLG metrics. To address this issue, we introduce a comprehensive UE-TS benchmark incorporating 31 NLG metrics across four dimensions. The benchmark evaluates the uncertainty estimation capabilities of two large language models and one pre-trained language model on three datasets, with human-annotation analysis incorporated where applicable. We also assess the performance of 14 common uncertainty estimation methods within this benchmark. Our findings emphasize the importance of considering multiple uncorrelated NLG metrics and diverse uncertainty estimation methods to ensure reliable and efficient evaluation of UE-TS techniques. Our code and data are available \href{https://github.com/he159ok/Benchmark-of-Uncertainty-Estimation-Methods-in-Text-Summarization}{here}.

\end{abstract}

\section{Introduction}

Text summarization~\cite{tas2007survey} is a representative NLG task that generates summaries for given texts. This study researches abstractive summarization~\cite{nallapati2016abstractive}, which is more flexible than extractive summarization~\cite{gupta2010survey}. In many scenarios (e.g., finance and health), there are serious consequences if relying on false predicted summaries. For instance, an inaccurate financial report summary could lead to incorrect financial decisions, resulting in financial losses~\cite{gomez2003usefulness}. Consequently, the task of UE-TS, which measures the likelihood that a generated summary is low-quality, has garnered significant interest~\cite{gidiotis2021should,fadeeva2023lm,he2024semi}. However, an overlooked question persists regarding the \textit{validity} (the ability to accurately measure what it intends) and \textit{robustness} (consistency across various scenarios and datasets) of the UE-TS evaluation framework.

This concern arises due to two primary factors. First, common UE-TS evaluation metrics, such as forced-choice evaluation~\cite{he2024semi} and Prediction Rejection Ratio (PRR)~\cite{malinin-etal-2017-incorporating,fadeeva2023lm}, rely on the alignment between two types of sample rankings: those based on uncertainty scores from an estimation method (e.g., entropy of generation semantics~\cite{kuhn2023semantic}) and those derived from a specific NLG metric (e.g., ROUGE~\cite{lin2004rouge}). While higher alignment suggests better uncertainty estimation, the reliance on a single NLG metric may not fully capture the nuances of summary quality.

Second, the inherent label diversity in NLG tasks, particularly text summarization, necessitates the use of various NLG metrics. For instance, in text summarization, summaries like ``The cat chased the mouse'' and ``The mouse was pursued by the kitty'' convey identical messages using different expressions. Due to the multiple valid responses possible in such tasks~\cite{vijayakumar2016diverse}, it is unreasonable to expect NLG predictions to perfectly match the labels, unlike in NLU tasks~\cite{hu2021uncertainty,he2020towards}. Consequently, each NLG metric emphasizes different aspects of the diverse labels. For instance, ROUGE~\cite{lin2004rouge} and BLEU~\cite{papineni2002bleu} assess the similarity between the prediction and label based on n-gram overlap,  while SummaC~\cite{laban2022summac} examines the level of hallucination between the generation and reference text.

Collectively, previous evaluations of UE-TS models~\cite{gidiotis2021should,fadeeva2023lm,he2024semi}  typically relied on one or two uncertainty estimation metrics, each dependent on a single type of NLG metric. This limited evaluation raises a critical question:

\begin{quote}
\textit{How reliable and valid are current methods for evaluating uncertainty estimation in text summarization?
}
\end{quote}

Answering this question is essential for developing reliable UE-TS methods that can mitigate the risks associated with inaccurate summaries in critical applications, such as financial decision-making or healthcare. Moreover, our research aims to explore the broader relationship between uncertainty estimation and NLG metric scores, the impact of NLG metric diversity on experimental design and analysis, and the effective selection of diverse uncertainty estimation methods. This comprehensive exploration will inform and enhance future UE-TS research. To address these questions, we introduce a UE-TS benchmark incorporating a diverse set of text generation metrics. Our contributions are summarized below.

\begin{itemize}
    \item  To the best of our knowledge, we are the first to highlight a critical issue regarding the reliability of performance evaluation for UE-TS methods. We propose addressing this concern by evaluating UE-TS methods using a diverse set of NLG metrics. These NLG metrics span four dimensions crucial for NLG evaluation, including coherence, consistency, fluency, and relevance, following~\citet{zhong2022towards}.

    \item We present a UE-TS benchmark for evaluating uncertainty estimation through various NLG metric perspectives, marking the first endeavor to our knowledge. This benchmark assesses two Large Language Models (LLMs) and one Pre-trained Language Model (PLM) across three datasets focusing on UE-TS. Within this benchmark, we incorporate 31 NLG metrics and fourteen uncertainty estimation methods. To facilitate future research, the intermediate results encompass each sample's NLG metric and uncertainty scores. An intermediate result for an original text includes the generated summaries obtained from respective summarization models, the uncertainty scores for the sample via different uncertainty estimation methods, and the NLG metric scores via different NLG metrics for the sample. The statistics reported in our tables and figures can be reproduced with these intermediate results.

    \item We also conduct human experimental analysis via a publicly available human-annotation dataset. It finds that the NLG metrics do not always correlate closely with the human annotation. However, the NLG metrics can still be used to find the best uncertainty estimation method, which saves the cost and shows some similar uncertainty estimation method rank as that ranked via human annotation.

    \item We have uncovered intriguing findings outlined in Sec.~\ref{sec:summary_findings}. Because text summarization is a representative NLG task, our findings in text summarization tasks could also motivate design and analysis on uncertainty estimation in other NLG tasks.

\end{itemize}

\section{Related Work}
\textbf{Text summarization.} There are two types of text summarization: extractive text summarization~\cite{wong2008extractive,liu2019fine}, which extracts the original sentences from the text for summarization, and abstractive text summarization~\cite{liu2019text,nallapati2016abstractive}, which directly generates summaries from the text. Due to the flexibility of abstractive text summarization, we focus on abstractive text summarization. Recent models for abstractive text summarization are typically divided into two categories: PLMs (e.g., BART~\cite{lewis2019bart}) and LLMs (e.g., Llama 2~\cite{touvron2023llama}). To conduct comprehensive research on our question, we test abstractive summarization models from both PLM and LLMs.

\begin{figure*}[!htbp]
\centering
\includegraphics[width=0.85\textwidth]{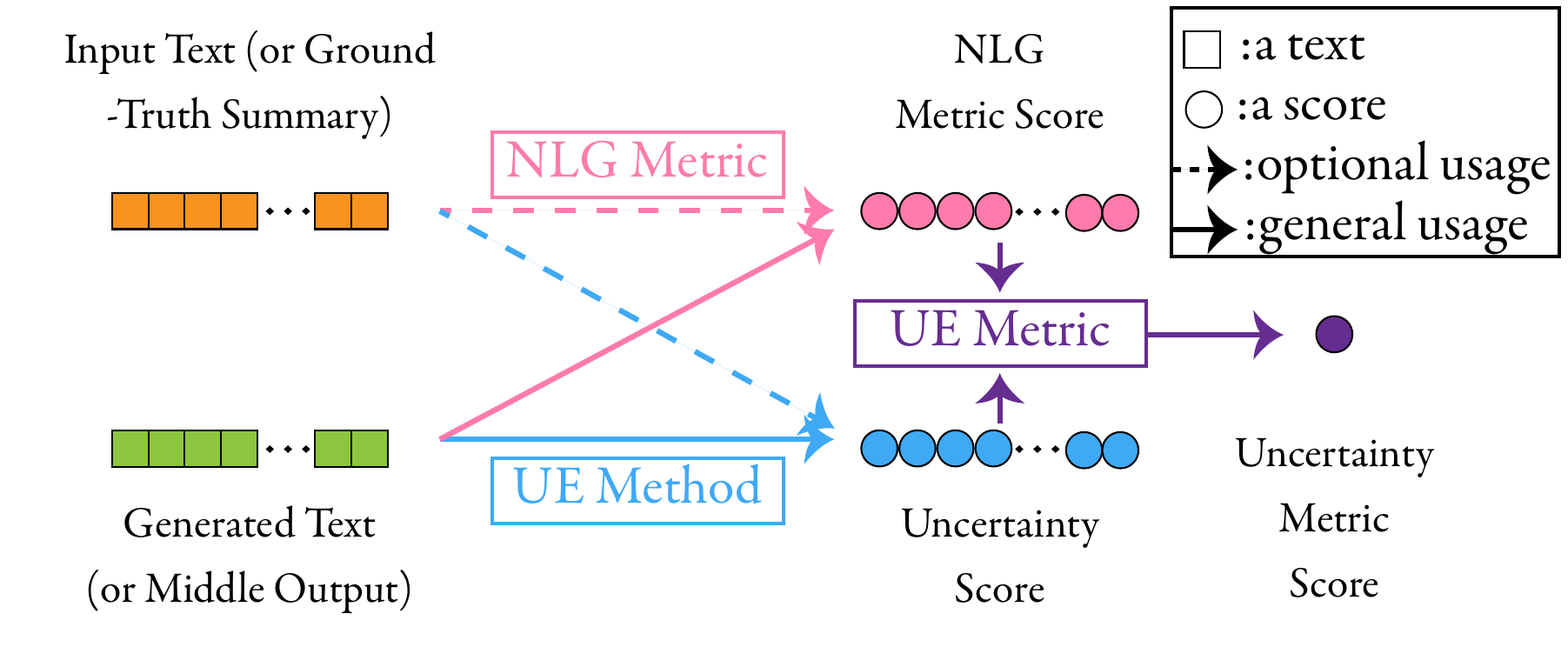}
\caption{Diagram of the relationship between the Uncertainty Estimation (UE) metric, NLG metrics, and UE methods in the evaluation process. Specifically, the evaluation process for UE-TS methods involves using the generated texts (or intermediate outputs, such as token probabilities) and the optional input text (or ground-truth summary) to obtain NLG metric scores and uncertainty scores for all test samples, through an NLG metric and a UE method, respectively. Finally, the NLG metric scores and uncertainty scores for all testing samples are both inputted into a UE metric to obtain an uncertainty metric score of the UE method.}
\label{fig:benchmark_frame}
\end{figure*}

\noindent\textbf{Uncertainty estimation on text summarization.}
The UE-TS methods primarily fall into four categories~\cite{fadeeva2023lm}: information-based, density-based, ensemble-based, and prompt-based methods. Information-based methods use middle output (e.g., token probability) to obtain uncertainty scores~\cite{beigi2024internalinspector}. For example, \citet{tsvigun2023active} calculate uncertainty scores based on token logits in two ways: mean and Monte Carlo Dropout~\cite{wang2019improving}. Similarly,~\citet{simpson2020interactive,zhang2022momentum} also calculate uncertainty based on token logits.
Density-based methods leverage latent representations of instances, which are further used to construct a probability density. For example,~\citet{ren2022out} detect low-quality generations by a density learned on the given sample embeddings.
Ensemble-based methods use ensembles to approximate Bayesian Neural Networks (BNN)~\cite{mukhoti2023deep} or use variance of ensembled generation to obtain uncertainty scores. For example, dropout~\cite{gal2016dropout} is used in~\citet{gidiotis2021should} to approximate BNN. Also,~\citet{chuang2024spec} obtain the uncertainty score by the variance of ensembling predictions. As for prompt-based methods, they refer to the methods that ask the generation model via a prompt to obtain the uncertainty score~\cite{kadavath2022language}.
Some methods, like the SiCF score~\cite{he2024semi}, integrate aspects of multiple uncertainty estimation methods, such as information and ensemble methods. To ensure comprehensive research on UE-TS methods, our benchmark includes representative methods from each category, following~\citet{fadeeva2023lm}.

\noindent\textbf{Performance evaluation of uncertainty estimation in text summarization.} 
As for the uncertainty evaluation methods, ~\citet{gidiotis2021should,gidiotis2021bayesian} obtain the uncertainty score by measuring the variance among ensembled generations based on an NLG metric, BLEU~\cite{papineni2002bleu}. BLEU solely assesses uncertainty estimation method performance based on n-gram similarity. \citet{kolagar2024aligning} compare the annotated expression of uncertainty between human annotations and annotations made by LLM based on an NLG metric, semantic similarity via SentenceBERT~\cite{reimers2019sentence}. ~\citet{lei2024polarity} evaluates UE-TS models based on the polarity score between prediction and ground truth summaries. ~\citet{zhao2022calibrating,zablotskaia2023uncertainty} evaluate uncertainty estimation methods solely using one NLG metric, ROUGE~\cite{lin2004rouge}. Only two NLG metrics, ROUGE and position accuracy, have been used separately for evaluation in ~\citet{zhang2022momentum}. Additionally, ROUGE and BERTScore~\cite{zhang2019bertscore}, measuring semantic similarity, are both considered in ~\citet{he2024semi,fadeeva2023lm}.

Thus, we observed that current evaluations of uncertainty estimation in text summarization rely on a limited set of NLG metric scores, which may lead to inconsistent performance rankings across other NLG metrics. Hence, our research aims to comprehensively explore the relationship between different uncertainty methods and various NLG metrics.

\section{Metrics \& Methods in Benchmark} 

\begin{figure*}[!htbp]
\centering
\includegraphics[width=0.85\textwidth]{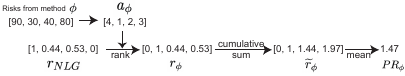}
\caption{Diagram of the $PR_{\phi}$ calculation example with testing sample size $N=4$. In this example, we have min-max normalized $\hat{s}_{NLG}=[0, 0.56, 0.47, 1]$, which is not drawn in the figure. Once we have obtained the sample rank $a_{\phi}$ based on a score list from method $\phi$. We rerank $r_{NLG}$ via $a_{\phi}$ to get $r_{\phi}$. Then, we use Eq.~\ref{eq:cum_risk} to cumulatively sum the elements and obtain $\widetilde{r}_{\phi}$. Finally, the $PR_{\phi}$ is the mean of $\widetilde{r}_{\phi}$. }
\label{fig:pr_example}
\end{figure*}

\subsection{Our Benchmark: A Global Overview}

To answer ``How does the choice of NLG metric affect the evaluation of uncertainty estimation methods in text summarization?'', our benchmark aims to explore the uncertainty estimation metric score across different NLG metrics for a given uncertainty estimation method. The relationship of these three items in the model evaluation process is shown in Figure~\ref{fig:benchmark_frame}. Below, we formalize the definition of the three scores we used in our work.
\begin{itemize}
    \item \textbf{NLG metric score}: reflects the quality of a generation at the sample level via an NLG metric (e.g., ROUGE).

    \item \textbf{Uncertainty score}: reflects the likelihood that a model generation is low-quality at the sample level. The likelihood comes from an uncertainty estimation method (e.g., BNN).

    \item \textbf{Uncertainty estimation metric score}: reflects the performance of an uncertainty estimation method at the method level.
\end{itemize}

Regarding the \textit{NLG metrics}, we evaluate 31 different ones across four NLG evaluation dimensions, as proposed by~\citet{zhong2022towards}. For the \textit{uncertainty estimation methods} to obtain uncertainty scores, we examine fourteen common approaches outlined in~\citet{fadeeva2023lm}. For the \textit{uncertainty estimation metric}, we adopt a widely used one, the Prediction Rejection Ratio (PRR)~\cite{malinin2017incorporating,fadeeva2023lm}.  Further details about these three components are provided below.

\subsection{Uncertainty Estimation Metric: PRR}
\label{sec:uem_prr}
Although uncertainty estimation metrics, such as force-truth evaluation~\cite{he2024semi} and PRR~\cite{malinin2017incorporating,fadeeva2023lm}, rely on NLG metrics, we opt for PRR in our benchmark due to its efficiency. Unlike force-truth evaluation, which necessitates repeating NLG metric calculations ten times, PRR offers a more streamlined approach. It is formatted as follows:
\begin{equation}
\label{eq:prr}
        PRR = \frac{PR_{uncertainty}-\frac{1}{\alpha}\sum_{i=1}^{\alpha}PR_{random}}{PR_{oracle}-\frac{1}{\alpha}\sum_{i=1}^{\alpha}PR_{random}}
\end{equation}
where $PR_{\phi}$ is a scalar representing the cumulative risk, calculated based on a predicted sample rank $a_{\phi}$ from a method $\phi\in \{uncertainty, random, oracle\}$~\footnote{The usage of $i$ and $\alpha$ for $PR_{random}$ will be introduced in Sec.~\ref{sec:three_pr_cal}.} and a list of NLG metric scores $s_{NLG}$. Specifically, the $s_{NLG}\in\mathbb{R}^{N}$ for $N$ testing samples, is first min-max normalized to $\hat{s}_{NLG}\in \mathbb{R}^{N}$. Then, the risk for $N$ testing samples is as follows,
\begin{equation}
\label{eq:ini_risk}
        r_{NLG} = 1 - \hat{s}_{NLG}
\end{equation}
here, we assume that our chosen NLG metric score is positively correlated with generation performance, and common NLG metrics (e.g., ROUGE) meet this assumption. Thus, the NLG metric score (an element in $\hat{s}_{NLG}$) is negatively correlated with the risk (an element in $r_{NLG}$) of inaccurate generation. A higher risk indicates a higher chance of inaccurate generation. We then obtain $r_{\phi}$ by ranking $r_{NLG}$ based on sample rank in $a_{\phi}$. This process is shown in Fig.~\ref{fig:pr_example}.
Further, we obtain a cumulative risk vector $\widetilde{r}_{\phi}=[\widetilde{r}_{\phi, 1}, \widetilde{r}_{\phi, 2}, ..., \widetilde{r}_{\phi, N}]$. 
Its $k$-th element is calculated via cumulative summing the first $k$ elements,
\begin{equation}
\label{eq:cum_risk}
\widetilde{r}_{\phi, k}=\sum_{j=1}^{k} r_{\phi, j}
\end{equation}
finally, $PR_{\phi}$ is the mean of cumulative risk $\widetilde{r}_{\phi}$.

Because the risk is defined based on the normalized NLG metric score list $\hat{s}_{NLG}$ in Eq.~\ref{eq:ini_risk}, the $PR{\phi}$ is smaller if the predicted sample rank $a_{\phi}$ is more aligned with the sample rank based on the NLG metric score.  
A simple intuition is that if the predicted sample rank $a_{\phi}$ is identical to the sample rank based on $r_{NLG}$, $PR_{\phi}$ is minimized. 

\subsection{Details about Three PR Calculations}
\label{sec:three_pr_cal}
Then, we introduce $PR_{oracle}$, $PR_{random}$, and $PR_{uncertainty}$ as follows.

For $PR_{oracle}$, the predicted sample rank $a_{oracle}$ is obtained via an oracle score vector $s_{oracle}=-1\times s_{NLG}$. The $PR_{oracle}$ is the lowest cumulative risk we can obtain because $a_{oracle}$ is aligned with the sample rank via the NLG metric score.

For $PR_{random}$, we generate a random permutation of $N$ numbers each time. $\frac{1}{\alpha}\sum_{i=1}^{\alpha}PR_{random}$ represents the average of $\alpha$ different $PR_{random}$ values, each calculated using a random permutation. In our experiments, we set $\alpha$ to 1000.

For $PR_{uncertainty}$, the predicted sample rank is obtained from the uncertainty score, which is calculated via an uncertainty estimation method. We list our uncertainty estimation methods in Sec.~\ref{sec:ue_method}.

As a result, PRR in Eq.~\ref{eq:prr} calculates the relative risk between uncertainty scores from an uncertainty method and NLG metric scores from an NLG metric. The relative risk is normalized to random risk expectation $\frac{1}{\alpha}\sum_{i=1}^{\alpha}PR_{random}$. A higher PRR means a more accurate uncertain estimation model. This is because a smaller $PR_{\phi}$ leads to better alignment for method $\phi$ and the denominator in Eq.~\ref{eq:prr} is negative. Thus, a larger PRR means smaller $PR_{uncertainty}$ and higher generation quality.

\begin{table*}[]
\small
\centering
\begin{tabular}{l|l|l}
\hline
\multirow{10}{*}{White-box} & \multirow{3}{*}{Information-based methods}   & Maximum Sequence Probability (\textbf{MSP})                    \\
                            &                                              & Mean Token Entropy (\textbf{MTE})                              \\
                            &                                              & Monte Carlo Sequence Entropy (\textbf{MCSE})                   \\ \cline{2-3} 
                            & \multirow{2}{*}{Density-based methods}       & Mahalanobis Distance (\textbf{MD})                             \\
                            &                                              & Robust Density Estimation (\textbf{RDE})                       \\ \cline{2-3} 
                            & \multirow{4}{*}{Ensemble-based methods}      & Token-level Total Uncertainty (\textbf{T-TU})                  \\
                            &                                              & Token-level Reverse Mutual Information (\textbf{T-RMI})        \\
                            &                                              & Sequence-level Total Uncertainty (\textbf{S-TU})               \\
                            &                                              & Sequence-level Reverse Mutual Information RMI (\textbf{S-RMI}) \\ \cline{2-3} 
                            & Prompt-based                                 & \textbf{P(True)}                                      \\ \hline
\multirow{4}{*}{Black-box}  & \multirow{4}{*}{Mixture types} & Number of Semantic Sets (\textbf{NumSets})                     \\
                            &                                              & Eccentricity (\textbf{ECC})                                    \\
                            &                                              & Lexical Similarity (\textbf{LexSim})                           \\
                            &                                              & Sum of Eigenvalues of the Graph Laplacian (\textbf{EigV})      
                            \\ \hline
\end{tabular}
\caption{A summary of the fourteen uncertainty methods that are used in our benchmark.}
\label{tab:sum_ue_methods}
\end{table*}

\subsection{Uncertainty Estimation Methods}
\label{sec:unc_est_met}
\label{sec:ue_method}
The uncertainty estimation methods utilized in our benchmark adhere to the framework established by~\citet{fadeeva2023lm}, as the framework provides a publicly available uncertainty estimation tool, enhancing our results' reproducibility.

The uncertainty estimation methods can be divided into two kinds: white-box methods and black-box methods. White-box methods refer to uncertainty estimation methods using the intermediate output, model structure, or model parameters of the text summarization model (e.g., BART). Black-box methods refer to uncertainty estimation methods that require only the final text output from a text summarization model (e.g., GPT-3.5~\cite{OPENAI35}) for the uncertainty estimation. 
Below, we introduce our uncertainty estimation methods one by one in Sec.~\ref{app:ue_methods}., with a summary provided in Table~\ref{tab:sum_ue_methods}.

\subsection{NLG Metrics}
\label{sec:nlg_metrics}
We chose thirty-one commonly used NLG metrics. To better understand the relationship between these NLG metrics, we categorize them into four dimensions~\cite{zhong2022towards}. We first introduce these four dimensions, then present all thirty-one NLG metrics with a summary in Table~\ref{tab:sum_nlg_metrics}. In this table, except for the LLM-based metrics, most traditional NLG metrics are tested in~\citet{fadeeva2023lm}.

\subsubsection{Dimensions for NLG Metrics} 
\label{sec:dim_concepts}
We choose four commonly used dimensions for NLG metrics, proposed in~\citet{zhong2022towards} and listed below.

\noindent\textbf{Relevance} refers to whether the generated text contains only the important information from the input text.

\noindent\textbf{Consistency} is the factual alignment between the generated text and the input text.

\noindent\textbf{Coherence} refers to whether all the sentences in the given generated text form a coherent body.

\noindent\textbf{Fluency} represents the quality of individual sentences in the generated text.

Additionally, we also consider the \textbf{overall} dimension that rates the generated texts based on all the above four dimensions.

The detailed explanation for each NLG metric in Table~\ref{tab:sum_nlg_metrics} is in Sec.~\ref{app:nlg_metrics}.

\begin{table}[!htb]
\small
\centering
\begin{tabular}{l|l}
\hline
                              & ROUGE-L                             \\
                              & Spearman    \\
                              & Kendall-Tau \\
                              & UniEval (Relevance)                 \\
                              & wo-GPT-3.5 (Relevance)              \\
                              & wi-gt-GPT-3.5 (Relevance)              \\
                              & wi-in-GPT-3.5 (Relevance)              \\
\multirow{-8}{*}{Relevance}   & wi-ingt-GPT-3.5 (Relevance)                \\ \hline
                              & BARTSCORE                           \\
                              & SummaC                              \\
                              & CTC                                 \\
                              & UniEval (Consistency)               \\
                              & wo-GPT-3.5 (Consistency)            \\
                              & wi-gt-GPT-3.5 (Consistency)            \\
                              & wi-in-GPT-3.5 (Consistency)            \\                        
\multirow{-8}{*}{Consistency} & wi-ingt-GPT-3.5 (Consistency)              \\ \hline
                              & UniEval (Coherence)                 \\
                              & wo-GPT-3.5 (Coherence)              \\
                              & wi-gt-GPT-3.5 (Coherence)              \\
                              & wi-in-GPT-3.5 (Coherence)              \\                           
\multirow{-5}{*}{Coherence}   & wi-ingt-GPT-3.5 (Coherence)                \\ \hline
                              & UniEval (Fluency)                   \\
                              & wo-GPT-3.5 (Fluency)                \\
                              & wi-gt-GPT-3.5 (Fluency)                \\
                              & wi-in-GPT-3.5 (Fluency)                \\                              
\multirow{-5}{*}{Fluency}     & wi-ingt-GPT-3.5 (Fluency)                  \\ \hline
                              & UniEval (Overall)                   \\
                              & wo-GPT-3.5 (Overall)                \\
                              & wi-gt-GPT-3.5 (Overall)                \\
                              & wi-ingt-GPT-3.5 (Overall)                \\                              
\multirow{-5}{*}{Overall}     & wi-in-GPT-3.5(Overall)                  \\ \hline
\end{tabular}
\caption{A summary of the thirty-one NLG metrics that are used in our benchmark.}
\label{tab:sum_nlg_metrics}
\end{table}

\section{Experiments}
\subsection{Experimental Settings} 

\noindent\textbf{Dataset.} We employ two widely used text summarization datasets in our experiments. Firstly, we utilize the AESLC dataset~\cite{zhang2019email}, comprising 1,906 testing texts from the email domain. Secondly, we incorporate the XSUM dataset~\cite{narayan2018don}, featuring articles collected from the British Broadcasting Corporation (BBC) and encompassing 11,334 testing samples. 

Besides these two, we are also interested in human-related experiments. To meet our experimental requirements, we used the TofuEval~\cite{tang2024tofueval} dataset. This dataset contains 100 dialogue-summary pairs with human annotations in terms of quality, evaluated from seven dimensions, which will be introduced in Sec.~\ref{sec:human_seven_dimension}.

\noindent\textbf{Related methods and metrics.} We provide detailed descriptions of the uncertainty estimation methods in Sec.\ref{sec:unc_est_met}. Additionally, an introduction to the NLG metrics is presented in Sec.\ref{sec:nlg_metrics}. By combining the uncertainty scores obtained from the uncertainty estimation methods with the NLG metric scores, we calculate the uncertainty metric score using the PRR method introduced in Sec.~\ref{sec:uem_prr}.

\begin{table*}[!htb]
\centering
\small
\begin{tabular}{l|rrrrrr}
\hline
\multicolumn{1}{l|}{\textbf{NLG Metrics}} & \multicolumn{1}{c}{\textbf{MSP}} & \multicolumn{1}{c}{\textbf{MTE}} & \multicolumn{1}{c}{\textbf{MCSE}} & \multicolumn{1}{c}{\textbf{MD}} & \multicolumn{1}{c}{\textbf{RDE}} & \multicolumn{1}{c}{\textbf{P(True)}} \\
\hline
ROUGE-L                                  & 0.2107                           & 0.1668                           & 0.2082                            & \bf0.2650                          & 0.1608                           & -0.0233                              \\
BARTSCORE                                & 0.0372                           & 0.2015                           & 0.0418                            & \bf0.2451                          & 0.1573                           & 0.0615                               \\
SummaC                                   & -0.1301                          & -0.0440                          & -0.1189                           & 0.0739                          & -0.0182                          & \bf0.0998                               \\
CTC                                      & \bf0.0736                           & -0.1515                          & 0.0685                            & 0.0457                          & 0.0347                           & -0.1074                              \\
Spearman                                 & 0.0656                           & \bf0.1430                           & 0.0640                            & 0.1429                          & 0.0929                           & -0.0080                              \\
Kendall-Tau                              & 0.0649                           & \bf0.1412                           & 0.0630                            & 0.1404                          & 0.0904                           & -0.0084                              \\
UniEval (Relevance)                      & -0.0572                          & 0.0769                           & -0.0309                           & \bf0.1995                          & -0.0334                          & -0.2365                              \\
UniEval (Consistency)                    & \bf0.1408                           & 0.1007                           & 0.1143                            & 0.0224                          & 0.1392                           & -0.3789                              \\
UniEval (Coherence)                      & 0.1804                           & 0.1802                           & 0.1932                            & 0.2163                          & \bf0.2332                           & -0.7554                              \\
UniEval (Fluency)                        & -0.0524                          & -0.3809                          & \bf0.0079                            & 0.0568                          & -0.1954                          & -0.2343                              \\
UniEval (Overall)                        & 0.0300                           & 0.0000                           & \bf0.0563                            & 0.1702                          & 0.0148                           & -0.4410                              \\
wo-GPT-3.5 (Relevance)                   & -0.0863                          & \bf0.2315                           & -0.0451                           & 0.1784                          & -0.0533                          & 0.1029                               \\
wo-GPT-3.5 (Consistency)                 & -0.0280                          & 0.1825                           & 0.0025                            & \bf0.2586                          & 0.0157                           & 0.0849                               \\
wo-GP-T3.5 (Coherence)                   & -0.0796                          & 0.0771                           & -0.0479                           & \bf0.1672                          & -0.0712                          & 0.0969                               \\
wo-GPT-3.5 (Fluency)                     & -0.0148                          & 0.1498                           & 0.0023                            & \bf0.1992                          & -0.0347                          & 0.0615                               \\
wo-GPT-3.5 (Overall)                     & -0.0634                          & 0.0976                           & -0.0794                           & \bf0.1807                          & 0.0187                           & 0.0399                               \\
wi-gt-GPT-3.5 (Relevance)                & -0.2205                          & \bf0.0841                           & -0.2692                           & 0.0835                          & 0.0833                           & -0.1177                              \\
wi-gt-GPT-3.5 (Consistency)              & -0.2533                          & -0.0078                          & -0.2726                           & \bf -0.0076                         & -0.0268                          & -0.0773                              \\
wi-gt-GPT-3.5 (Coherence)                & -0.2991                          & -0.0493                          & -0.2979                           & \bf 0.0333                          & -0.0597                          & -0.1096                              \\
wi-gt-GPT-3.5 (Fluency)                  & -0.2344                          & -0.0392                          & -0.2084                           & \bf0.1485                          & 0.0882                           & -0.3175                              \\
wi-gt-GPT-3.5 (Overall)                  & -0.1829                          & \bf0.0045                           & -0.2133                           & -0.0836                         & -0.0842                          & -0.1202                              \\
wi-in-GPT-3.5 (Relevance)                & 0.0185                           & \bf0.0798                           & 0.0735                            & 0.0608                          & -0.0820                          & -0.0239                              \\
wi-in-GPT-3.5 (Consistency)              & 0.1015                           & 0.1054                           & 0.1041                            & \bf0.1972                          & 0.0545                           & -0.1937                              \\
wi-in-GPT-3.5 (Coherence)                & -0.0170                          & \bf0.0653                           & -0.0206                           & 0.0283                          & -0.1453                          & -0.2302                              \\
wi-in-GPT-3.5 (Fluency)                  & 0.0808                           & -0.1020                          & 0.0955                            & \bf0.1685                          & 0.0308                           & -0.4904                              \\
wi-in-GPT-3.5 (Overall)                  & -0.1816                          & \bf0.1285                           & -0.1818                           & 0.0457                          & -0.0564                          & -0.0854                              \\
wi-ingt-GPT-3.5 (Relevance)              & -0.1114                          & \bf0.1486                           & -0.1015                           & 0.1182                          & 0.0500                           & -0.0379                              \\
wi-ingt-GPT-3.5 (Consistency)            & -0.1391                          & \bf0.1558                           & -0.1462                           & 0.0934                          & 0.0114                           & -0.0660                              \\
wi-ingt-GPT-3.5 (Coherence)              & -0.1079                          & 0.0909                           & -0.0959                           & \bf0.1531                          & 0.0058                           & -0.1189                              \\
wi-ingt-GPT-3.5 (Fluency)                & -0.0283                          & 0.0799                           & 0.0065                            & \bf0.1402                          & -0.0077                          & -0.2216                              \\
wi-ingt-GPT-3.5 (Overall)                & -0.1149                          & \bf0.0485                           & -0.1011                           & -0.0094                         & -0.1268                          & -0.0594                              \\
\hline
Col Mean                                 & -0.0451                          & 0.0634                           & -0.0364                           & \bf 0.1204                          & 0.0092                           & -0.1263                             
                                              
\\
\hline
\end{tabular}
\caption{Main results of the relationship between the uncertainty estimation methods and NLG metrics on AESLC dataset using generation from Llama 2.}
\label{tab:main_aes_llama}
\end{table*}

\noindent\textbf{Implementation details.} To limit diverse generation,  the temperature was set to 0, and a random seed of 42 was used. The experiments were conducted on a server equipped with a single A100 GPU. For the AESLC dataset, it took approximately 17 hours to run on GPT-3.5 generation and around 22 hours to run on Llama 2. For the XSUM dataset, the experiments took approximately six times longer to run compared to those on AESLC. The total cost is about 1000 USD for the GPT-3.5-based generation and GPT-3.5-based evaluation. For the density-based method, we randomly sampled 1000 training samples with a fixed random seed from the respective dataset to represent the training data distribution, following the settings outlined in~\citet{fadeeva2023lm}.

\subsection{Experimental Results about Uncertainty Estimation Methods}
\label{sec:ex_res_ue}

\textbf{Analysis on ensemble-based uncertainty estimation methods.}
In the case of ensemble-based uncertainty estimation methods, T-TU, S-TU, and S-RMI generally exhibit positive correlations with all other white-box uncertainty estimation methods, except for T-RMI. This trend is evident in Figures~\ref{fig:spear_ue_aes_bart} and~\ref{fig:spear_ue_xsum_bart}. Additionally, T-RMI demonstrates negative correlations with T-TU and S-TU, as well as a weakly positive correlation with S-RMI. Consequently, we propose that for future applications of uncertainty estimation methods, focusing on one of T-TU, S-TU, and S-RMI could be advantageous, with T-RMI serving as a supplementary baseline.

\noindent\textbf{Analysis on information-based uncertainty estimation methods.}
In the realm of information-based uncertainty estimation methods, MSP and MCSE typically exhibit strongly positive correlations with each other, as evidenced in Figures~\ref{fig:spear_ue_aes_bart},~\ref{fig:spear_ue_aes_llama},~\ref{fig:spear_ue_xsum_bart}, and~\ref{fig:spear_ue_xsum_llama}.

Conversely, MTE exhibits inconsistently strong positive correlations with MSP and MCSE. For instance, while Figure~\ref{fig:spear_ue_aes_llama} demonstrates a weakly positive correlation between them, Figure~\ref{fig:spear_ue_xsum_bart} depicts a strongly positive correlation. Therefore, we recommend that in future comparison of uncertainty estimation methods, when considering MSP and MCSE, opting for one of them and optionally utilizing MTE as a supplementary baseline could be beneficial.

\noindent\textbf{Analysis on density-based uncertainty estimation methods.}
In the domain of density-based uncertainty estimation methods, MD and RDE typically demonstrate strongly positive correlations with each other. This trend is observed in Figures~\ref{fig:spear_ue_aes_bart},~\ref{fig:spear_ue_xsum_bart}, and~\ref{fig:spear_ue_xsum_llama}, with the exception of a weakly positive correlation in Figure~\ref{fig:spear_ue_aes_llama}.

Hence, we propose that in future applications of uncertainty estimation methods, utilizing either MD or RDE alone would suffice rather than using both of them as the baselines.

\noindent\textbf{Analysis on prompt-based uncertainty estimation methods.}
In the case of prompt-based uncertainty estimation methods, P(True) typically exhibits a weak or negative correlation with other uncertainty estimation methods. This trend is evident from Figures~\ref{fig:spear_ue_aes_llama} and~\ref{fig:spear_ue_xsum_llama}.

\noindent\textbf{Analysis on black-box uncertainty estimation methods.}
In terms of black-box uncertainty estimation methods, ECC, LexSim, and EigV typically exhibit positive correlations with each other, with ECC and EigV showing particularly strong correlations. Conversely, NumSets demonstrates a weak or negative correlation with ECC, LexSim, and EigV. These trends are shown in Figures~\ref{fig:spear_ue_aes_gpt35} and~\ref{fig:spear_ue_xsum_gpt35}.

Hence, we recommend that in future applications of uncertainty estimation methods involving ECC, LexSim, and EigV, opting for one of them and employing NumSets as a complementary measure could be beneficial.

\noindent\textbf{Analysis on specific uncertainty estimation methods.}
Besides, we observed that each UE method consistently achieves positive UniEval (Consistency) scores across all eight methods listed in Table~\ref{tab:main_aes_bart}. This indicates that all of the aforementioned methods outperform random ranking in terms of the NLG metric, UniEval (Consistency).

\begin{table*}[!htb]
\centering
\small
\begin{tabular}{l|ccccccc|c}
\hline
\textbf{NLG Metrics}          & \multicolumn{1}{c}{\textbf{EI}} & \multicolumn{1}{c}{\textbf{MR}} & \multicolumn{1}{c}{\textbf{SOAF}} & \multicolumn{1}{c}{\textbf{RE}} & \multicolumn{1}{c}{\textbf{TME}} & \multicolumn{1}{c}{\textbf{CO}} & \multicolumn{1}{c}{\textbf{NMS}} & \multicolumn{1}{|c}{\textbf{Row Mean}}\\
\hline
ROUGE-L                       & -0.5357                         & 0.2070                          & -0.3931                           & -0.7112                         & -0.4778                          & 0.0084                          & 0.0713                           & -0.2616                               \\
SummaC                        & -0.1109                         & 0.0525                          & 0.0981                            & 0.0838                          & -0.2313                          & -0.5984                         & 0.1242                           & -0.0831                               \\
CTC                           & -0.4205                         & 0.2899                          & 0.2378                            & -0.3425                         & -0.4008                          & 0.0238                          & 0.1431                           & -0.0670                               \\
Spearman                      & -0.4254                         & 0.2311                          & -0.0991                           & -0.2111                         & -0.1136                          & -0.2723                         & 0.0096                           & -0.1258                               \\
Kendall-Tau                   & -0.4216                         & 0.2387                          & -0.1086                           & -0.2151                         & -0.1111                          & -0.2905                         & 0.0112                           & -0.1281                               \\
UniEval (Relevance)           & 0.0388                          & -0.1822                         & 0.3394                            & -0.2541                         & -0.1908                          & -0.1578                         & -0.1172                          & -0.0748                               \\
UniEval (Consistency)         & -0.6647                         & -0.1144                         & 0.3829                            & -0.1344                         & 0.2585                           & -0.0797                         & -0.1606                          & -0.0732                               \\
UniEval (Coherence)           & -0.5794                         & 0.2330                          & 0.6794                            & -0.4737                         & -0.2264                          & 0.1767                          & -0.1442                          & \bf-0.0478                               \\
UniEval (Fluency)             & -0.0665                         & 0.2300                          & -0.6612                           & -0.3210                         & -0.3194                          & 0.2168                          & -0.3231                          & -0.1778                               \\
UniEval (Overall)             & -0.5925                         & -0.0105                         & 0.5262                            & -0.3591                         & -0.0376                          & 0.0002                          & -0.2080                          & -0.0973                               \\
wi-ingt-GPT-3.5 (Relevance)   & -0.0991                         & 0.1672                          & -0.2758                           & -0.2624                         & -0.1556                          & -0.1358                         & -0.2762                          & -0.1482                               \\
wi-ingt-GPT-3.5 (Consistency) & -0.3081                         & 0.1279                          & 0.0396                            & -0.3236                         & -0.4197                          & -0.0768                         & -0.0286                          & -0.1413                               \\
wi-ingt-GPT-3.5 (Coherence)   & -0.2569                         & 0.0856                          & -0.0278                           & -0.3163                         & -0.4721                          & 0.0666                          & 0.1028                           & -0.1169                               \\
wi-ingt-GPT-3.5 (Fluency)     & -0.1679                         & 0.0791                          & 0.0177                            & -0.3292                         & -0.2071                          & 0.0501                          & 0.0548                           & -0.0718                               \\
wi-ingt-GPT-3.5 (Overall)     & -0.2945                         & 0.0100                          & 0.1369                            & -0.2878                         & -0.3577                          & 0.0233                          & 0.2251                           & -0.0778                              
                     
\\
\hline
\end{tabular}
\caption{Main results of the relationship between the NLG metrics and human annotations on TofuEval dataset using generation from GPT-3.5.}
\label{tab:main_tofu_gpt35_nlg_hum}
\end{table*}

Each uncertainty estimation method exhibits dissatisfaction performance with negative ROUGE-L scores across all eight methods in Table~\ref{tab:main_xsum_bart}. This suggests that all the methods listed above perform worse than random ranking in terms of the NLG metric, ROUGE-L (Consistency).

Each uncertainty estimation method demonstrates strong performance, achieving positive scores for both UniEval (Consistency) and UniEval (Coherence) across all four methods presented in Table~\ref{tab:main_aes_gpt35}. This indicates that all aforementioned methods outperform random ranking in terms of both consistency and coherence according to UniEval assessment. We summarized our findings in Sec.~\ref{sec:human_seven_dimension}.

\subsection{Results about NLG Metrics}
The experimental results and experimental analysis from the view NLG metrics are in Sec.~\ref{sec:exp_res_ana}. Based on the information presented in these tables and figures in Sec.~\ref{sec:exp_res_ana}, we can address our key question: 
``How does the choice of NLG metric affect the evaluation of uncertainty estimation methods in text summarization?''
The answer is that \textbf{using different NLG metrics could lead to different ranks for uncertainty estimation methods}. Therefore, it is important to design uncertainty estimation metrics that are robust across various NLG metrics. Besides Sec.~\ref{sec:exp_res_ana}, more findings are summarized in Sec.~\ref{sec:summary_findings}.

\subsection{Experimental Results Involving Human Annotations}
The experimental results and experimental analysis involving human annotations are in Sec.~\ref{sec:human_seven_dimension}. Based on the human annotation from seven dimensions, we found that the rankings of certain uncertainty estimation methods differ between NLG metrics and human annotation. Also, current uncertainty estimation methods struggle to attain positive PRR scores across various perspectives of human annotations. This could be a future direction for the research. Besides Sec.~\ref{sec:human_seven_dimension}, more findings are summarized in Sec.~\ref{sec:summary_findings}.

\section{Conclusion}

Recognizing the dependency of uncertainty model metrics on diverse NLG metrics, we introduce a comprehensive UE-TS benchmark to investigate the impact of NLG metric choice on the evaluation of uncertainty estimation methods in text summarization. This benchmark encompasses \textbf{three} datasets, \textbf{fourteen} uncertainty estimation methods (both white-box and black-box), and \textbf{thirty-one} NLG metrics spanning \textbf{four} evaluation dimensions. Additionally, we assess one black-box LLM, one white-box LLM, and one PLM for text summarization.

Our findings not only highlight the current limitations in UE-TS model evaluation but also contribute to a deeper understanding of the complex relationship between NLG metrics, uncertainty estimation methods, and human annotation. This knowledge can guide future research, inform the development of more effective uncertainty estimation methods, and ultimately enhance evaluation protocols for UE-TS and other NLG tasks.

\section{Summary of Our Findings}
\label{sec:summary_findings}

Below, we summarize our findings, which are takeaway knowledge. The evidence to obtain these findings is detailed in Sec.~\ref{sec:ex_res_ue},~\ref{sec:exp_res_ana} and~\ref{sec:human_seven_dimension}.

From the view of \textbf{NLG metrics} (see Sec.~\ref{sec:exp_res_ana}), 

\begin{itemize}
    \item It is evident that evaluating uncertainty estimation models using different NLG metrics leads to variations in the performance ranking of these models.

    \item Some evaluations of uncertainty estimation models using different NLG metrics could result in different performance ranks. However, some evaluations of uncertainty estimation models using different NLG metrics may result in the same performance rankings.

    \item Generation models of the same type across different datasets usually result in similar correlations among various methods. However, this does not apply to the fluency dimension.

    \item When utilizing LLMs as a type of relevance NLG metric, the choice of target text source can greatly impact the final conclusion. Specifically, using ground-truth summaries versus using input text as the target text source can result in different performance rankings. 

    \item When using LLMs as an NLG metric, if both ground-truth summaries and input text are employed together as the target text source, ground-truth summaries typically impact metric results more than using only the input text in most cases.

    \item GPT-3.5 knows the concept of relevance, overall. But it might not know the concept of consistency, coherence. As for the concept of fluency, it is hard to tell whether the GPT-3.5 itself understand it or not. Therefore, using GPT-3.5 as an evaluation tool, it would be better to provide the concept definition in the prompt. 
\end{itemize}

Specific recommendations for \textbf{NLG Metrics}:
\begin{itemize}

    \item \textbf{Relevance:} Spearman and Kendall-Tau metrics show positive correlations, so either can be used. When using LLMs as generation models, UniEval (Relevance) and wo-GPT-3.5 (Relevance) typically show positive correlations with most other NLG metrics. Therefore, either of them could serve as a representative NLG metric in the relevance dimension (see Sec.~\ref{sec:app_nlg_relevance_dimension}).

    \item \textbf{Consistency:} CTC may be preferable to SummaC or UniEval due to positive correlations. For GPT-3.5-based metrics, it is recommended to use wi-in-GPT-3.5 (Consistency) alongside wo-gt-GPT-3.5, wi-gt-GPT-3.5 (Consistency), or wi-ingt-GPT-3.5 (Consistency) (see Sec.~\ref{sec:app_nlg_consistency_dimension}).

    
    \item \textbf{Coherence:} Determining the superior GPT-3.5-based metric in the coherence dimension is challenging. But, due to their strongly positive correlation, either wi-gt-GPT-3.5 (Coherence) or wi-in-GPT-3.5 (Coherence) is a suitable choice. In the coherence dimension, UniEval (Coherence) could complement either wi-gt-GPT-3.5 (Coherence) or wi-in-GPT-3.5 (Coherence) (see Sec.~\ref{sec:app_nlg_coherence_dimension}).


    \item \textbf{Fluency:} We recommend wi-gt-GPT-3.5 (Fluency) and wi-ingt-GPT-3.5 (Fluency) due to their correlation patterns. UniEval (Fluency) can supplement a GPT-3.5-based metric (see Sec.~\ref{sec:app_nlg_fluency_dimension}).




    \item The UniEval (Overall) can serve as an optional supplementary tool for GPT-3.5-based NLG metrics in the overall dimension (see Sec.~\ref{sec:app_nlg_overall_dimension}).




\end{itemize}

Key insights for \textbf{uncertainty estimation methods}  (see Sec.~\ref{sec:ex_res_ue}): For future applications of uncertainty estimation methods:
\begin{itemize}
    \item Prioritize one of T-TU, S-TU, and S-RMI, with T-RMI serving as a supplementary baseline.

    \item When considering MSP and MCSE, opting for one of them and utilizing MTE as an optional supplementary measure could be beneficial.

    \item Using either MD or RDE alone would suffice rather than using both of them as the baselines.

    \item When considering ECC, LexSim, and EigV, opting for one of them and employing NumSets as a complementary measure could be beneficial.
\end{itemize}

From the view of \textbf{human annotations} (see Sec.~\ref{sec:app_ue_hum_dim},~\ref{sec:app_nlg_hum_dim},~\ref{sec:app_ue_nlg_dim}), 
\begin{itemize}
    \item Rankings of certain uncertainty estimation methods differ between NLG metrics and human annotation.

    \item Utilizing NLG metrics as an oracle remains meaningful, as it saves the cost but leads to an uncertainty estimation method rank that is inconsistent with human annotations.

    \item Current uncertainty estimation methods struggle to attain positive PRR scores across various perspectives of human annotations. This could be a future direction for the research.

\end{itemize}

\section{Acknowledgment}
This research is supported in part by the National Science Foundation
(NSF) grants IIS-2107449, FAI-2147375, and IIS-1954376.

\section{Ethical Consideration}
This study explores the overlooked relationship between uncertainty metrics and NLG metrics. During our study, we answer the question ``Can We Trust the Performance Evaluation of Uncertainty Estimation Methods in Text Summarization?''.

Our research employs datasets that are publicly available, ensuring transparency and accessibility. The datasets integral to our work are utilized in adherence to their respective licenses~\cite{tang2024tofueval,fadeeva2023lm}.

We recommend that any future expansion of this research into areas involving personal or sensitive data should be approached with stringent ethical guidelines in place.

\section{Limitations}
Due to the resources (GPU and budget for calling GPT APIs), we only conduct one-time experiments. In contrast, the experiments on BART are set to use greedy generation for each sample rather than diverse generations for the robustness of the experiments.

\bibliography{main}

\clearpage
\appendix


\section{More Explanation of Uncertainty Estimation Methods}
\label{app:ue_methods}

Below we give a detailed explanation of the uncertainty estimation methods in Table~\ref{tab:sum_ue_methods}. 

\subsection{White-Box Methods}
The UE-TS methods primarily fall into four categories~\cite{fadeeva2023lm}: information-based, density-based, ensemble-based, and prompt-based methods.

\noindent\textbf{Information-based methods} utilize token probability to obtain uncertainty scores. Within this category, we employ the following methods:

\noindent (1) Maximum Sequence Probability (\textbf{MSP}): estimates uncertainty score as the log-probability of the generation in a greedy search way. It is calculated as the sum of log probabilities in each token.

\noindent (2) Mean Token Entropy (\textbf{MTE}):  estimates the uncertainty score as the mean entropy for all tokens in a generation.
    
\noindent (3) Monte Carlo Sequence Entropy (\textbf{MCSE}): calculates the generation entropy estimations using Monte-Carlo estimation. It is the ``predictive entropy'' in~\citet{kuhn2023semantic}.

\noindent\textbf{Density-based methods} utilize latent representations of instances to construct a probability density. Below, we list the related methods compared in our benchmark.

\noindent (4) Mahalanobis Distance (\textbf{MD}): calculates a Gaussian distribution for training samples. Then, the MD calculates the distance between the testing sample and the Gaussian distribution as the uncertainty score~\cite{lee2018simple}. 

\noindent (5) Robust Density Estimation (\textbf{RDE}): improves over MD by reducing the dimensionality via principal component analysis decomposition~\cite{yoo2022detection}.

\noindent\textbf{Ensemble-based methods} utilize ensembles to approximate Bayesian neural networks (BNN) or the variance of ensemble generations to obtain uncertainty scores. Below are the related methods we used: 

\noindent (6) Token-level Total Uncertainty (\textbf{T-TU}): calculates the entropy of the predictive posterior at token level as the uncertainty score~\cite{malinin2020uncertainty,he2024semi}.  

\noindent (7) Token-level Reverse Mutual Information (\textbf{T-RMI}): uses reverse-KL divergence counterpart to the mutual information at token level as the uncertainty score~\cite{malinin2020uncertainty}.  

\noindent (8) Sequence-level Total Uncertainty (\textbf{S-TU}): calculates the entropy of the predictive posterior at sequence level as the uncertainty score~\cite{malinin2020uncertainty,he2024semi}.  

\noindent (9) Sequence-level Reverse Mutual Information RMI (\textbf{S-RMI}): uses reverse-KL divergence counterpart to the mutual information at the sequence level as the uncertainty score~\cite{malinin2020uncertainty}.

\noindent \textbf{Prompt-based methods} refer to methods that prompt the generation model to obtain the uncertainty score~\cite{kadavath2022language}. We employ (10) \textbf{P(True)}~\cite{kadavath2022language}, which takes the uncertainty score as the probability of asking whether the proposed answer is true or false. This method assumes that the generation model possesses superior zero-shot prompt abilities.

\subsection{Black-Box Methods}
For scenarios where only the final textual output is available, we utilize the following black-box uncertainty estimation methods, which have demonstrated effectiveness in previous studies~\cite{fadeeva2023lm}.

\noindent (11) Number of Semantic Sets (\textbf{NumSets}): takes the number of diverse semantic interpretations for the generation as uncertainty score~\cite{lin2023generating}.

\noindent (12) Eccentricity (\textbf{ECC}): gets the uncertainty via calculating a distance between all eigenvectors that are informative embeddings of graph Laplacian~\cite{lin2023generating}.

\noindent (13) Lexical Similarity (\textbf{LexSim}): obtains the uncertainty scores via calculating mean similarity between all pairs of sampled generations~\cite{fomicheva2020unsupervised}. 

\noindent (14) Sum of Eigenvalues of the Graph Laplacian (\textbf{EigV}): extends the NumSets from integer case into a continuous case, where the uncertainty score is calculated based on a matrix trace~\cite{lin2023generating}.

\section{More Explanation of NLG metrics}
\label{app:nlg_metrics}

Below we give a detailed explanation of the NLG metrics in Table~\ref{tab:sum_nlg_metrics}.

\subsection{Specific NLG Metrics}
\noindent (1) \textbf{ROUGE-L}: measures the longest common subsequence between the generated text and ground-truth text~\cite{lin2004rouge}. It emphasizes relevance.

\noindent (2) \textbf{BARTSCORE}: uses an encoder-decoder pretrained model to compute a similarity score for each token in the generation with each token in the reference text~\citet{yuan2021bartscore}. It emphasizes consistency.

\noindent (3) \textbf{CTC}: uses a pretrained model to measure the alignment between the generation text and ground-truth text~\cite{deng2021compression}. It emphasizes consistency.

\noindent (4) \textbf{SummaC}: uses pretrained models to segment both generated and input texts into sentence units and aggregate scores between pairs of sentences~\cite{laban2022summac}. It emphasizes consistency.

\noindent (5) Spearman Correlation (\textbf{Spearman}) and (6) Kendall-Tau Correlation (\textbf{Kendall-Tau}) are designed to
measure the semantic overlap between the model output and the reference text via text embeddings. They have relatively high correlations in the relevance dimension~\cite{liu-etal-2021-explainaboard,zhong2022towards}.

\noindent (7-11) \textbf{UniEVAL}: takes NLG evaluation as a boolean question answering (QA) task and guides the model with different questions. UniEVAL~\cite{zhong2022towards} can use one evaluator to evaluate one of the dimensions in Sec.~\ref{sec:dim_concepts}.

\noindent (12-16) GPT-3.5 without dimension concepts \textbf{wo-GPT-3.5}: uses GPT-3.5~\cite{OPENAI35} to prompt about the generation quality. Among the prompts, we do not provide any prior knowledge about each dimension concept described in Sec.~\ref{sec:dim_concepts}. This method compares the difference between the generated summaries and \textit{ground-truth summaries}.

\noindent (17-21) GPT-3.5 with dimension concepts \textbf{wi-gt-GPT-3.5}: uses the GPT-3.5~\cite{OPENAI35} to prompt about the generation quality. However, the prior knowledge about each dimension concept is given in the prompt. This method compares the difference between the generated summaries and \textit{ground-truth summaries}.

\noindent (22-26) \textbf{wi-in-GPT-3.5}: is very similar to wi-gt-GPT-3.5. The only difference is that this method compares the difference between the generated summaries and \textit{input text}.

\noindent (27-31) \textbf{wi-ingt-GPT-3.5}: is very similar to wi-gt-GPT-3.5. The only difference is that this method compares generated summaries to the \textit{input text and ground-truth summaries}.

Although different NLP tasks may use various evaluation metrics, such as those in \citet{lin2024navigating}, we adopt the evaluation metric settings from \citet{zhong2022towards} due to their generalizability to other NLG tasks. Consequently, our research conclusions may be applicable to a broader range of NLG tasks.

\section{More Experimental Results}
In this subsection, we list more experimental results below. We first give the experimental results from the view of NLG metrics. Then, we show the experimental results involving human annotations.

\begin{table*}[]
\centering
\small
\begin{tabular}{l|rrrrrr}
\hline
\multicolumn{1}{l|}{\textbf{NLG Metrics}} & \multicolumn{1}{c}{\textbf{MSP}} & \multicolumn{1}{c}{\textbf{MTE}} & \multicolumn{1}{c}{\textbf{MCSE}} & \multicolumn{1}{c}{\textbf{MD}} & \multicolumn{1}{c}{\textbf{RDE}} & \multicolumn{1}{c}{\textbf{P(True)}} \\
\hline
ROUGE-L                                  & 0.1300                           & 0.1171                           & 0.1257                            & 0.0096                          & -0.0232                          & -0.1698                              \\
BARTSCORE                                & 0.1657                           & 0.2000                           & 0.1789                            & 0.2132                          & 0.1891                           & -0.1828                              \\
SummaC                                   & -0.0659                          & -0.0580                          & -0.0642                           & -0.1556                         & -0.2557                          & 0.0392                               \\
CTC                                      & 0.0448                           & -0.0112                          & 0.0699                            & -0.0021                         & -0.0759                          & -0.3161                              \\
Spearman                                 & 0.0189                           & 0.0516                           & 0.0094                            & 0.0569                          & 0.0190                           & -0.2855                              \\
Kendall-Tau                              & 0.0138                           & 0.0437                           & 0.0034                            & 0.0495                          & 0.0151                           & -0.2735                              \\
UniEval (Relevance)                      & 0.4496                           & 0.4654                           & 0.4586                            & 0.3709                          & 0.1655                           & -0.8063                              \\
UniEval (Consistency)                    & 0.5711                           & 0.5589                           & 0.5746                            & 0.4069                          & 0.2830                           & -0.7399                              \\
UniEval (Coherence)                      & 0.7198                           & 0.7069                           & 0.7219                            & 0.5884                          & 0.2795                           & -1.0973                              \\
UniEval (Fluency)                        & 0.3714                           & 0.3008                           & 0.3969                            & 0.3526                          & 0.2316                           & -0.5880                              \\
UniEval (Overall)                        & 0.6020                           & 0.5904                           & 0.6113                            & 0.4893                          & 0.2567                           & -0.9452                              \\
wo-GPT-3.5 (Relevance)                   & 0.3447                           & 0.3852                           & 0.3756                            & 0.2945                          & -0.0065                          & -0.3841                              \\
wo-GPT-3.5 (Consistency)                 & 0.0789                           & 0.1104                           & 0.0990                            & 0.0993                          & 0.0142                           & -0.0140                              \\
wo-GP-T3.5 (Coherence)                   & -0.0197                          & -0.0238                          & -0.0012                           & -0.0316                         & 0.0102                           & 0.0612                               \\
wo-GPT-3.5 (Fluency)                     & 0.1701                           & 0.1997                           & 0.1935                            & 0.1850                          & -0.0118                          & -0.0961                              \\
wo-GPT-3.5 (Overall)                     & 0.1249                           & 0.1626                           & 0.1442                            & 0.1562                          & 0.0138                           & -0.1296                              \\
wi-gt-GPT-3.5 (Relevance)                & -0.1070                          & -0.1089                          & -0.0933                           & -0.1349                         & -0.1222                          & -0.1304                              \\
wi-gt-GPT-3.5 (Consistency)              & -0.0985                          & -0.1148                          & -0.0832                           & -0.0625                         & -0.0538                          & -0.0531                              \\
wi-gt-GPT-3.5 (Coherence)                & -0.0546                          & -0.0504                          & -0.0270                           & -0.0813                         & -0.0471                          & -0.0909                              \\
wi-gt-GPT-3.5 (Fluency)                  & -0.1158                          & -0.1192                          & -0.1139                           & -0.1233                         & -0.1168                          & -0.1173                              \\
wi-gt-GPT-3.5 (Overall)                  & -0.0327                          & -0.0285                          & -0.0130                           & -0.0718                         & -0.0911                          & -0.0383                              \\
wi-in-GPT-3.5 (Relevance)                & -0.0105                          & -0.0084                          & -0.0035                           & -0.0576                         & -0.0280                          & -0.0266                              \\
wi-in-GPT-3.5 (Consistency)              & -0.0276                          & -0.0289                          & -0.0069                           & -0.0458                         & -0.0211                          & 0.0216                               \\
wi-in-GPT-3.5 (Coherence)                & -0.0454                          & -0.0456                          & -0.0234                           & -0.0526                         & -0.0171                          & 0.0429                               \\
wi-in-GPT-3.5 (Fluency)                  & -0.1349                          & -0.1423                          & -0.1161                           & -0.1589                         & -0.1039                          & 0.0210                               \\
wi-in-GPT-3.5 (Overall)                  & -0.1067                          & -0.1069                          & -0.0861                           & -0.1170                         & -0.0915                          & -0.0361                              \\
wi-ingt-GPT-3.5 (Relevance)              & -0.0269                          & -0.0038                          & -0.0295                           & -0.0894                         & -0.1017                          & -0.0178                              \\
wi-ingt-GPT-3.5 (Consistency)            & -0.0940                          & -0.0760                          & -0.0917                           & -0.1211                         & -0.0636                          & -0.0263                              \\
wi-ingt-GPT-3.5 (Coherence)              & -0.0587                          & -0.0155                          & -0.0530                           & -0.1187                         & -0.1089                          & -0.0441                              \\
wi-ingt-GPT-3.5 (Fluency)                & -0.1545                          & -0.1254                          & -0.1790                           & -0.1492                         & -0.1588                          & -0.0200                              \\
wi-ingt-GPT-3.5 (Overall)                & -0.0362                          & -0.0156                          & -0.0098                           & -0.0193                         & -0.0316                          & -0.0162                              \\
\hline
Col Mean                                 & 0.0844                           & 0.0906                           & \bf0.0957                            & 0.0542                          & -0.0017                          & -0.2084                             

\\
\hline
\end{tabular}
\caption{Main results of the relationship between the uncertainty estimation methods and NLG metrics on XSUM dataset using generation from Llama 2.}
\label{tab:main_xsum_llama}
\end{table*}

\begin{table*}[]
\centering
\small
\begin{tabular}{l|rrrr}
\hline
\multicolumn{1}{l|}{\textbf{NLG Metrics}} & \multicolumn{1}{c}{\textbf{NumSets}} & \multicolumn{1}{c}{\textbf{ECC}} & \multicolumn{1}{c}{\textbf{LexSim}} & \multicolumn{1}{c}{\textbf{EigV}} \\
\hline
ROUGE-L                                  & 0.0415                               & 0.1071                           & 0.2121                              & 0.0964                            \\
SummaC                                   & 0.0919                               & -0.1259                          & -0.1362                             & -0.0994                           \\
CTC                                      & 0.0387                               & 0.0999                           & 0.0506                              & 0.1132                            \\
Spearman                                 & 0.0326                               & 0.0101                           & 0.1344                              & 0.0112                            \\
Kendall-Tau                              & 0.0326                               & 0.0102                           & 0.1324                              & 0.0114                            \\
UniEval (Relevance)                      & 0.1131                               & -0.0728                          & -0.0165                             & -0.0747                           \\
UniEval (Consistency)                    & 0.0559                               & 0.0898                           & 0.1260                              & 0.1211                            \\
UniEval (Coherence)                      & 0.0992                               & 0.1668                           & 0.2259                              & 0.1754                            \\
UniEval (Fluency)                        & -0.0252                              & -0.1101                          & -0.2024                             & -0.0921                           \\
UniEval (Overall)                        & 0.0946                               & -0.0379                          & -0.0055                             & -0.0284                           \\
wo-GPT-3.5 (Relevance)                   & 0.0714                               & -0.0908                          & 0.1165                              & -0.1115                           \\
wo-GPT-3.5 (Consistency)                 & 0.0977                               & -0.0382                          & 0.0806                              & -0.0488                           \\
wo-GP-T3.5 (Coherence)                   & 0.1004                               & -0.0981                          & 0.0624                              & -0.1043                           \\
wo-GPT-3.5 (Fluency)                     & 0.1217                               & -0.1114                          & 0.0466                              & -0.1265                           \\
wo-GPT-3.5 (Overall)                     & 0.1008                               & -0.0548                          & 0.0785                              & -0.0734                           \\
wi-gt-GPT-3.5 (Relevance)                & 0.1465                               & -0.0595                          & 0.1097                              & -0.0893                           \\
wi-gt-GPT-3.5 (Consistency)              & 0.1002                               & -0.0101                          & 0.0344                              & -0.0330                           \\
wi-gt-GPT-3.5 (Coherence)                & 0.1418                               & -0.0267                          & 0.0512                              & -0.0519                           \\
wi-gt-GPT-3.5 (Fluency)                  & 0.1501                               & 0.0117                           & 0.1928                              & 0.0091                            \\
wi-gt-GPT-3.5 (Overall)                  & 0.1226                               & -0.0892                          & 0.0206                              & -0.0953                           \\
wi-in-GPT-3.5 (Relevance)                & 0.0303                               & -0.0291                          & 0.0783                              & -0.0334                           \\
wi-in-GPT-3.5 (Consistency)              & 0.0792                               & 0.1707                           & 0.2616                              & 0.1626                            \\
wi-in-GPT-3.5 (Coherence)                & 0.1428                               & 0.0940                           & 0.2092                              & 0.0840                            \\
wi-in-GPT-3.5 (Fluency)                  & 0.0167                               & 0.1683                           & 0.0820                              & 0.1765                            \\
wi-in-GPT-3.5 (Overall)                  & 0.0774                               & -0.1700                          & 0.0117                              & -0.1711                           \\
wi-ingt-GPT-3.5 (Relevance)              & 0.1462                               & -0.1226                          & 0.0659                              & -0.1473                           \\
wi-ingt-GPT-3.5 (Consistency)            & 0.0677                               & -0.1344                          & 0.0135                              & -0.1543                           \\
wi-ingt-GPT-3.5 (Coherence)              & 0.0395                               & -0.1203                          & 0.0317                              & -0.1438                           \\
wi-ingt-GPT-3.5 (Fluency)                & 0.0971                               & -0.0075                          & 0.1304                              & -0.0011                           \\
wi-ingt-GPT-3.5 (Overall)                & 0.1281                               & -0.0642                          & 0.0760                              & -0.0918                           \\
\hline
Col Mean                                 & \bf 0.0851                               & -0.0215                          & 0.0758                              & -0.0270                          
                
\\
\hline
\end{tabular}
\caption{Main results of the relationship between the uncertainty estimation methods and NLG metrics on AESLC dataset using generation from GPT-3.5.}
\label{tab:main_aes_gpt35}
\end{table*}

\begin{table*}[]
\centering
\small
\begin{tabular}{l|rrrr}
\hline
\multicolumn{1}{l|}{\textbf{NLG Metrics}} & \multicolumn{1}{c}{\textbf{NumSets}} & \multicolumn{1}{c}{\textbf{ECC}} & \multicolumn{1}{c}{\textbf{LexSim}} & \multicolumn{1}{c}{\textbf{EigV}} \\
\hline
ROUGE-L                                  & -0.0207                              & 0.0867                           & 0.0759                              & 0.0565                            \\
SummaC                                   & 0.0445                               & -0.0485                          & -0.1196                             & -0.0297                           \\
CTC                                      & 0.0058                               & 0.1267                           & 0.0278                              & 0.0956                            \\
Spearman                                 & -0.0045                              & -0.0721                          & -0.1082                             & -0.0781                           \\
Kendall-Tau                              & -0.0046                              & -0.0664                          & -0.1084                             & -0.0716                           \\
UniEval (Relevance)                      & 0.0468                               & -0.0740                          & -0.1099                             & -0.0728                           \\
UniEval (Consistency)                    & -0.0325                              & 0.3751                           & 0.4828                              & 0.3161                            \\
UniEval (Coherence)                      & -0.0033                              & 0.3327                           & 0.4128                              & 0.2822                            \\
UniEval (Fluency)                        & 0.0573                               & 0.2495                           & 0.1347                              & 0.2297                            \\
UniEval (Overall)                        & 0.0259                               & 0.1686                           & 0.1814                              & 0.1385                            \\
wo-GPT-3.5 (Relevance)                   & 0.0714                               & -0.1213                          & 0.0464                              & -0.1533                           \\
wo-GPT-3.5 (Consistency)                 & 0.0266                               & -0.0478                          & -0.0526                             & -0.0624                           \\
wo-GP-T3.5 (Coherence)                   & 0.0325                               & -0.0464                          & -0.0550                             & -0.0558                           \\
wo-GPT-3.5 (Fluency)                     & 0.0105                               & -0.0094                          & -0.0039                             & -0.0272                           \\
wo-GPT-3.5 (Overall)                     & -0.0007                              & -0.0300                          & -0.0094                             & -0.0379                           \\
wi-gt-GPT-3.5 (Relevance)                & -0.0598                              & -0.1062                          & -0.0832                             & -0.0932                           \\
wi-gt-GPT-3.5 (Consistency)              & -0.0066                              & -0.0486                          & -0.0582                             & -0.0516                           \\
wi-gt-GPT-3.5 (Coherence)                & -0.0918                              & -0.0589                          & -0.0727                             & -0.0541                           \\
wi-gt-GPT-3.5 (Fluency)                  & -0.0402                              & -0.1248                          & -0.0733                             & -0.1125                           \\
wi-gt-GPT-3.5 (Overall)                  & -0.0563                              & -0.0768                          & -0.0341                             & -0.0771                           \\
wi-in-GPT-3.5 (Relevance)                & 0.0122                               & -0.0074                          & -0.0050                             & -0.0112                           \\
wi-in-GPT-3.5 (Consistency)              & -0.0130                              & -0.0686                          & -0.0892                             & -0.0739                           \\
wi-in-GPT-3.5 (Coherence)                & -0.0295                              & -0.0117                          & -0.0832                             & -0.0214                           \\
wi-in-GPT-3.5 (Fluency)                  & 0.0059                               & -0.1526                          & -0.1198                             & -0.1627                           \\
wi-in-GPT-3.5 (Overall)                  & -0.0506                              & -0.1875                          & -0.1962                             & -0.1827                           \\
wi-ingt-GPT-3.5 (Relevance)              & -0.0152                              & -0.0612                          & -0.0431                             & -0.0502                           \\
wi-ingt-GPT-3.5 (Consistency)            & -0.0176                              & -0.0290                          & -0.0326                             & -0.0167                           \\
wi-ingt-GPT-3.5 (Coherence)              & -0.0870                              & -0.1265                          & -0.0924                             & -0.1268                           \\
wi-ingt-GPT-3.5 (Fluency)                & -0.1197                              & -0.0844                          & -0.0136                             & -0.0609                           \\
wi-ingt-GPT-3.5 (Overall)                & -0.0136                              & -0.1040                          & -0.1022                             & -0.0979                           \\
\hline
Col Mean                                 & -0.0109                              & -0.0142                          & \bf-0.0101                             & -0.0221                          
   
\\
\hline
\end{tabular}
\caption{Main results of the relationship between the uncertainty estimation methods and NLG metrics on XSUM dataset using generation from GPT-3.5.}
\label{tab:main_xsum_gpt35}
\end{table*}

\begin{table*}[]
\centering
\small
\begin{tabular}{l|rrrrrrrrr}
\hline
\multicolumn{1}{l|}{\textbf{NLG Metrics}} & \multicolumn{1}{c}{\textbf{MSP}} & \multicolumn{1}{c}{\textbf{MTE}} & \multicolumn{1}{c}{\textbf{MCSE}} & \multicolumn{1}{c}{\textbf{MD}} & \multicolumn{1}{c}{\textbf{RDE}} & \multicolumn{1}{c}{\textbf{T-TU}} & \multicolumn{1}{c}{\textbf{T-RMI}} & \multicolumn{1}{c}{\textbf{S-TU}} & \multicolumn{1}{c}{\textbf{S-RMI}} \\
\hline
ROUGE-L                                  & 0.1763                           & -0.0385                          & 0.1539                            & -0.0333                         & -0.0844                          & -0.0571                           & -0.0752                            & -0.0849                           & 0.0577                             \\
BARTSCORE                                & -0.0541                          & 0.0314                           & 0.0252                            & 0.1228                          & 0.0930                           & 0.0525                            & -0.0867                            & 0.0026                            & 0.0590                             \\
SummaC                                   & 0.0106                           & -0.0024                          & 0.0304                            & 0.0215                          & -0.0719                          & -0.0247                           & -0.0422                            & 0.0229                            & 0.0267                             \\
CTC                                      & 0.1101                           & -0.2225                          & 0.0569                            & -0.2217                         & -0.2708                          & -0.1256                           & -0.0538                            & -0.1675                           & 0.0765                             \\
Spearman                                 & 0.0581                           & 0.0108                           & 0.1008                            & 0.0358                          & -0.0096                          & 0.0393                            & -0.0444                            & -0.0038                           & 0.0472                             \\
Kendall-Tau                              & 0.0659                           & 0.0080                           & 0.1043                            & 0.0316                          & -0.0141                          & 0.0345                            & -0.0443                            & -0.0077                           & 0.0458                             \\
UniEval (Relevance)                      & -0.0262                          & 0.1157                           & 0.0480                            & 0.0851                          & 0.0233                           & 0.1165                            & 0.1175                             & 0.1388                            & 0.1046                             \\
UniEval (Consistency)                    & 0.3493                           & 0.4330                           & 0.2544                            & 0.3288                          & 0.2752                           & 0.2014                            & 0.0470                             & 0.2281                            & 0.1353                             \\
UniEval (Coherence)                      & 0.4194                           & 0.5342                           & 0.2205                            & 0.4139                          & 0.3656                           & 0.2693                            & 0.0839                             & 0.3005                            & 0.0952                             \\
UniEval (Fluency)                        & 0.1141                           & 0.0538                           & 0.0782                            & -0.0489                         & -0.0556                          & 0.0228                            & 0.1079                             & 0.0917                            & 0.0844                             \\
UniEval (Overall)                        & 0.1508                           & 0.2423                           & 0.1274                            & 0.1586                          & 0.1071                           & 0.1490                            & 0.1154                             & 0.1902                            & 0.1172                             \\
wo-GPT-3.5 (Relevance)                   & -0.0183                          & 0.1433                           & 0.0632                            & 0.1613                          & 0.0862                           & 0.0948                            & -0.1168                            & 0.0700                            & 0.0527                             \\
wo-GPT-3.5 (Consistency)                 & 0.1177                           & 0.1945                           & 0.0912                            & 0.1519                          & 0.1255                           & 0.0422                            & -0.0873                            & 0.0435                            & 0.0546                             \\
wo-GP-T3.5 (Coherence)                   & 0.2006                           & 0.2308                           & 0.0798                            & 0.1344                          & 0.1176                           & 0.0535                            & -0.0777                            & 0.0568                            & 0.0143                             \\
wo-GPT-3.5 (Fluency)                     & 0.1277                           & 0.2753                           & 0.0995                            & 0.2308                          & 0.1792                           & 0.0913                            & -0.0589                            & 0.1319                            & 0.0489                             \\
wo-GPT-3.5 (Overall)                     & 0.0287                           & 0.2327                           & 0.0661                            & 0.1784                          & 0.1434                           & 0.1107                            & -0.1695                            & 0.0788                            & 0.0688                             \\
\hline
Col Mean                                 & 0.1144                           & \bf0.1402                           & 0.1000                            & 0.1094                          & 0.0631                           & 0.0669                            & -0.0241                            & 0.0682                            & 0.0681   
              
\\
\hline
\end{tabular}
\caption{Main results of the relationship between the uncertainty estimation methods and NLG metrics on AESLC dataset using generation from BART.}
\label{tab:main_aes_bart}
\end{table*}

\begin{table*}[]
\centering
\small
\begin{tabular}{l|rrrrrrrrr}
\hline
\multicolumn{1}{l|}{\textbf{NLG Metrics}} & \multicolumn{1}{c}{\textbf{MSP}} & \multicolumn{1}{c}{\textbf{MTE}} & \multicolumn{1}{c}{\textbf{MCSE}} & \multicolumn{1}{c}{\textbf{MD}} & \multicolumn{1}{c}{\textbf{RDE}} & \multicolumn{1}{c}{\textbf{T-TU}} & \multicolumn{1}{c}{\textbf{T-RMI}} & \multicolumn{1}{c}{\textbf{S-TU}} & \multicolumn{1}{c}{\textbf{S-RMI}} \\
\hline
ROUGE-L                                  & -0.0978                          & -0.1359                          & -0.0587                           & -0.0375                         & -0.0996                          & -0.0511                           & -0.0126                            & -0.0733                           & -0.0041                            \\
BARTSCORE                                & -0.0558                          & 0.0009                           & 0.0752                            & 0.1515                          & 0.1324                           & 0.0037                            & 0.1136                             & -0.0457                           & -0.0040                            \\
SummaC                                   & 0.0760                           & 0.0148                           & 0.0696                            & -0.0004                         & -0.0542                          & -0.0432                           & 0.0611                             & -0.0227                           & -0.0144                            \\
CTC                                      & -0.0486                          & -0.3754                          & -0.0001                           & -0.2301                         & -0.3298                          & -0.1879                           & 0.0176                             & -0.2027                           & -0.0095                            \\
Spearman                                 & -0.2040                          & -0.1310                          & -0.1209                           & 0.0028                          & -0.0424                          & -0.0579                           & -0.0500                            & -0.0895                           & -0.0048                            \\
Kendall-Tau                              & -0.2013                          & -0.1321                          & -0.1206                           & -0.0005                         & -0.0452                          & -0.0586                           & -0.0511                            & -0.0906                           & -0.0048                            \\
UniEval (Relevance)                      & -0.1475                          & -0.1092                          & -0.0562                           & -0.0126                         & -0.0966                          & -0.0858                           & 0.0156                             & -0.1038                           & -0.0235                            \\
UniEval (Consistency)                    & 0.7162                           & 0.7064                           & 0.6192                            & 0.5025                          & 0.5037                           & 0.1936                            & 0.1523                             & 0.2284                            & 0.0223                             \\
UniEval (Coherence)                      & 0.6467                           & 0.6474                           & 0.5416                            & 0.4556                          & 0.4740                           & 0.1358                            & 0.1423                             & 0.1790                            & 0.0095                             \\
UniEval (Fluency)                        & 0.0345                           & -0.1334                          & 0.0338                            & 0.0493                          & 0.0469                           & -0.1619                           & 0.1130                             & -0.2010                           & 0.0375                             \\
UniEval (Overall)                        & 0.2380                           & 0.2421                           & 0.2456                            & 0.2264                          & 0.1792                           & 0.0076                            & 0.0920                             & 0.0113                            & -0.0033                            \\
wo-GPT-3.5 (Relevance)                   & 0.0667                           & 0.1933                           & 0.0761                            & 0.1921                          & 0.1629                           & 0.0292                            & -0.0060                            & 0.0261                            & 0.0081                             \\
wo-GPT-3.5 (Consistency)                 & 0.1220                           & 0.1545                           & 0.0544                            & 0.0828                          & 0.0777                           & 0.0235                            & -0.0382                            & 0.0299                            & 0.0051                             \\
wo-GP-T3.5 (Coherence)                   & 0.1436                           & 0.2104                           & 0.0479                            & 0.1176                          & 0.1313                           & 0.0454                            & -0.0740                            & 0.0527                            & 0.0046                             \\
wo-GPT-3.5 (Fluency)                     & 0.1924                           & 0.2679                           & 0.0675                            & 0.1547                          & 0.1815                           & 0.0769                            & -0.0528                            & 0.0816                            & 0.0102                             \\
wo-GPT-3.5 (Overall)                     & 0.1137                           & 0.2039                           & 0.0448                            & 0.1446                          & 0.1546                           & 0.0848                            & -0.1038                            & 0.0681                            & 0.0115                             \\
\hline
Col Mean                                 & 0.0997                           & 0.1015                           & 0.0950                            & \bf0.1124                          & 0.0860                           & -0.0029                           & 0.0199                             & -0.0095                           & 0.0025

\\
\hline
\end{tabular}
\caption{Main results of the relationship between the uncertainty estimation methods and NLG metrics on XSUM dataset using generation from BART.}
\label{tab:main_xsum_bart}
\end{table*}


\begin{figure*}[!htbp]
\centering
\includegraphics[width=\textwidth]{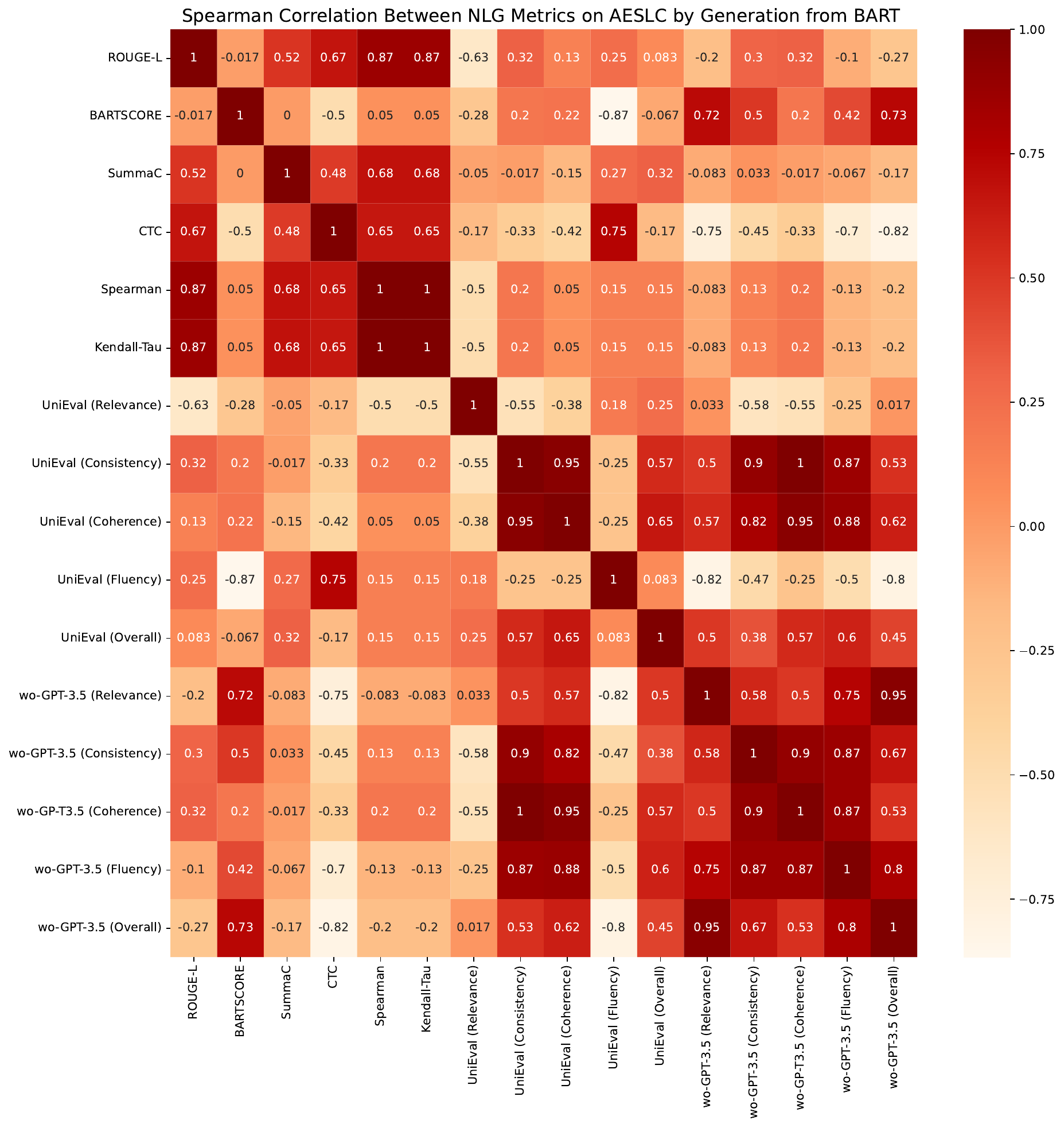}
\caption{Diagram of Spearman correlation between NLG metrics on AESLC dataset from the view of uncertainty estimation methods used in Fig.~\ref{fig:spear_ue_aes_bart}.  The generated summaries are from BART. For the GPT-3.5-based NLG metrics, we only conduct wo-GPT-3.5 on the BART generation model setting.}
\label{fig:spear_nlg_aes_bart}
\end{figure*}

\begin{figure*}[!htbp]
\centering
\includegraphics[width=\textwidth]{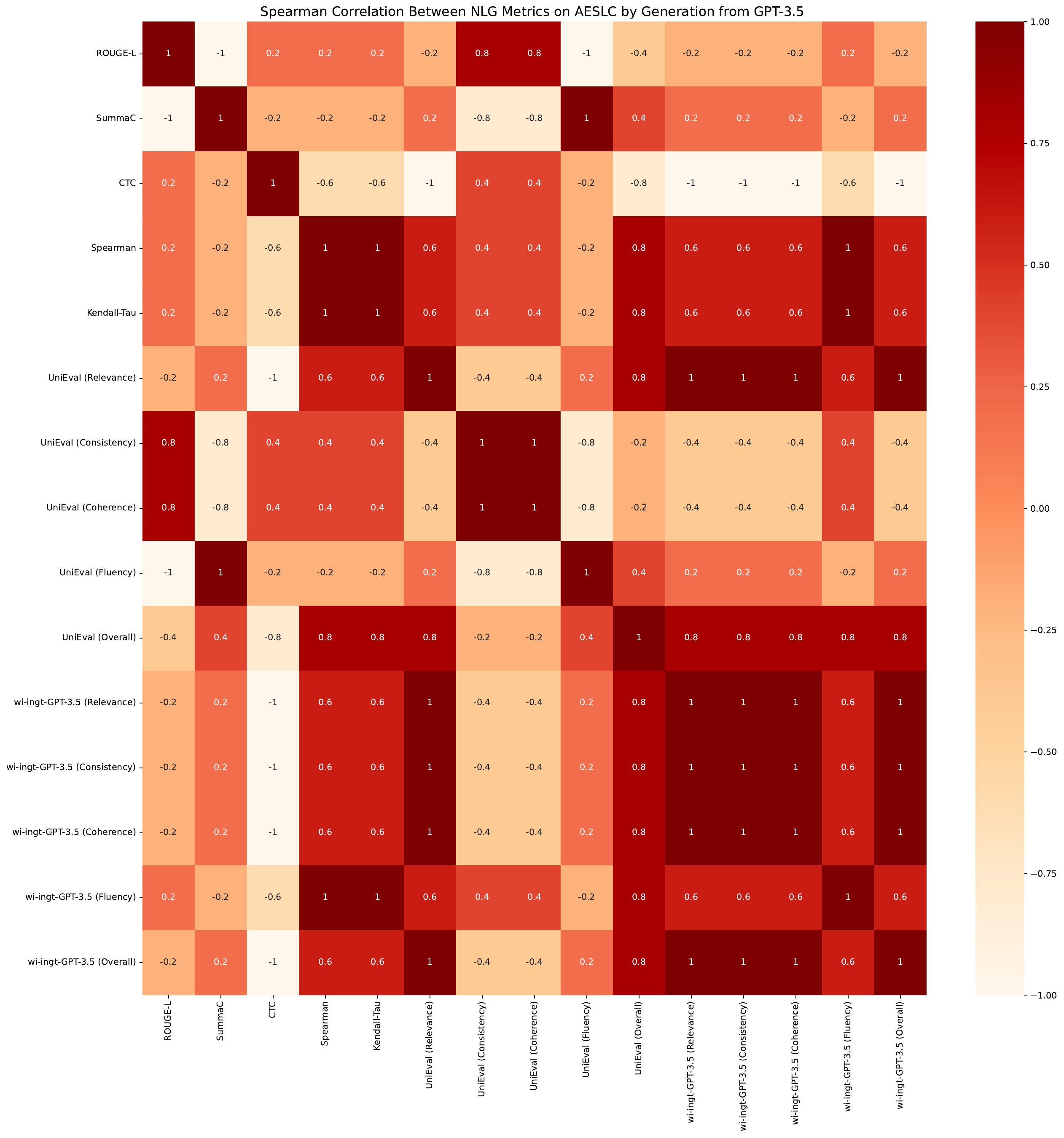}
\caption{Diagram of Spearman correlation between NLG metrics on AESLC dataset from the view of uncertainty estimation methods used in Fig.~\ref{fig:spear_ue_aes_gpt35}. The generated summaries are from GPT-3.5. For the GPT-3.5-based NLG metrics, we only draw wi-ingt-GPT-3.5 results to save space.}
\label{fig:spear_nlg_aes_gpt35}
\end{figure*}

\begin{figure*}[!htbp]
\centering
\includegraphics[width=\textwidth]{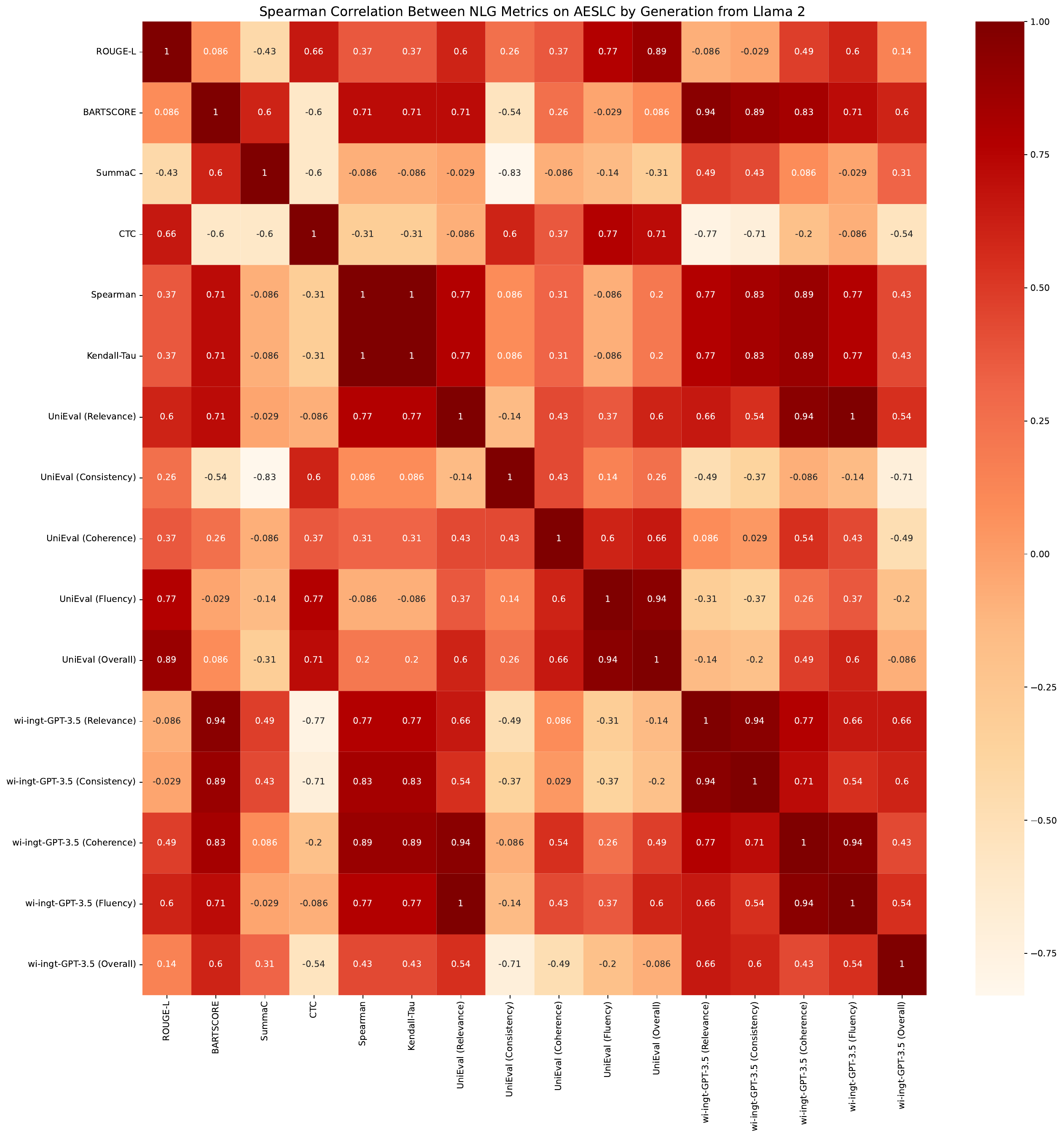}
\caption{Diagram of Spearman correlation between NLG metrics on AESLC dataset from the view of uncertainty estimation methods used in Fig.~\ref{fig:spear_ue_aes_llama}. The generated summaries are from Llama 2. For the GPT-3.5-based NLG metrics, we only draw wi-ingt-GPT-3.5 results to save space.}
\label{fig:spear_nlg_aes_llama}
\end{figure*}

\begin{figure*}[!htbp]
\centering
\includegraphics[width=\textwidth]{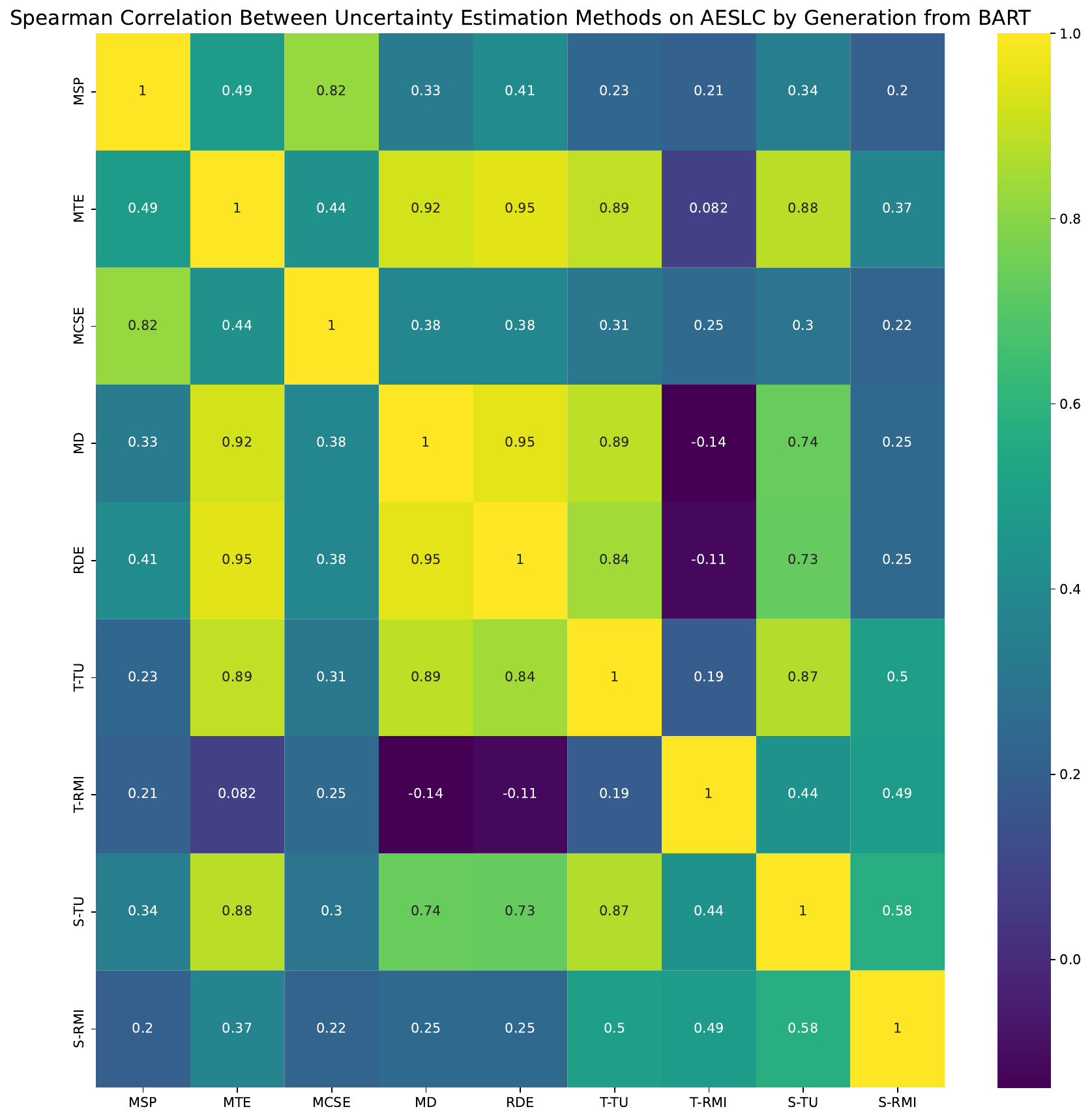}
\caption{Diagram of Spearman correlation between uncertainty estimation methods on AESLC dataset from the view of NLG metrics used in Fig.~\ref{fig:spear_nlg_aes_bart}. The generated summaries are from BART.}
\label{fig:spear_ue_aes_bart}
\end{figure*}

\begin{figure*}[!htbp]
\centering
\includegraphics[width=\textwidth]{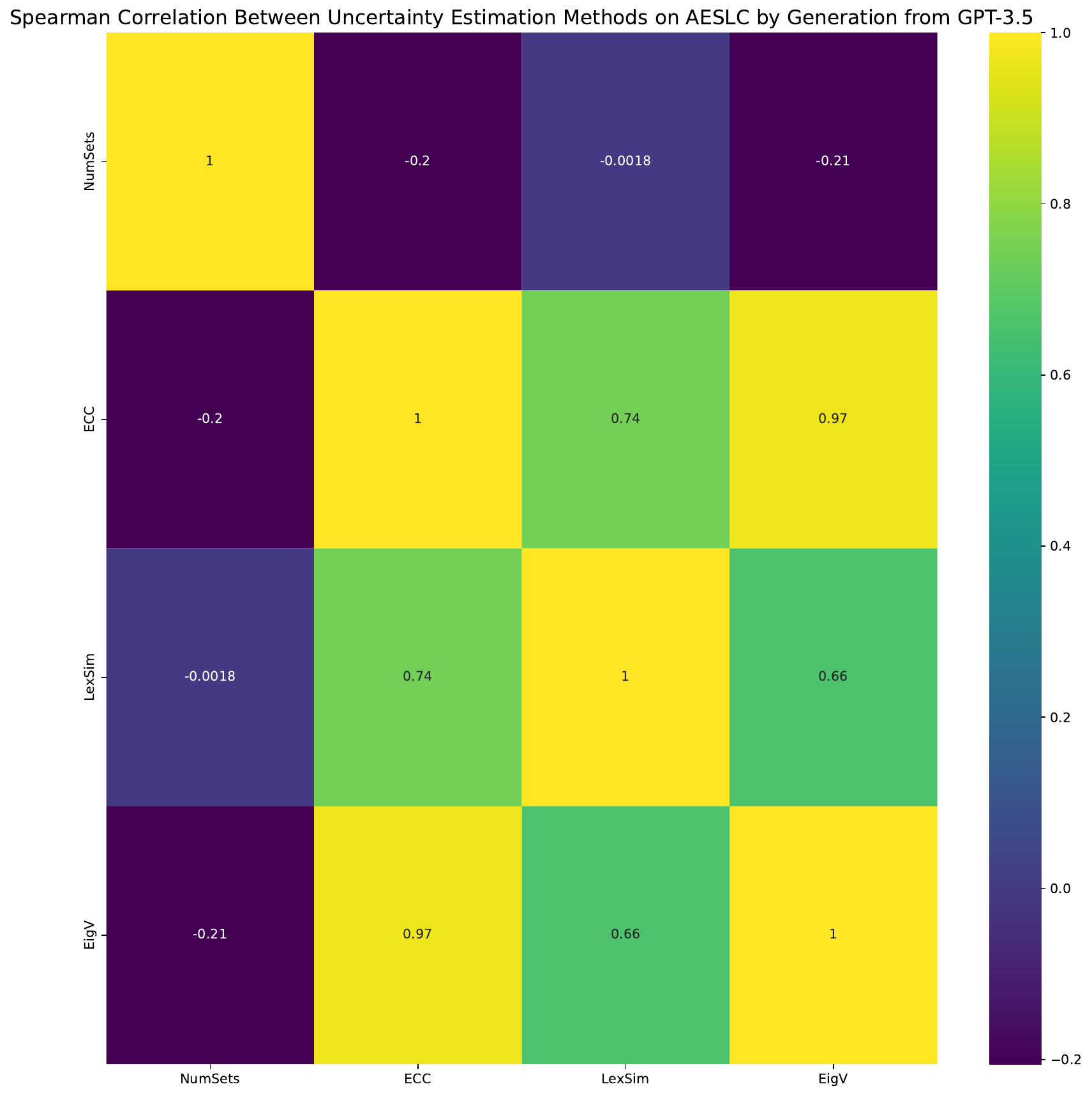}
\caption{Diagram of Spearman correlation between uncertainty estimation methods on AESLC dataset from the view of NLG metrics used in Fig.~\ref{fig:spear_nlg_aes_gpt35}.  The generated summaries are from GPT-3.5.}
\label{fig:spear_ue_aes_gpt35}
\end{figure*}

\begin{figure*}[!htbp]
\centering
\includegraphics[width=\textwidth]{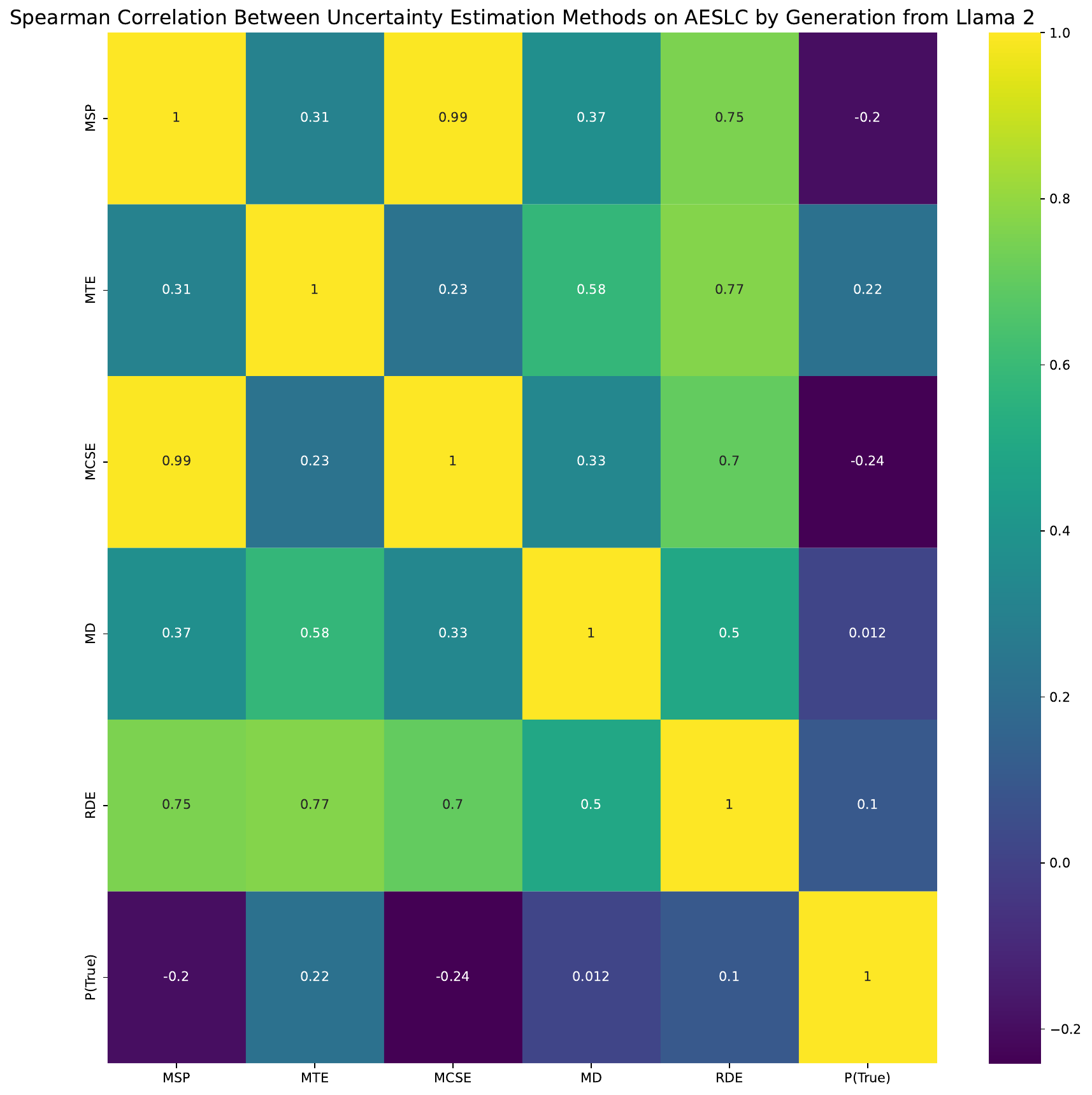}
\caption{Diagram of Spearman correlation between uncertainty estimation methods on AESLC dataset from the view of NLG metrics used in Fig.~\ref{fig:spear_nlg_aes_llama}.  The generated summaries are from Llama 2.}
\label{fig:spear_ue_aes_llama}
\end{figure*}


\begin{figure*}[!htbp]
\centering
\includegraphics[width=\textwidth]{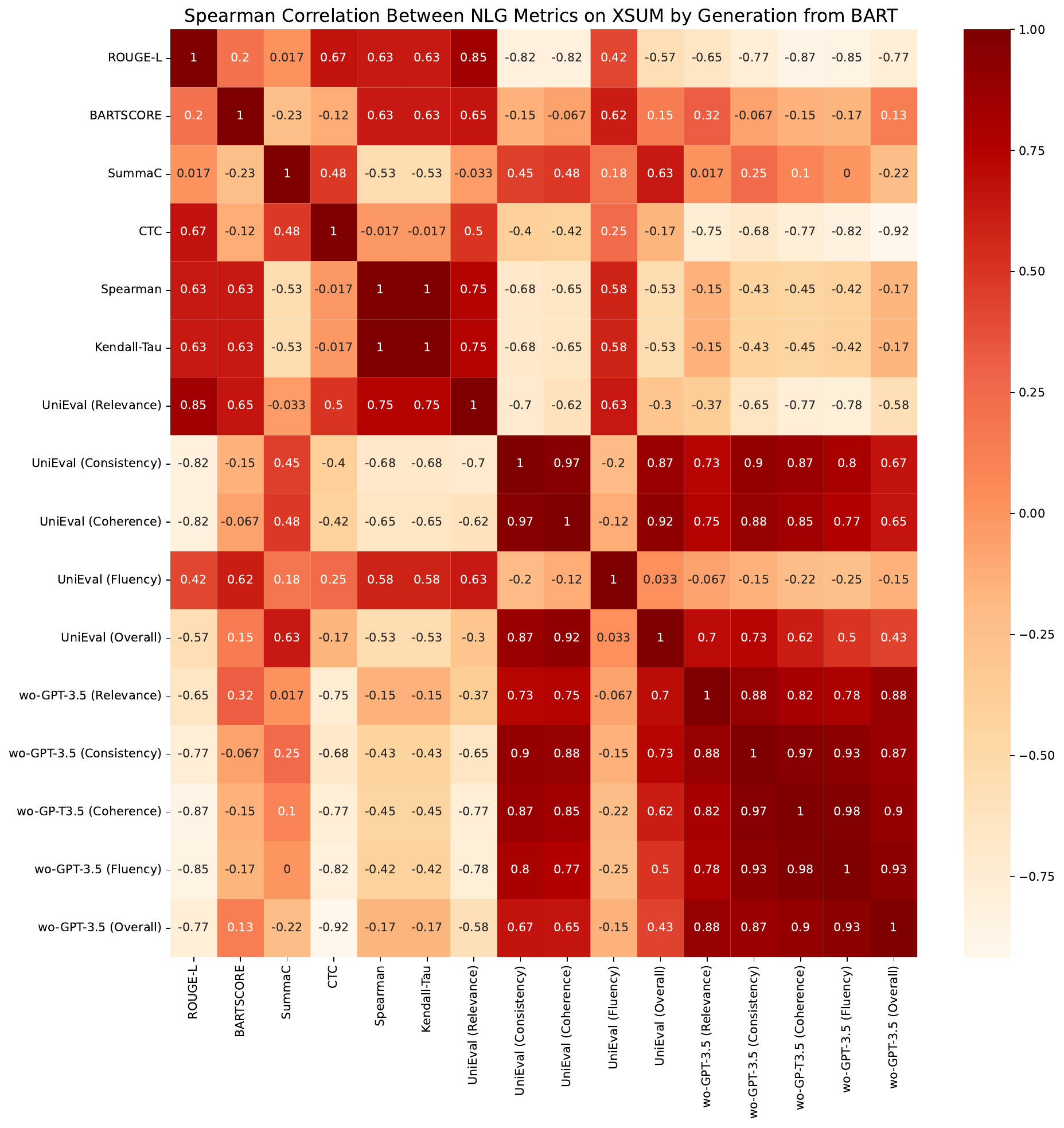}
\caption{Diagram of Spearman correlation between NLG metrics on XSUM dataset from the view of uncertainty estimation methods used in Fig.~\ref{fig:spear_ue_xsum_bart}.  The generated summaries are from BART. For the GPT-3.5-based NLG metrics, we only conduct wo-GPT-3.5 on the BART generation model setting.}
\label{fig:spear_nlg_xsum_bart}
\end{figure*}

\begin{figure*}[!htbp]
\centering
\includegraphics[width=\textwidth]{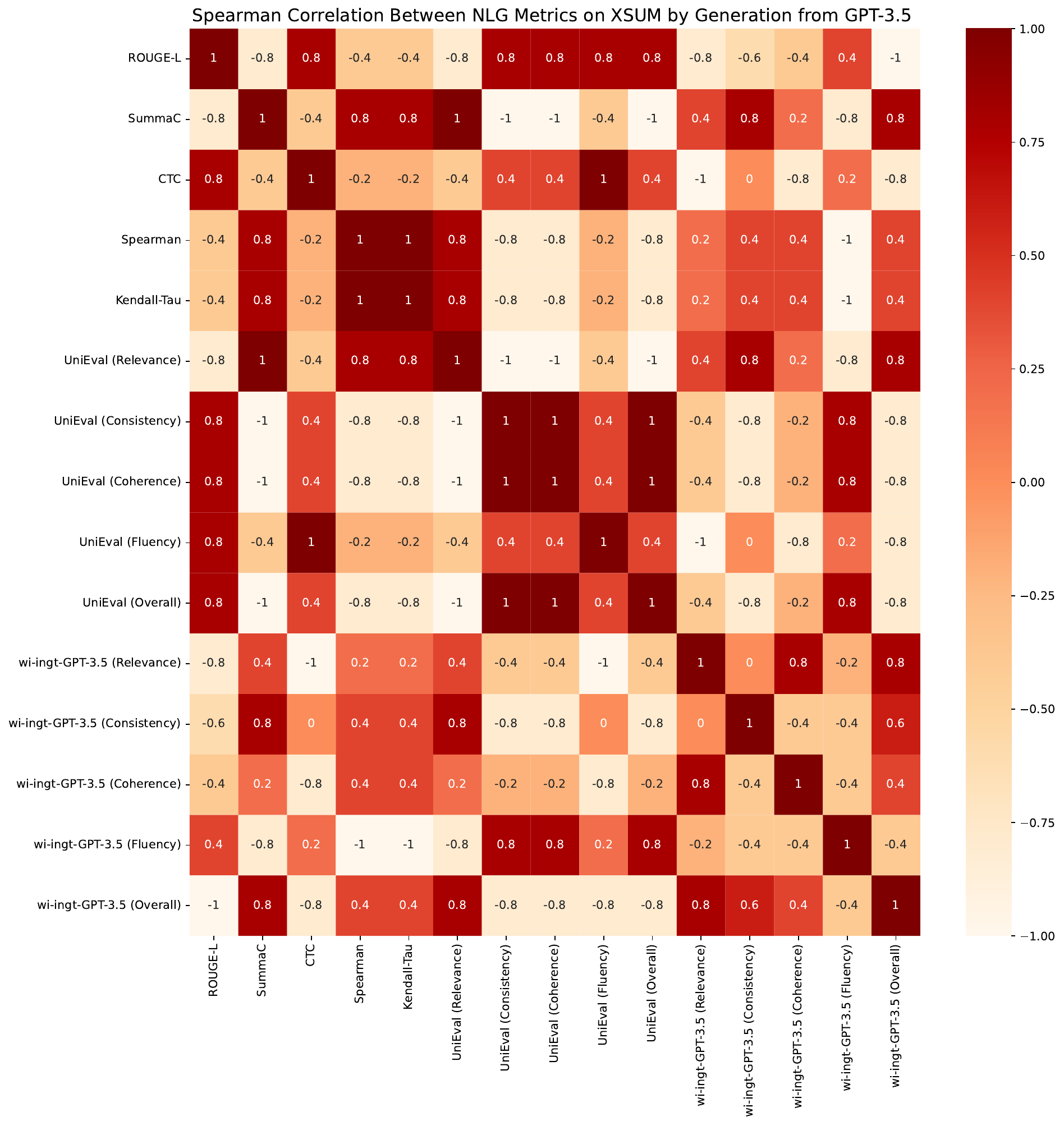}
\caption{Diagram of Spearman correlation between NLG metrics on XSUM dataset from the view of uncertainty estimation methods used in Fig.~\ref{fig:spear_ue_xsum_gpt35}. The generated summaries are from GPT-3.5. For the GPT-3.5-based NLG metrics, we only draw wi-ingt-GPT-3.5 results to save space.}
\label{fig:spear_nlg_xsum_gpt35}
\end{figure*}

\begin{figure*}[!htbp]
\centering
\includegraphics[width=\textwidth]{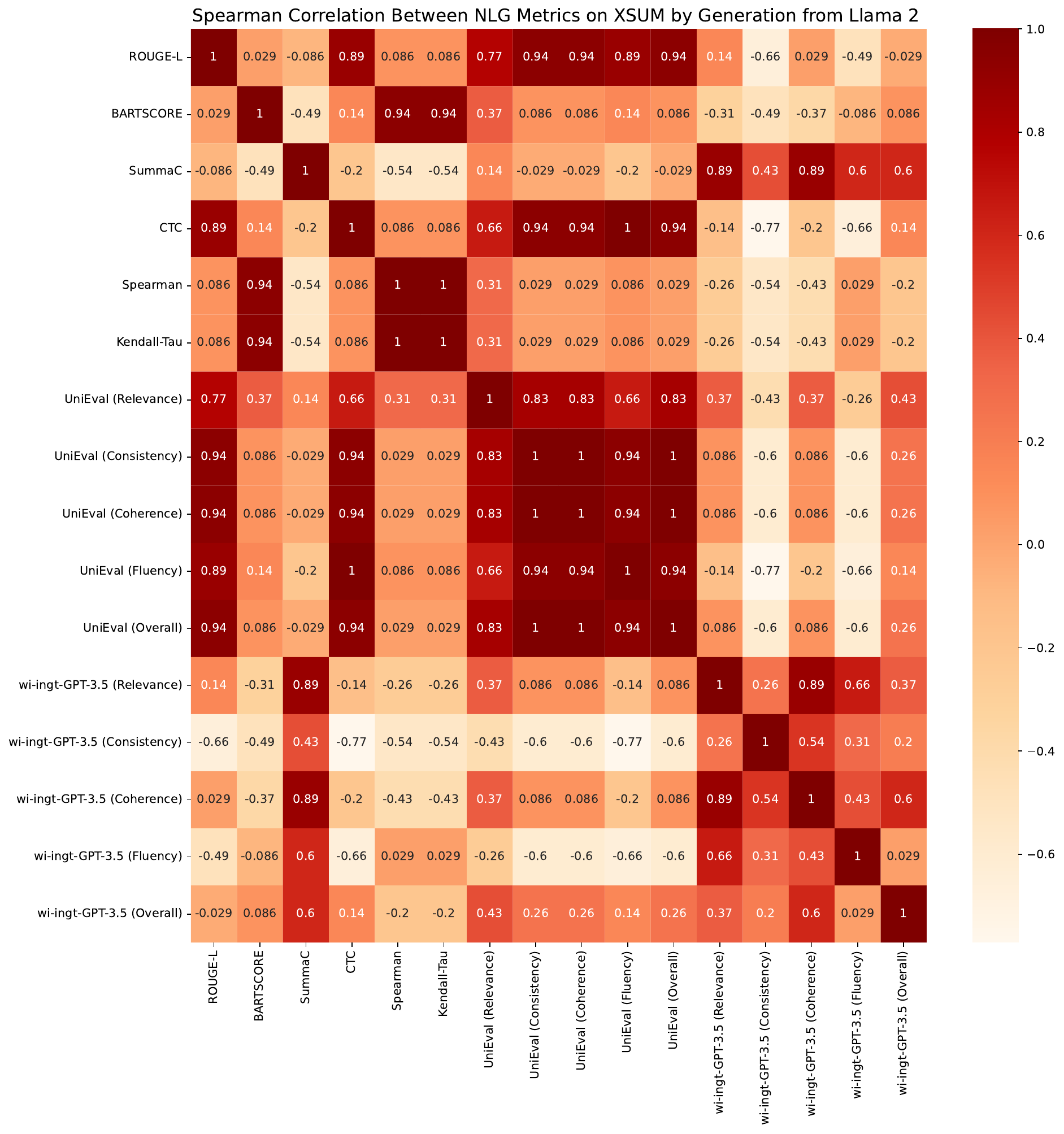}
\caption{Diagram of Spearman correlation between NLG metrics on XSUM dataset from the view of uncertainty estimation methods used in Fig.~\ref{fig:spear_ue_xsum_llama}. The generated summaries are from Llama 2. For the GPT-3.5-based NLG metrics, we only draw wi-ingt-GPT-3.5 results to save space.}
\label{fig:spear_nlg_xsum_llama}
\end{figure*}

\begin{figure*}[!htbp]
\centering
\includegraphics[width=\textwidth]{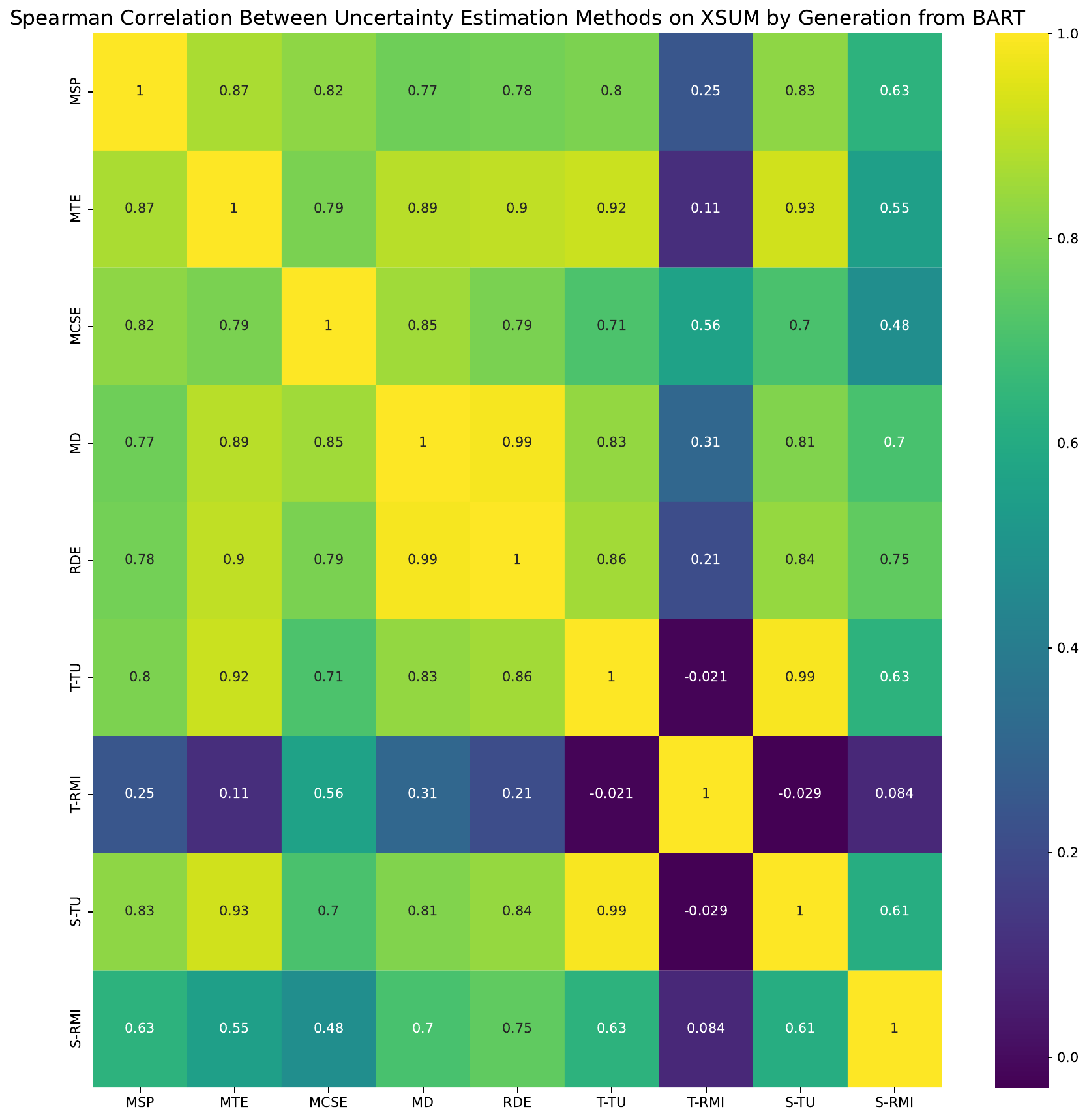}
\caption{Diagram of Spearman correlation between uncertainty estimation methods on XSUM dataset from the view of NLG metrics used in Fig.~\ref{fig:spear_nlg_xsum_bart}. The generated summaries are from BART.}
\label{fig:spear_ue_xsum_bart}
\end{figure*}

\begin{figure*}[!htbp]
\centering
\includegraphics[width=\textwidth]{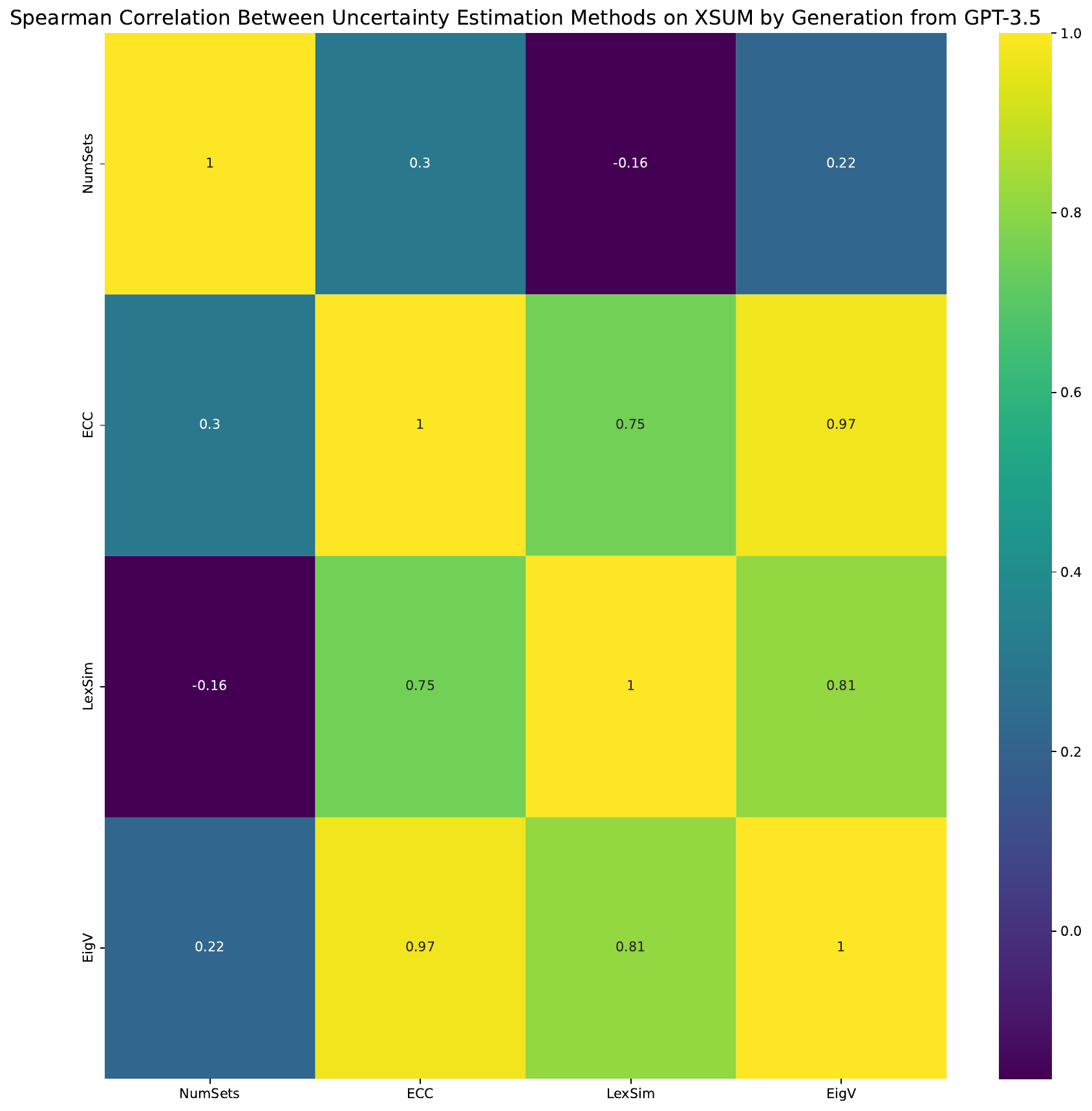}
\caption{Diagram of Spearman correlation between uncertainty estimation methods on XSUM dataset from the view of NLG metrics used in Fig.~\ref{fig:spear_nlg_xsum_gpt35}. The generated summaries are from GPT-3.5.}
\label{fig:spear_ue_xsum_gpt35}
\end{figure*}

\begin{figure*}[!htbp]
\centering
\includegraphics[width=\textwidth]{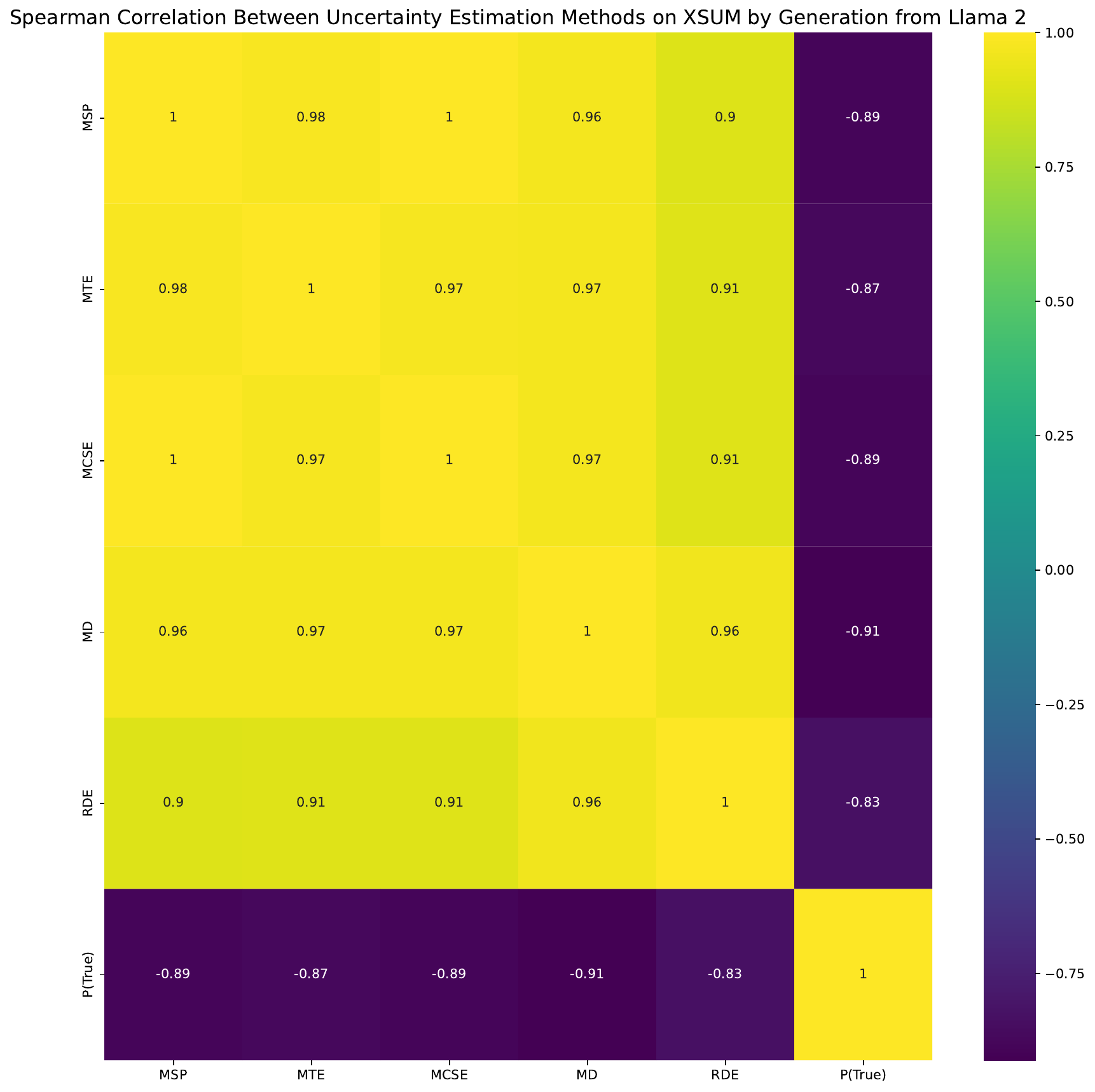}
\caption{Diagram of Spearman correlation between uncertainty estimation methods on XSUM dataset from the view of NLG metrics used in Fig.~\ref{fig:spear_nlg_xsum_llama}. The generated summaries are from Llama 2.}
\label{fig:spear_ue_xsum_llama}
\end{figure*}

\subsection{Experimental Results about NLG Metrics}
\label{sec:exp_res_ana}

Tables~\ref{tab:main_aes_bart},~\ref{tab:main_aes_gpt35},~\ref{tab:main_aes_llama},~\ref{tab:main_xsum_bart},~\ref{tab:main_xsum_gpt35}, and~\ref{tab:main_xsum_llama} present various uncertainty metric scores obtained from different uncertainty methods alongside different NLG metrics.
Figures~\ref{fig:spear_ue_aes_bart},~\ref{fig:spear_ue_aes_gpt35},~\ref{fig:spear_ue_aes_llama},~\ref{fig:spear_ue_xsum_bart},~\ref{fig:spear_ue_xsum_gpt35}, and~\ref{fig:spear_ue_xsum_llama} illustrate the correlation of uncertainty methods in terms of uncertainty metric scores.
Additionally, Figures~\ref{fig:spear_nlg_aes_bart},~\ref{fig:spear_nlg_aes_gpt35},~\ref{fig:spear_nlg_aes_llama},~\ref{fig:spear_nlg_xsum_bart},~\ref{fig:spear_nlg_xsum_gpt35}, and~\ref{fig:spear_nlg_xsum_llama} depict the correlation of NLG metrics with uncertainty metric scores.

Based on the information presented in these tables and figures, we can address our key question: 
``How does the choice of NLG metric affect the evaluation of uncertainty estimation methods in text summarization?''
The answer is that \textbf{using different NLG metrics could lead to different ranks for uncertainty estimation methods}. Therefore, it is important to design uncertainty estimation metrics that are robust across various NLG metrics.

Next, we provide a detailed analysis from the perspectives of NLG metrics and uncertainty estimation methods, respectively.

\subsubsection{Analysis Based on All NLG Metrics}
\label{sec:ana_all_nlg}
\noindent\textbf{Analysis from all five dimensions.} Figures~\ref{fig:spear_nlg_aes_bart},~\ref{fig:spear_nlg_aes_gpt35}, and~\ref{fig:spear_nlg_aes_llama} show the Spearman correlation between NLG metrics in a comprehensive view of the AESLC dataset. Figures~\ref{fig:spear_nlg_xsum_bart},~\ref{fig:spear_nlg_xsum_gpt35}, and~\ref{fig:spear_nlg_xsum_llama} show the Spearman correlation between NLG metrics in a comprehensive view of the XSUM dataset. We use Spearman correlation~\cite{myers2004spearman,zheng2022analysis} rather than Pearson correlation~\cite{sedgwick2012pearson,xue2022relationships}. This is because the Spearman correlation is calculated based on the ranks of the data while the Pearson correlation is calculated via the raw data values. Since our work wants to know the impact on the performance rank of the different uncertainty estimation methods. We use the Spearman correlation.
From these figures, we can draw the following conclusions.

It is evident that evaluating uncertainty estimation models using different NLG metrics leads to variations in the performance ranking of these models. This discrepancy arises because there are no rows or columns in these figures where all elements are greater than 0.5. Thus, the reliability of evaluating uncertainty estimation models becomes a concern.

Additionally, some evaluations of uncertainty estimation models using different NLG metrics may result in different performance rankings. For instance, in Figure~\ref{fig:spear_nlg_xsum_gpt35}, it is observed that the correlation between ROUGE-L and wi-ingt-GPT-3.5 (overall) is -1, indicating a completely different ranking.

However, some evaluations of uncertainty estimation models using different NLG metrics could result in the same performance ranks. For instance, Figure~\ref{fig:spear_nlg_aes_gpt35} shows that the correlation between wi-ingt-GPT-3.5 (Fluency) and UniEval (Relevance) is 1, indicating identical ranks.

\subsubsection{Analysis on NLG Relevance Dimension}
\label{sec:app_nlg_relevance_dimension}

\begin{figure*}[!htbp]
\centering
\includegraphics[width=\textwidth]{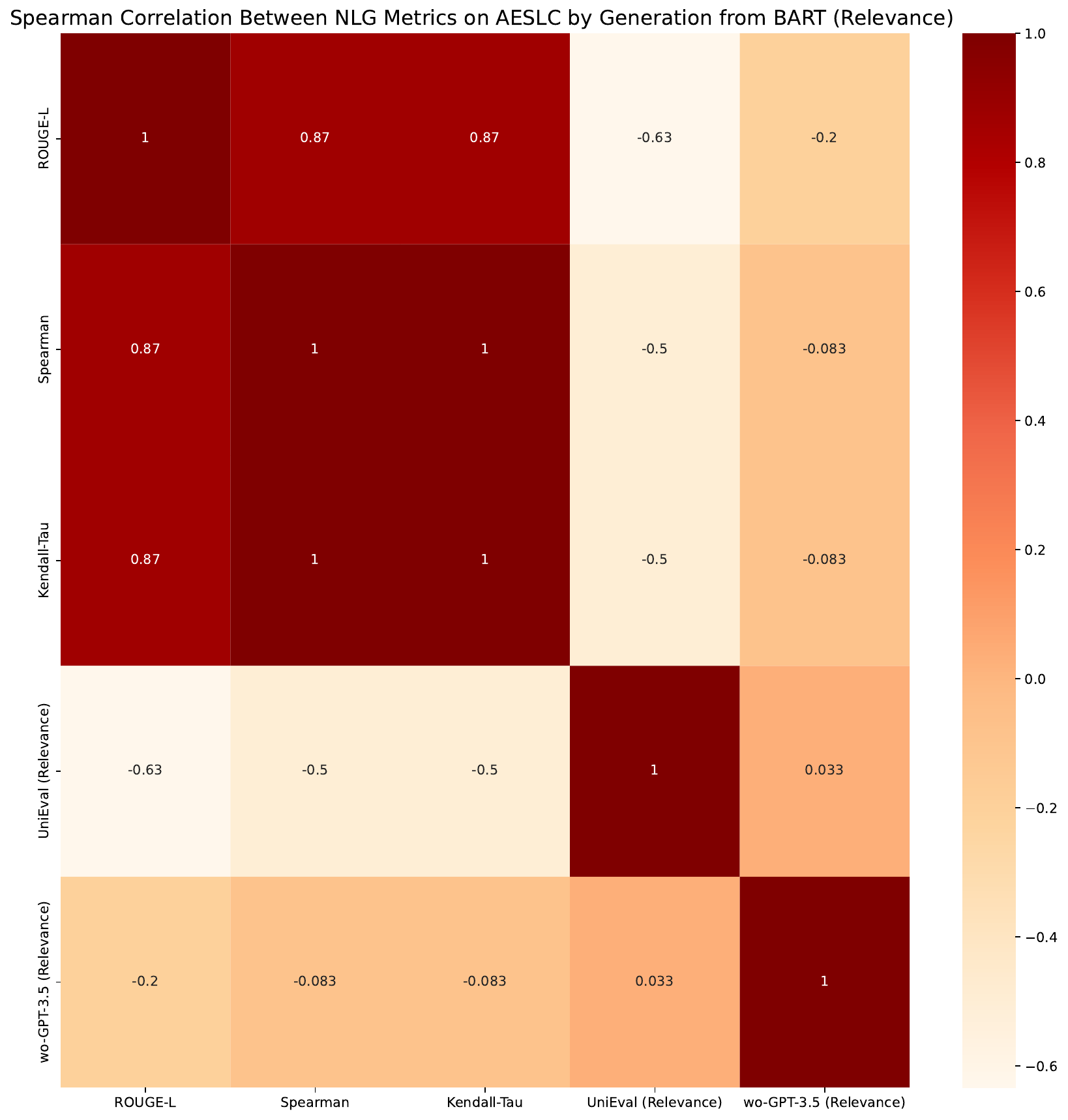}
\caption{Diagram of Spearman correlation in terms of relevance between NLG metrics on AESLC dataset from the view of uncertainty estimation methods used in Fig.~\ref{fig:spear_ue_aes_bart}.  The generated summaries are from BART. For the GPT-3.5-based NLG metrics, we only conduct wo-GPT-3.5 on the BART generation model setting.}
\label{fig:spear_nlg_aes_relevance_bart}
\end{figure*}

\begin{figure*}[!htbp]
\centering
\includegraphics[width=\textwidth]{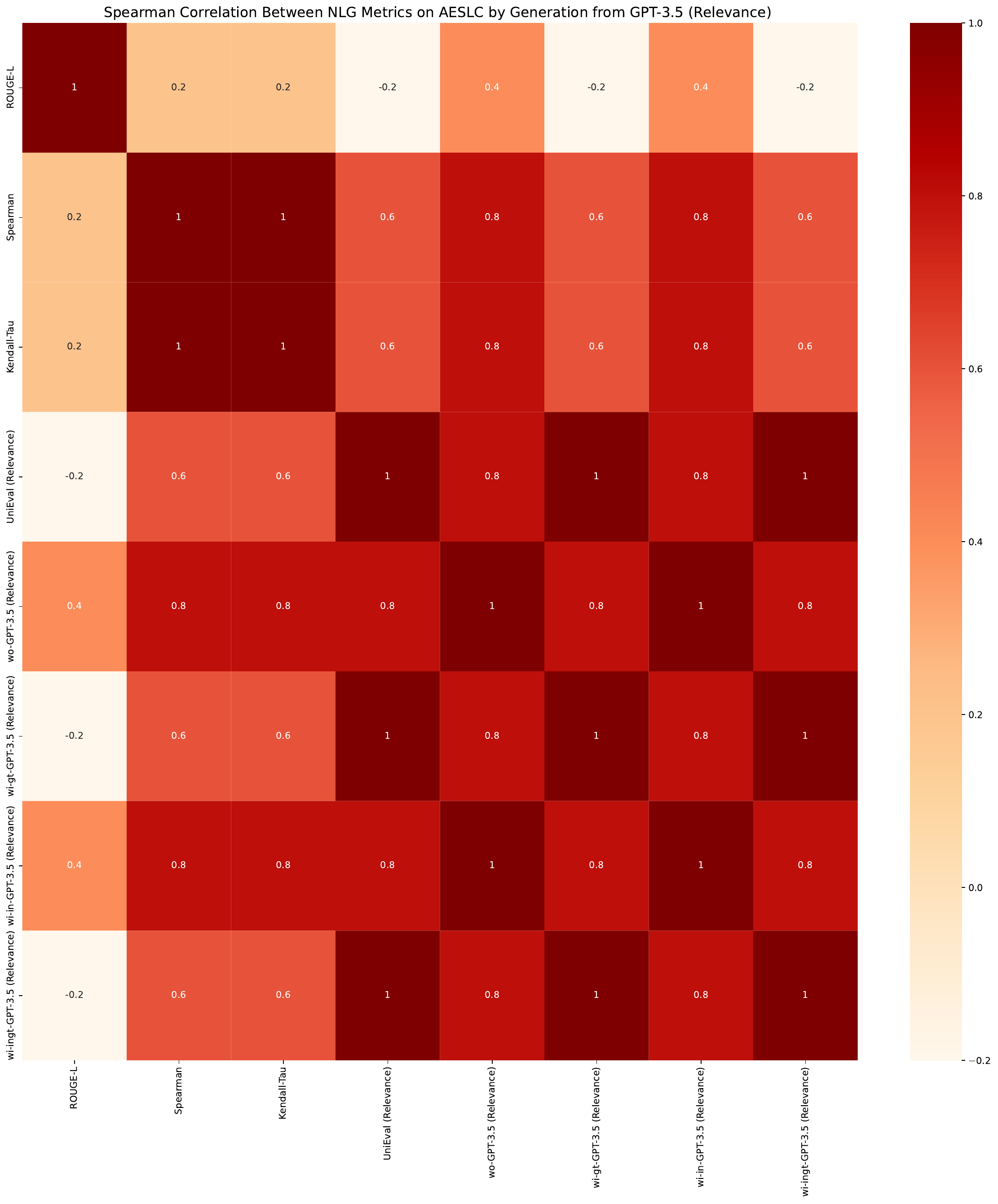}
\caption{Diagram of Spearman correlation in terms of relevance between NLG metrics on AESLC dataset from the view of uncertainty estimation methods used in Fig.~\ref{fig:spear_ue_aes_gpt35}. The generated summaries are from GPT-3.5. For the GPT-3.5-based NLG metrics, we only draw wi-ingt-GPT-3.5 results to save space.}
\label{fig:spear_nlg_aes_relevance_gpt35}
\end{figure*}

\begin{figure*}[!htbp]
\centering
\includegraphics[width=\textwidth]{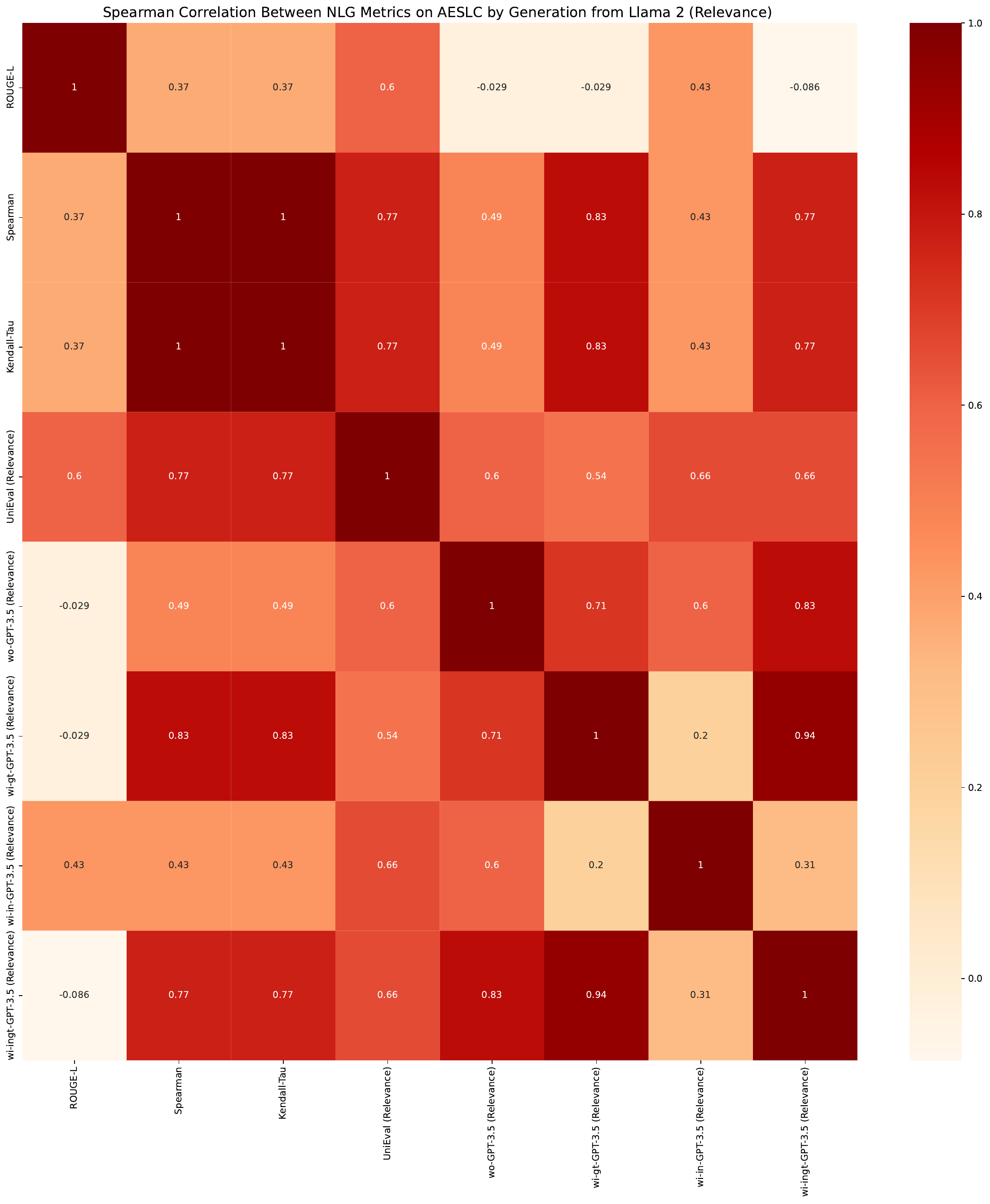}
\caption{Diagram of Spearman correlation in terms of relevance between NLG metrics on AESLC dataset from the view of uncertainty estimation methods used in Fig.~\ref{fig:spear_ue_aes_llama}. The generated summaries are from Llama 2. For the GPT-3.5-based NLG metrics, we only draw wi-ingt-GPT-3.5 results to save space.}
\label{fig:spear_nlg_aes_relevance_llama}
\end{figure*}

\begin{figure*}[!htbp]
\centering
\includegraphics[width=\textwidth]{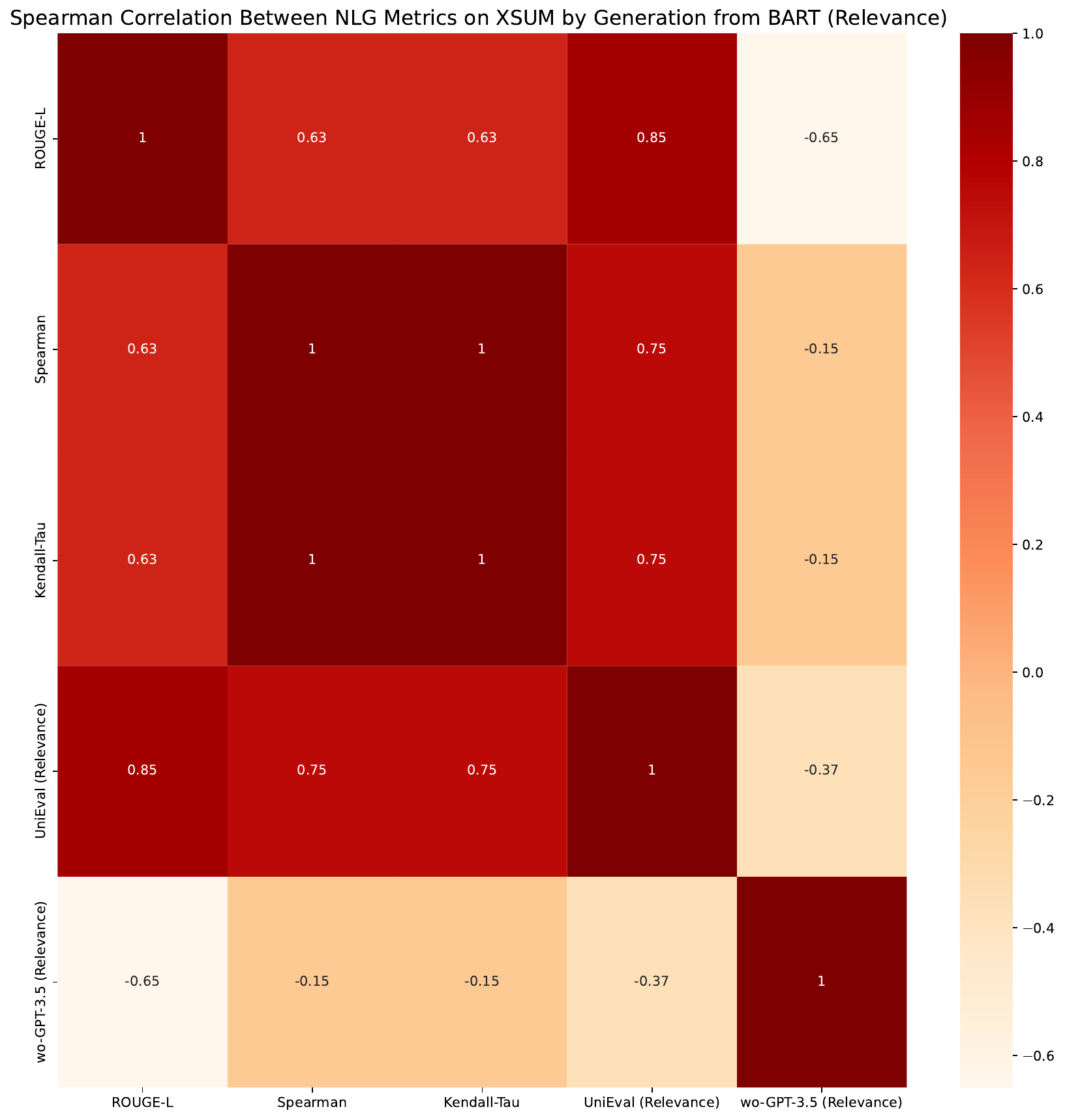}
\caption{Diagram of Spearman correlation in terms of relevance between NLG metrics on XSUM dataset from the view of uncertainty estimation methods used in Fig.~\ref{fig:spear_ue_xsum_bart}.  The generated summaries are from BART. For the GPT-3.5-based NLG metrics, we only conduct wo-GPT-3.5 on the BART generation model setting.}
\label{fig:spear_nlg_xsum_relevance_bart}
\end{figure*}

\begin{figure*}[!htbp]
\centering
\includegraphics[width=\textwidth]{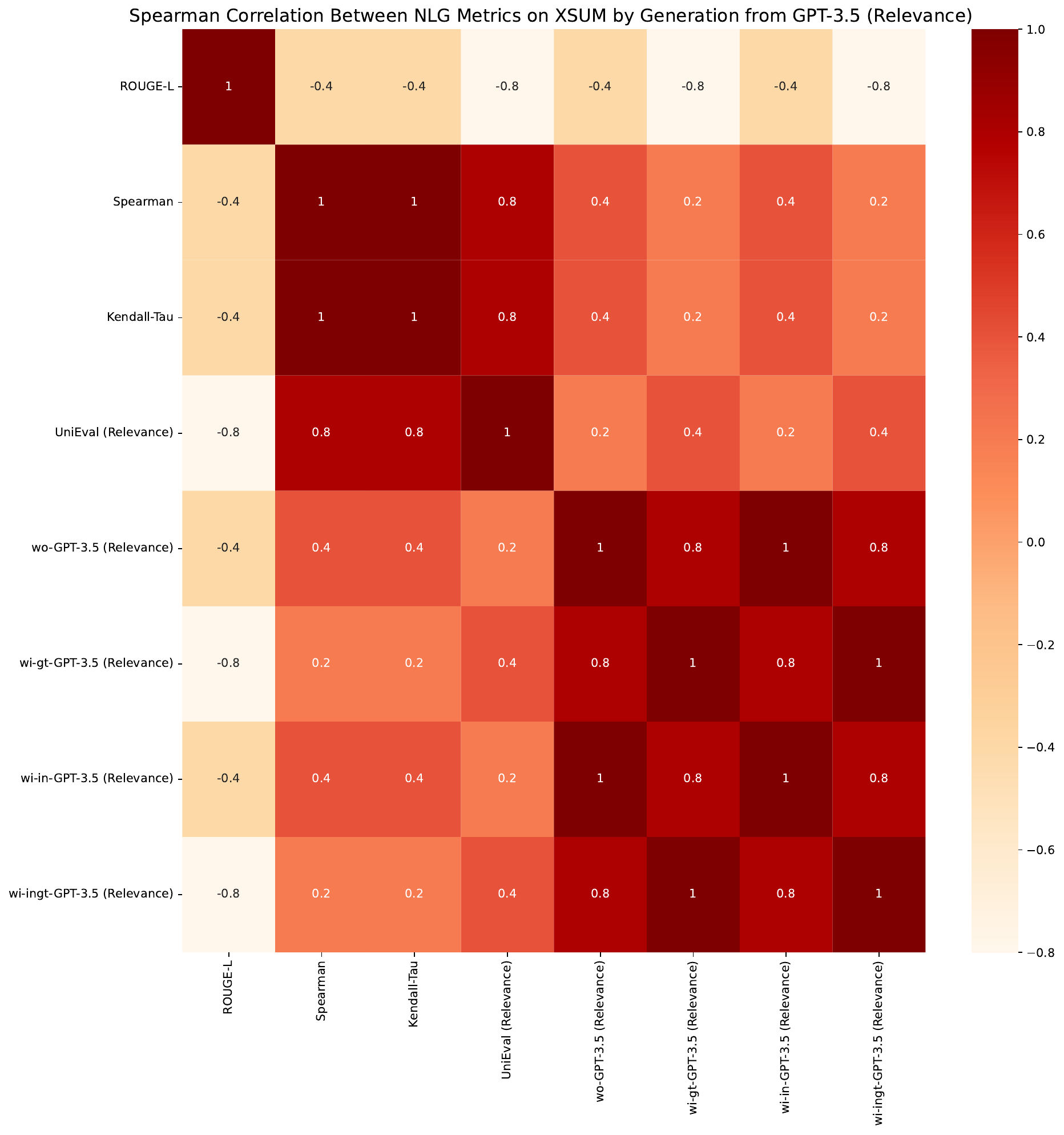}
\caption{Diagram of Spearman correlation in terms of relevance between NLG metrics on XSUM dataset from the view of uncertainty estimation methods used in Fig.~\ref{fig:spear_ue_xsum_gpt35}. The generated summaries are from GPT-3.5. For the GPT-3.5-based NLG metrics, we only draw wi-ingt-GPT-3.5 results to save space.}
\label{fig:spear_nlg_xsum_relevance_gpt35}
\end{figure*}

\begin{figure*}[!htbp]
\centering
\includegraphics[width=\textwidth]{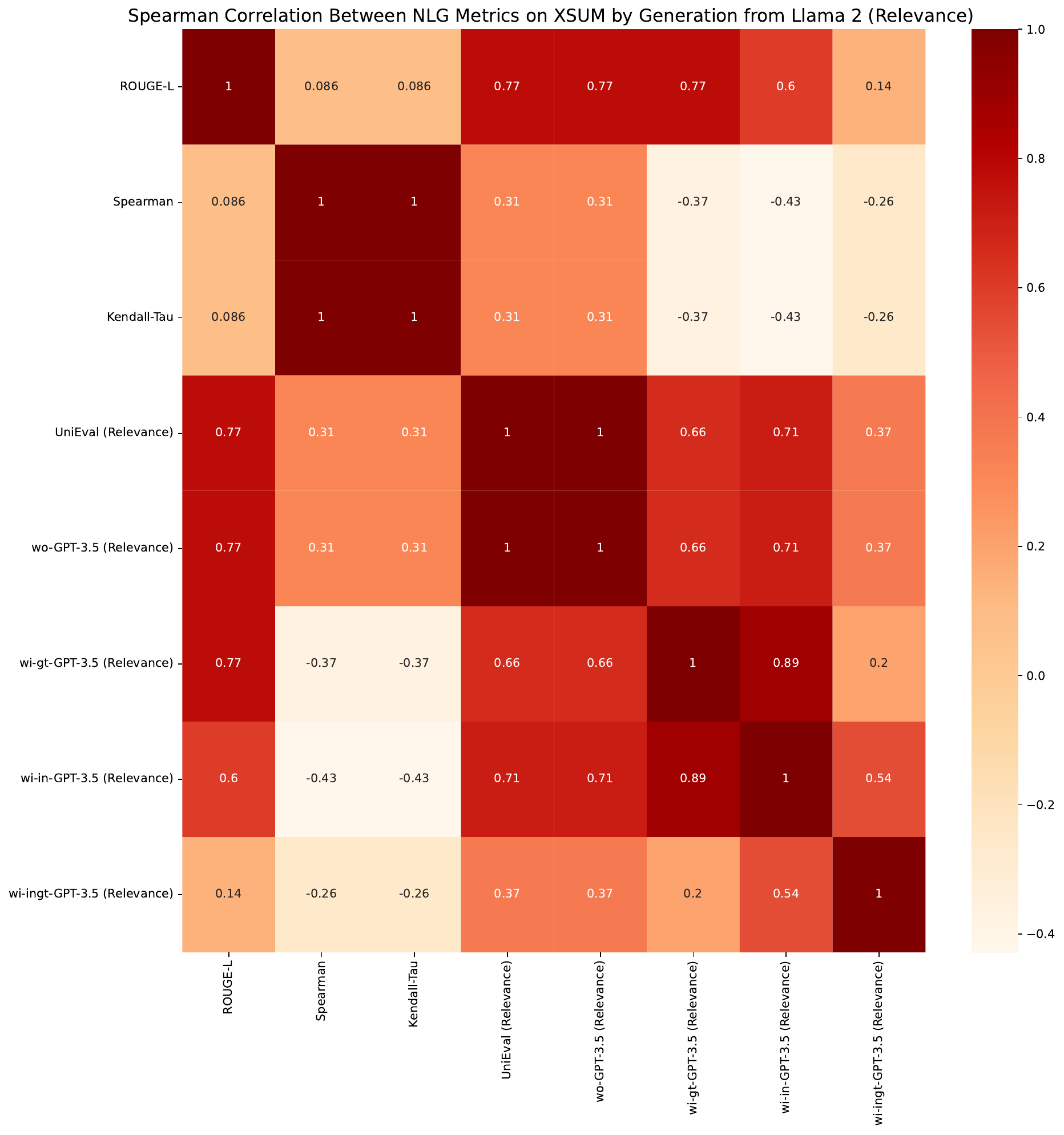}
\caption{Diagram of Spearman correlation in terms of relevance between NLG metrics on XSUM dataset from the view of uncertainty estimation methods used in Fig.~\ref{fig:spear_ue_xsum_llama}. The generated summaries are from Llama 2. For the GPT-3.5-based NLG metrics, we only draw wi-ingt-GPT-3.5 results to save space.}
\label{fig:spear_nlg_xsum_relevance_llama}
\end{figure*}

Figures~\ref{fig:spear_nlg_aes_relevance_bart},~\ref{fig:spear_nlg_aes_relevance_gpt35},~\ref{fig:spear_nlg_aes_relevance_llama},~\ref{fig:spear_nlg_xsum_relevance_bart},~\ref{fig:spear_nlg_xsum_relevance_gpt35}, and~\ref{fig:spear_nlg_xsum_relevance_llama} show the evaluation of UE-TS models from the relevance-dimension NLG metrics.

Based on the BART generation model, Figure~\ref{fig:spear_nlg_aes_relevance_bart} shows strongly positive correlations among ROUGE-L, Spearman, and Kendall-Tau. However, the Unieval (Relevance) and wo-GPT-3.5 (Relevance) show weak correlations with all other NLG metrics in the relevance dimension.

Also, based on the BART generation model, Figure~\ref{fig:spear_nlg_xsum_relevance_bart} shows strongly positive correlations among ROUGE-L, Spearman, Kendall-Tau, and UniEval (Relevance). However, the wo-GPT-3.5 (Relevance) shows weak correlation with all other NLG metrics in the relevance dimension.

Based on the GPT-3.5 generation model, Figure~\ref{fig:spear_nlg_aes_relevance_gpt35} shows strongly positive correlations among Spearman, Kendall-Tau, UniEval (Relevance), and all GPT-3.5-based NLG metrics in the relevance dimension. However, Rouge-L shows weak correlation with all other NLG metrics in the relevance dimension. As for GPT-3.5-based NLG metrics, except for wi-in-GPT-3.5, we see that all other GPT-3.5-based models show identical performance ranks with each other. This implies that the relevance metric using GPT-3.5 is strongly related to the target text source.

Also, based on the GPT-3.5 generation model, Figure~\ref{fig:spear_nlg_xsum_relevance_gpt35} shows a positive correlation among Spearman, Kendall-Tau, UniEval (Relevance), and all GPT-3.5-based NLG metrics in the relevance dimension, except for ROUGE-L.

Based on the Llama-2 generation model, in Figure~\ref{fig:spear_nlg_aes_relevance_llama}, UniEval (Relevance) shows a strongly positive correlation with all other NLG metrics in the relevance dimension. This is because each element in the row of UniEval (Relevance) is greater than 0.5. Similarly, wo-GPT-3.5 (Relevance) also shows a strongly positive correlation with all other NLG metrics in the relevance dimension except for ROUGE-L. Additionally, we found that wi-in-GPT-3.5 shows weak correlation with other GPT-3.5-based NLG metrics. This weak correlation indicates that if we only provide the ground-truth summaries for the relevance evaluation, the relevance metrics may ignore many details.

Also, based on the Llama-2 generation model, Figure~\ref{fig:spear_nlg_xsum_relevance_llama} shows that both UniEval (Relevance) and wo-GPT-3.5 (Relevance) show a positive correlation with other methods. The other metrics do not exhibit consistent positive or negative correlations with each other.

Besides, by comparing wo-GPT-3.5 (relevance) and wi-GPT-3.5 (relevance), where the only difference lies in whether GPT-3.5 is given the concept of relevance or not, their high positive correlation indicates that GPT-3.5 incorporates the concept of relevance. Consequently, the difference in providing the relevance concept to GPT-3.5 is not apparent.

For the evaluation of UE-TS models using relevance-dimension NLG metrics, we have the following conclusions.

\noindent 1. Generation models of the same type across different datasets could result in similar correlations among various methods.

\noindent 2. Spearman and Kendall-Tau usually exhibit positive correlations. Therefore, in future experiments, choosing one of them is sufficient.

\noindent 3. When employing LLMs as generation models, UniEval (Relevance) and wo-GPT-3.5 (Relevance) tend to exhibit positive correlations with most other NLG metrics. Therefore, one of them could serve as a representative NLG metric.

\noindent 4. When utilizing LLMs as a type of relevance NLG metric, the choice of target text source can greatly impact the final conclusion. Specifically, using ground-truth summaries versus using input text as the target text source can result in different performance rankings.

\noindent 5. When using LLMs as a type of relevance NLG metric, if both ground-truth summaries and input text are employed together as the target text source, the ground-truth summaries will dominate the metric results.

\noindent 6. GPT-3.5 knows the concept of relevance. Consequently, the difference in providing the relevance concept to GPT-3.5 is not apparent.

\subsubsection{Analysis on NLG Consistency Dimension}
\label{sec:app_nlg_consistency_dimension}

\begin{figure*}[!htbp]
\centering
\includegraphics[width=\textwidth]{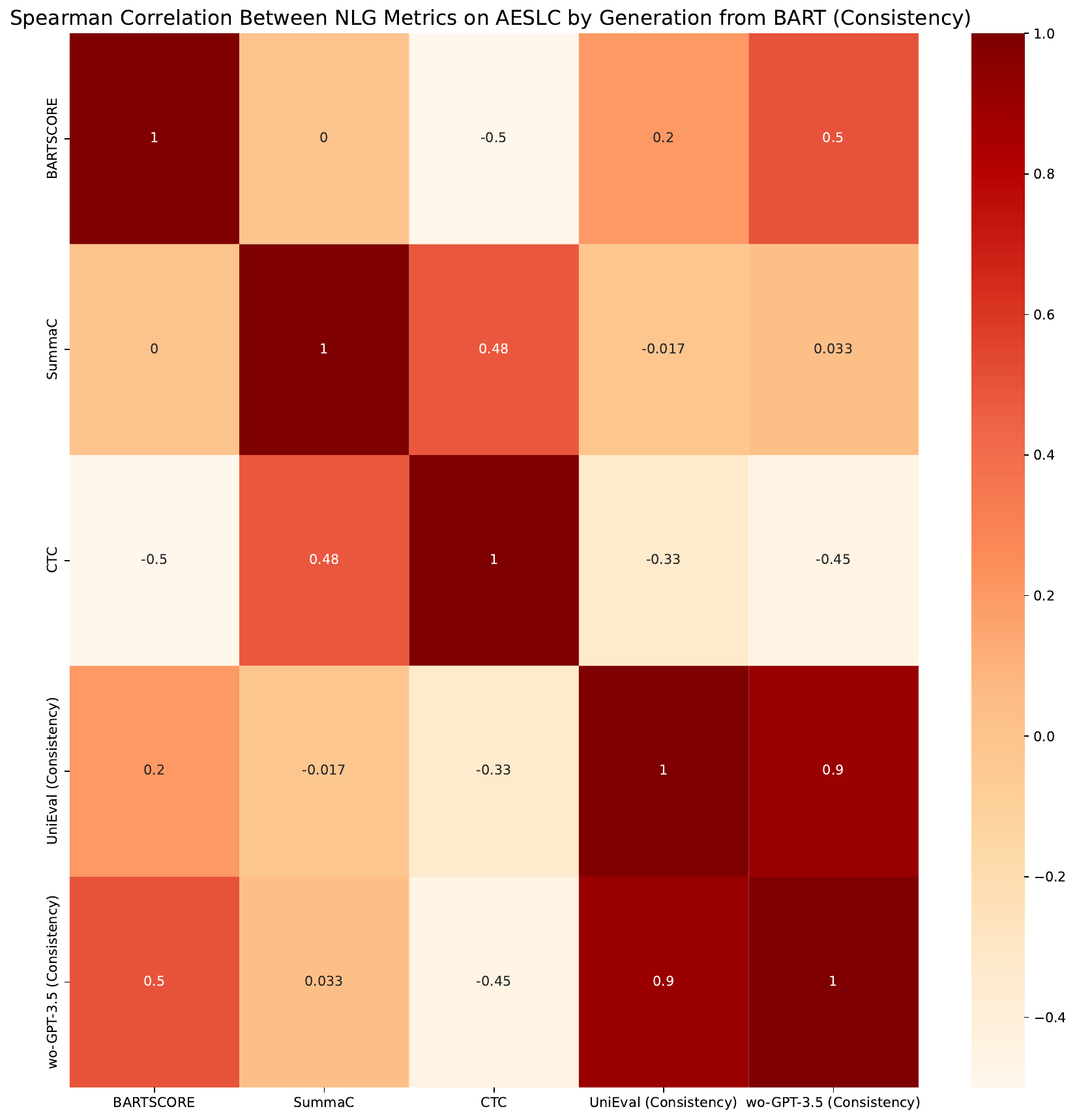}
\caption{Diagram of Spearman correlation in terms of consistency between NLG metrics on AESLC dataset from the view of uncertainty estimation methods used in Fig.~\ref{fig:spear_ue_aes_bart}.  The generated summaries are from BART. For the GPT-3.5-based NLG metrics, we only conduct wo-GPT-3.5 on the BART generation model setting.}
\label{fig:spear_nlg_aes_consistency_bart}
\end{figure*}

\begin{figure*}[!htbp]
\centering
\includegraphics[width=\textwidth]{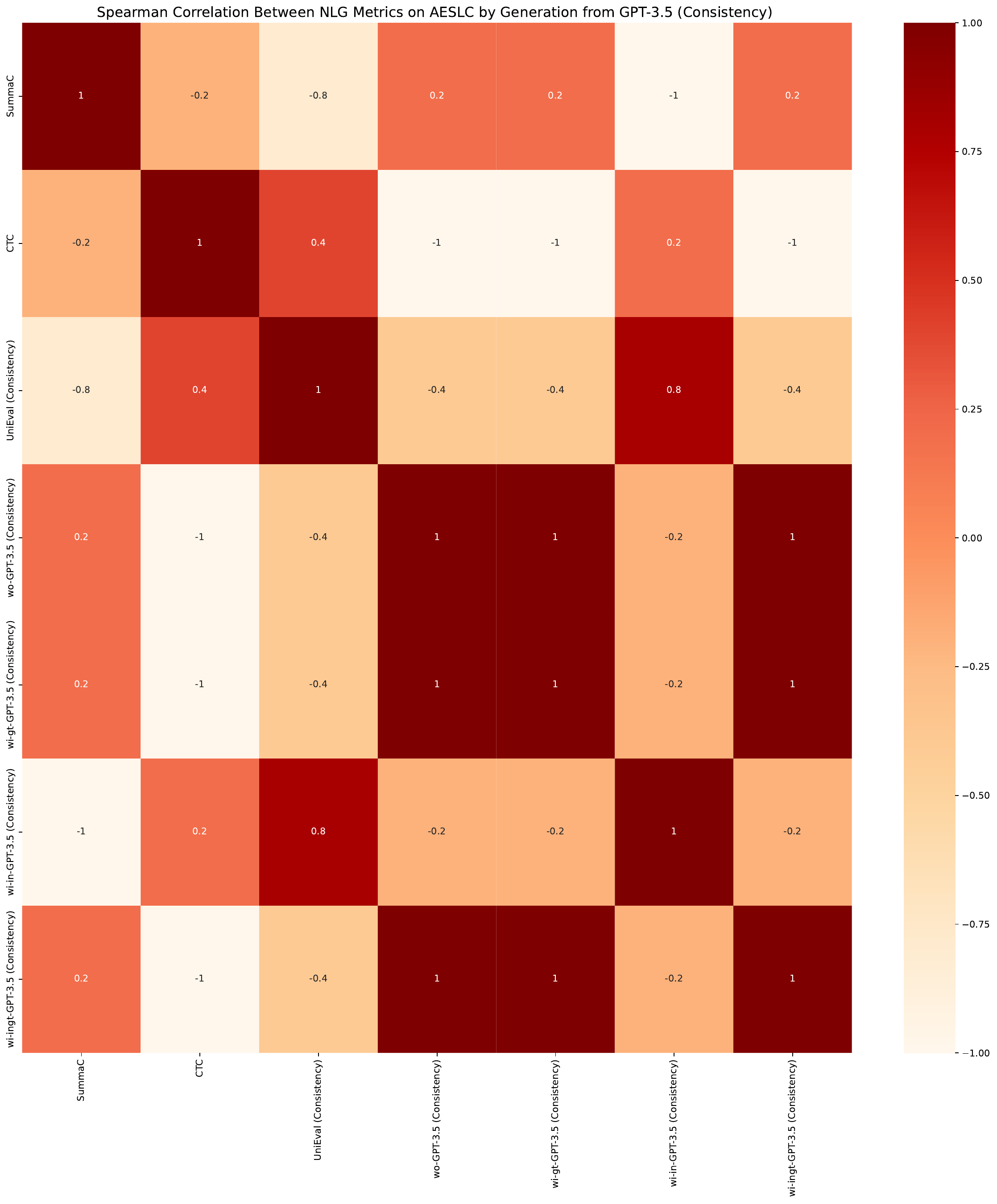}
\caption{Diagram of Spearman correlation in terms of consistency between NLG metrics on AESLC dataset from the view of uncertainty estimation methods used in Fig.~\ref{fig:spear_ue_aes_gpt35}. The generated summaries are from GPT-3.5. For the GPT-3.5-based NLG metrics, we only draw wi-ingt-GPT-3.5 results to save space.}
\label{fig:spear_nlg_aes_consistency_gpt35}
\end{figure*}

\begin{figure*}[!htbp]
\centering
\includegraphics[width=\textwidth]{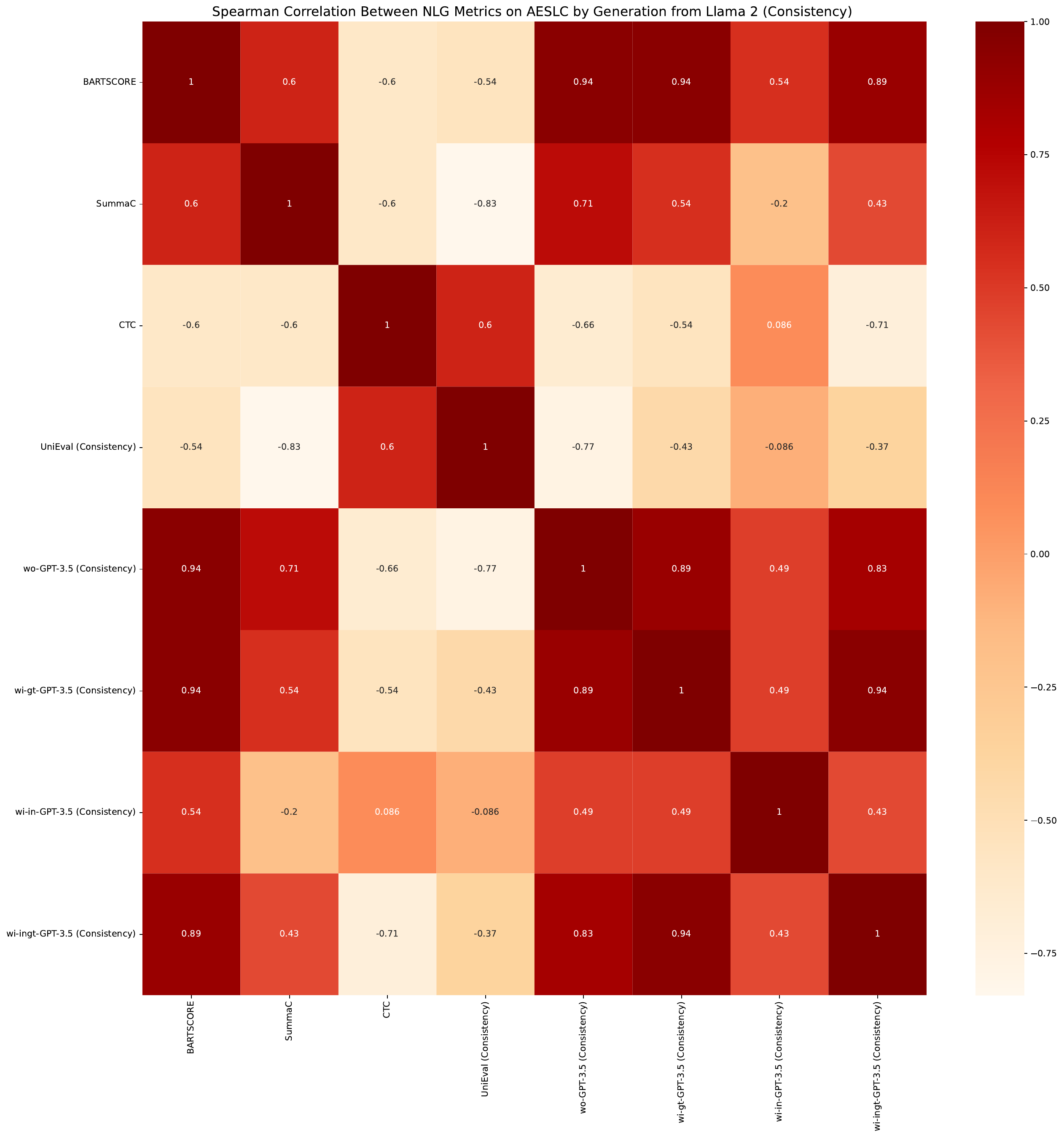}
\caption{Diagram of Spearman correlation in terms of consistency between NLG metrics on AESLC dataset from the view of uncertainty estimation methods used in Fig.~\ref{fig:spear_ue_aes_llama}. The generated summaries are from Llama 2. For the GPT-3.5-based NLG metrics, we only draw wi-ingt-GPT-3.5 results to save space.}
\label{fig:spear_nlg_aes_consistency_llama}
\end{figure*}

\begin{figure*}[!htbp]
\centering
\includegraphics[width=\textwidth]{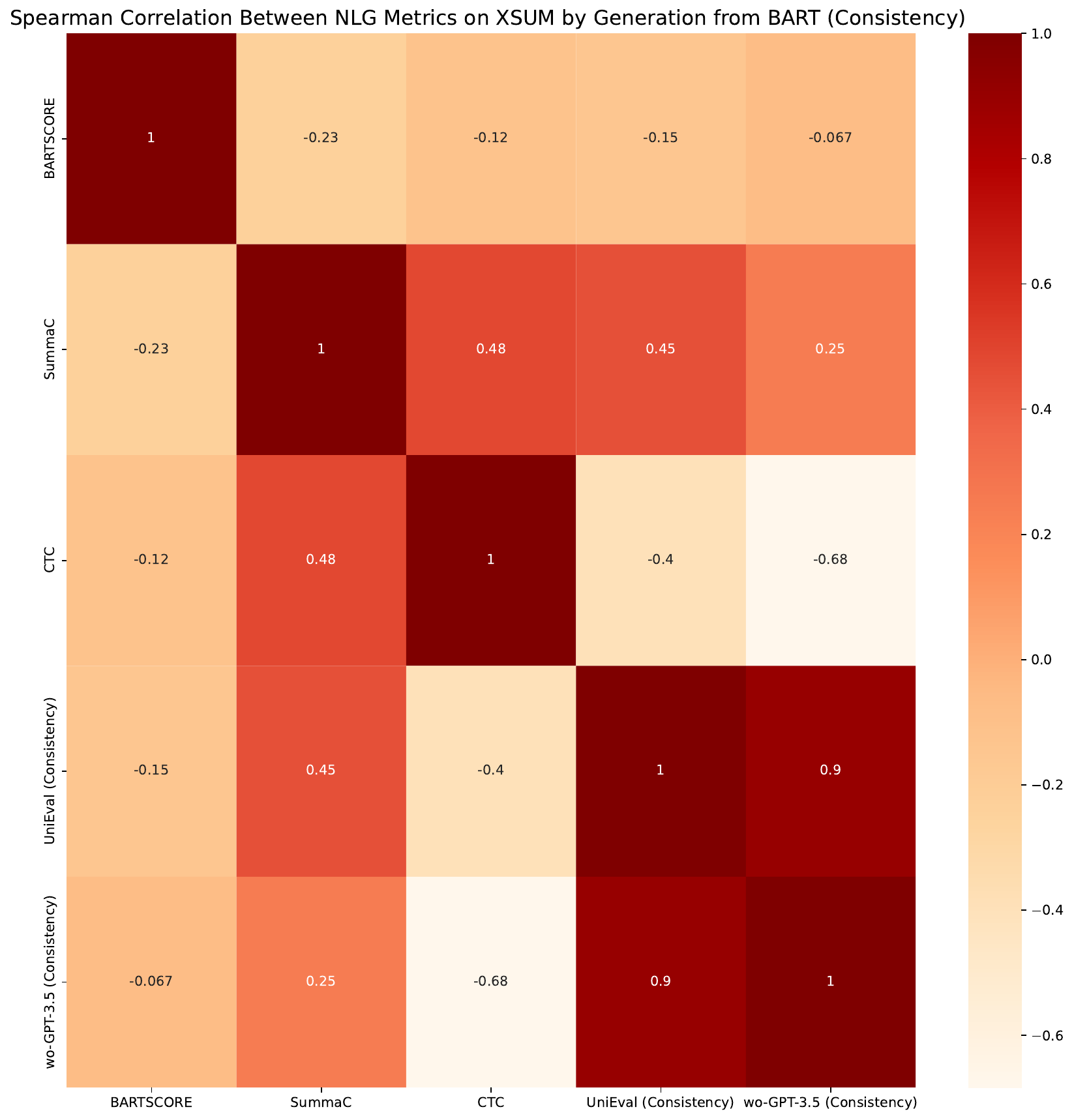}
\caption{Diagram of Spearman correlation in terms of consistency between NLG metrics on XSUM dataset from the view of uncertainty estimation methods used in Fig.~\ref{fig:spear_ue_xsum_bart}.  The generated summaries are from BART. For the GPT-3.5-based NLG metrics, we only conduct wo-GPT-3.5 on the BART generation model setting.}
\label{fig:spear_nlg_xsum_consistency_bart}
\end{figure*}

\begin{figure*}[!htbp]
\centering
\includegraphics[width=\textwidth]{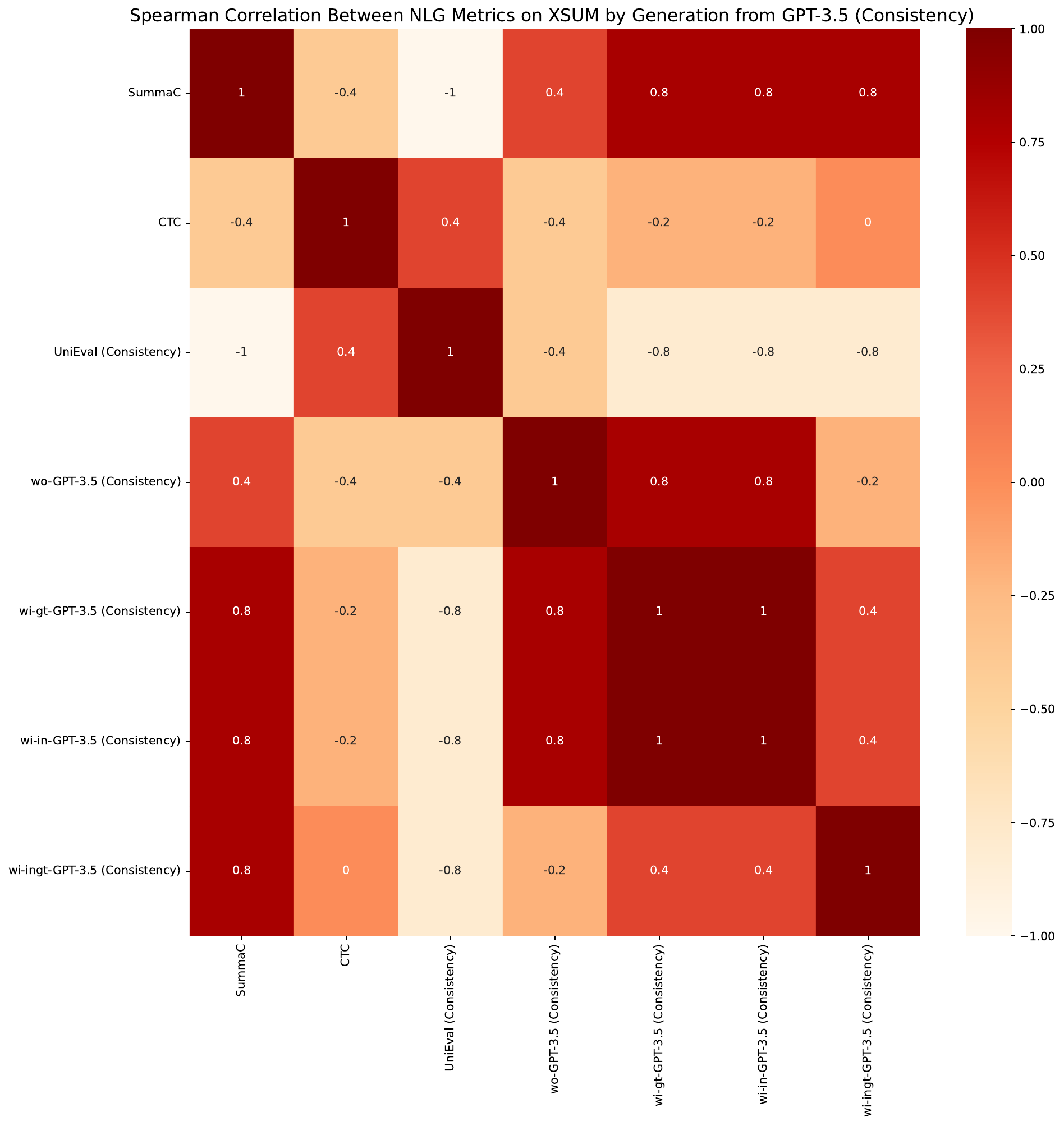}
\caption{Diagram of Spearman correlation in terms of consistency between NLG metrics on XSUM dataset from the view of uncertainty estimation methods used in Fig.~\ref{fig:spear_ue_xsum_gpt35}. The generated summaries are from GPT-3.5. For the GPT-3.5-based NLG metrics, we only draw wi-ingt-GPT-3.5 results to save space.}
\label{fig:spear_nlg_xsum_consistency_gpt35}
\end{figure*}

\begin{figure*}[!htbp]
\centering
\includegraphics[width=\textwidth]{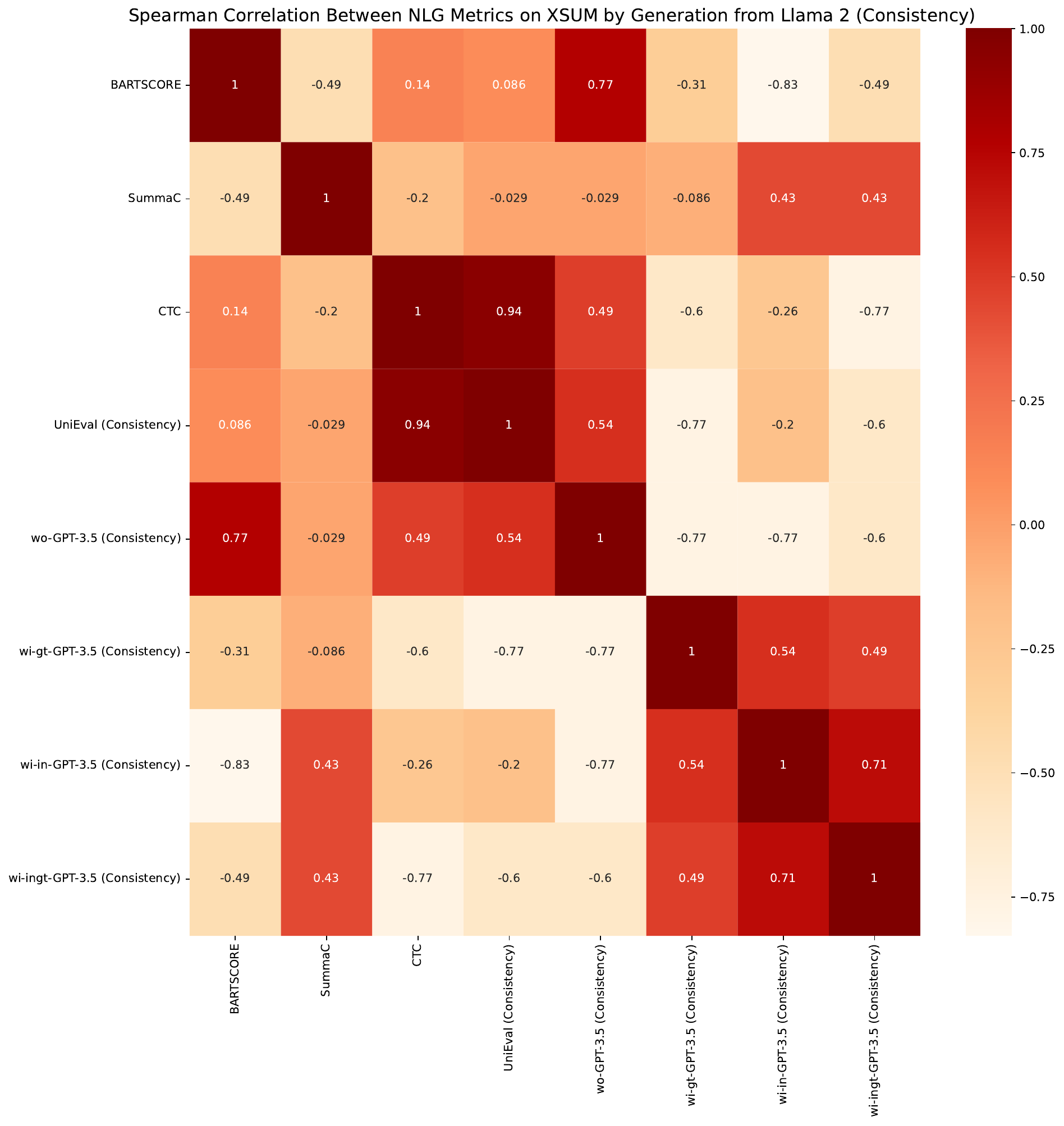}
\caption{Diagram of Spearman correlation in terms of consistency between NLG metrics on XSUM dataset from the view of uncertainty estimation methods used in Fig.~\ref{fig:spear_ue_xsum_llama}. The generated summaries are from Llama 2. For the GPT-3.5-based NLG metrics, we only draw wi-ingt-GPT-3.5 results to save space.}
\label{fig:spear_nlg_xsum_consistency_llama}
\end{figure*}

Figures~\ref{fig:spear_nlg_aes_consistency_bart},~\ref{fig:spear_nlg_aes_consistency_gpt35},~\ref{fig:spear_nlg_aes_consistency_llama},~\ref{fig:spear_nlg_xsum_consistency_bart},~\ref{fig:spear_nlg_xsum_consistency_gpt35}, and~\ref{fig:spear_nlg_xsum_consistency_llama} display the evaluation results of UE-TS models based on the consistency-dimension NLG metrics.

Specifically, based on BART generation, Figure~\ref{fig:spear_nlg_aes_consistency_bart} illustrates that UniEval (Consistency) and wo-GPT-3.5 (Consistency) exhibit a strongly positive correlation. SummaC and CTC show a positive correlation as well. Additionally, wo-GPT-3.5 (Consistency) displays a strongly positive correlation with BARTScore. However, there is no or even a negative correlation between the group of UniEval (Consistency) and wo-GPT-3.5 (Consistency) and the group of SummaC and CTC.

Also, based on BART generation, Figure~\ref{fig:spear_nlg_xsum_consistency_bart} illustrates that UniEval (Consistency) and wo-GPT-3.5 (Consistency) exhibit a strongly positive correlation. SummaC and CTC show a positive correlation as well. However, SummaC displays a positive correlation with UniEval (Consistency) and wo-GPT-3.5 (Consistency), whereas CTC does not.

Based on GPT-3.5 generation, in Figure~\ref{fig:spear_nlg_aes_consistency_gpt35}, positive correlations are found in a group consisting of CTC, UniEval (Consistency), and wi-in-GPT-3.5 (Consistency). Additionally, positive correlations are observed in another group comprising SummaC, wo-GPT-3.5 (Consistency), wi-gt-GPT-3.5 (Consistency), and wi-ingt-GPT-3.5 (Consistency). However, these two groups display negative correlations. Thus, different metrics in the NLG consistency dimension can lead to different evaluation performance rankings of UE-TS models.

Also, based on GPT-3.5 generation, Figure~\ref{fig:spear_nlg_xsum_consistency_gpt35} depicts positive correlations between CTC and UniEval (Consistency). Additional positive correlations are observed in a group comprising SummaC and all GPT-3.5-based NLG metrics. The only exception is that wi-ingt-GPT-3.5 (Consistency) and wo-ingt-GPT-3.5 (Consistency) exhibit negative correlations.

Based on Llama 2 generation, Figure~\ref{fig:spear_nlg_aes_consistency_llama} illustrates positive correlations in a group comprising BARTScore and all GPT-3.5-based NLG metrics. Additionally, positive correlations are found in a group consisting of CTC and UniEval (Consistency). However, negative correlations are observed for both of these groups. Regarding SummaC, it shows positive correlations with most of the first group except for wi-in-GPT-3.5 (Consistency) and negative correlations with the second group.

Also, based on Llama 2 generation, Figure~\ref{fig:spear_nlg_xsum_consistency_gpt35} shows strong positive correlations in a group comprising CTC, UniEval (Consistency), and wo-GPT-3.5 (Consistency). Another set of strong positive correlations is found in a group including wi-gt-GPT-3.5, wi-in-GPT-3.5, and wi-ingt-GPT-3.5 (Consistency). However, these two groups exhibit negative correlations with each other. BARTSCORE and SummaC display negative correlations.

According to the above analysis, we can draw the following conclusions regarding the evaluation of UE-TS models using consistency-dimension NLG metrics.

\noindent 1. Different metrics within the NLG consistency dimension can result in varying evaluation performance rankings of UE-TS models.

\noindent 2. CTC tends to exhibit a positive correlation with SummaC or UniEval (Consistency). Thus, when faced with scenarios where we must choose between CTC, SummaC, or UniEval, we can opt for CTC due to its positive correlations with the other two in most cases.

\noindent 3. For GPT-3.5-based NLG metrics, differences in the target text source could lead to discrepancies in some cases. Using ground-truth summaries as the target text source will have a more dominated impact compared to using input text as the target text source.

\noindent 4. For GPT-3.5-based NLG metrics, GPT-3.5 might not fully comprehend the concept of consistency, as indicated by the negative correlation between wo-GPT-3.5 and wi-GPT-3.5 in Figure~\ref{fig:spear_nlg_xsum_coherence_llama}. However, the impact of understanding the concept of consistency is not as pronounced as the impact of using different target text sources.

\noindent 5. For GPT-3.5-based NLG metrics, it is recommended to use wi-in-GPT-3.5 (Consistency) along with one of wo-gt-GPT-3.5, wi-gt-GPT-3.5 (Consistency), or wi-ingt-GPT-3.5.

\subsubsection{Analysis on NLG Coherence Dimension}
\label{sec:app_nlg_coherence_dimension}

\begin{figure*}[!htbp]
\centering
\includegraphics[width=\textwidth]{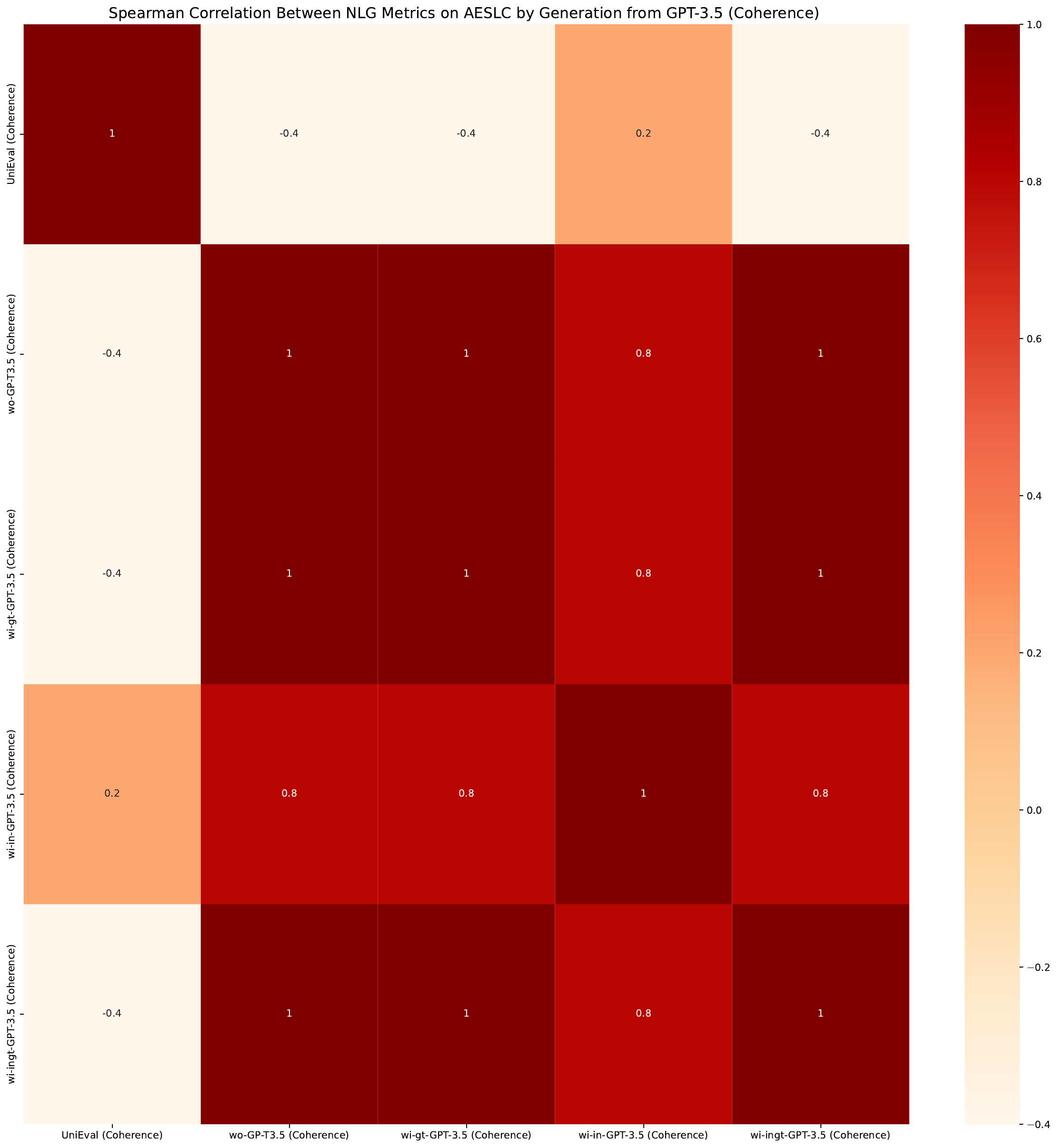}
\caption{Diagram of Spearman correlation in terms of coherence between NLG metrics on AESLC dataset from the view of uncertainty estimation methods used in Fig.~\ref{fig:spear_ue_aes_gpt35}. The generated summaries are from GPT-3.5. For the GPT-3.5-based NLG metrics, we only draw wi-ingt-GPT-3.5 results to save space.}
\label{fig:spear_nlg_aes_coherence_gpt35}
\end{figure*}

\begin{figure*}[!htbp]
\centering
\includegraphics[width=\textwidth]{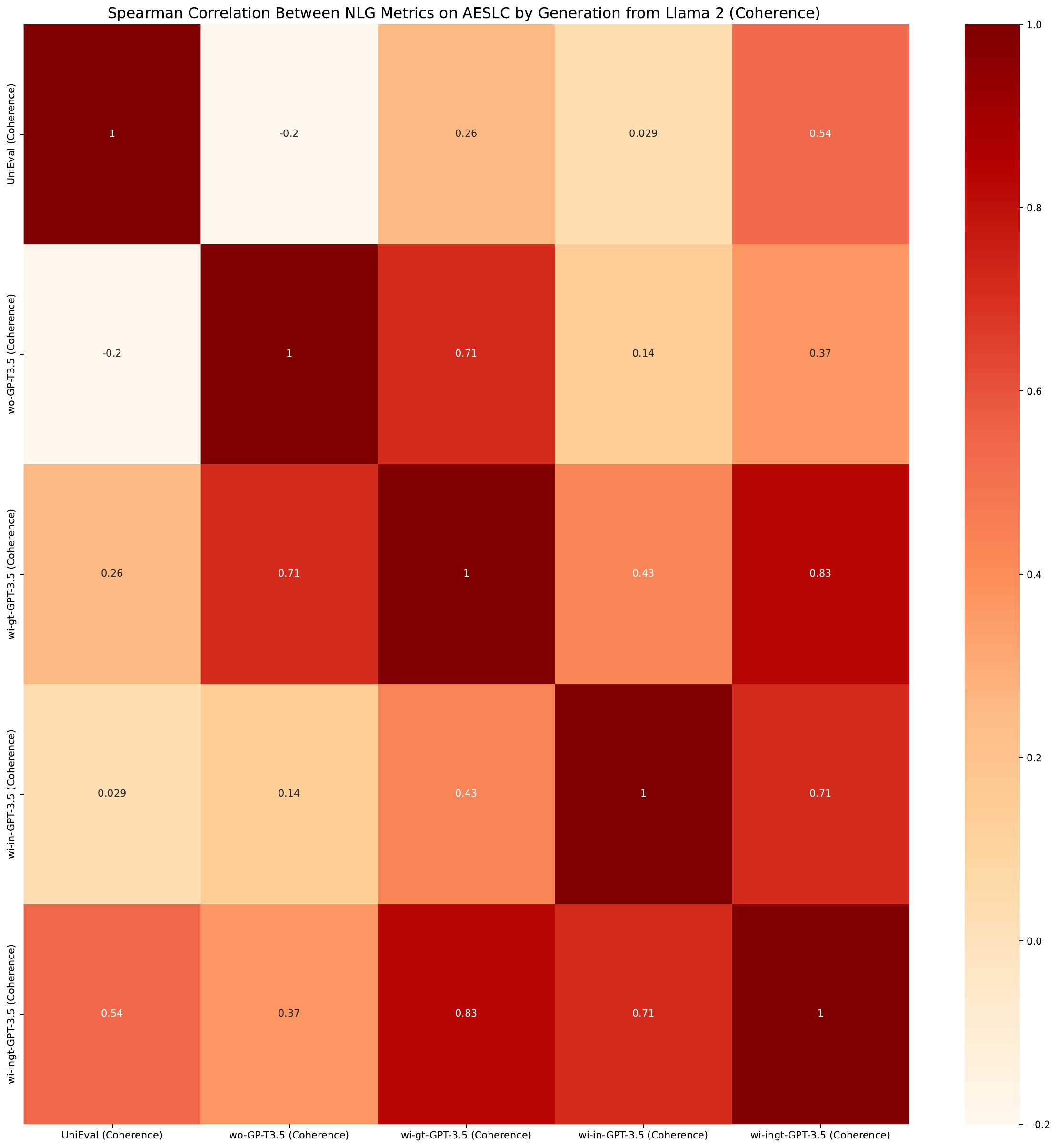}
\caption{Diagram of Spearman correlation in terms of coherence between NLG metrics on AESLC dataset from the view of uncertainty estimation methods used in Fig.~\ref{fig:spear_ue_aes_llama}. The generated summaries are from Llama 2. For the GPT-3.5-based NLG metrics, we only draw wi-ingt-GPT-3.5 results to save space.}
\label{fig:spear_nlg_aes_coherence_llama}
\end{figure*}

\begin{figure*}[!htbp]
\centering
\includegraphics[width=\textwidth]{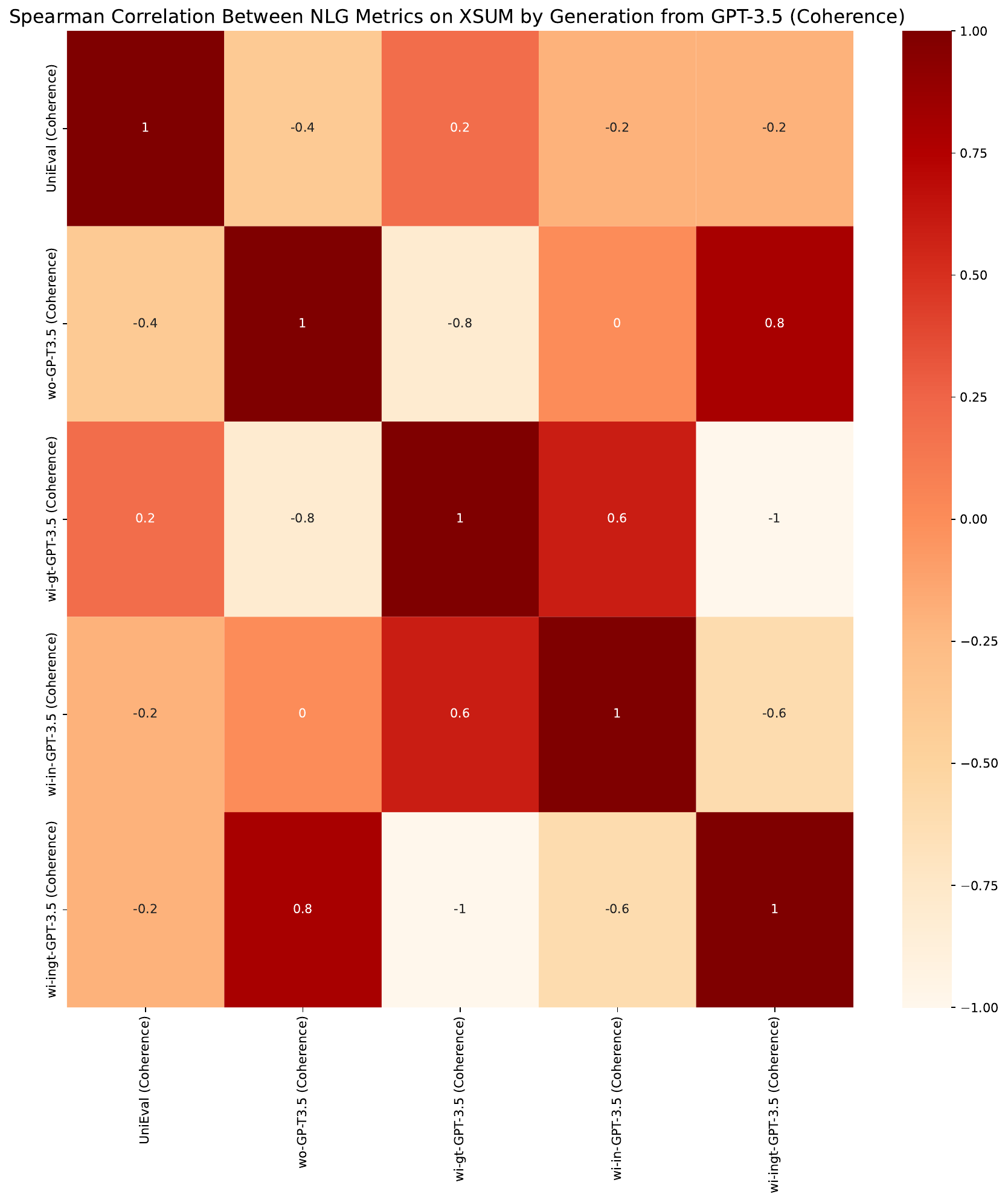}
\caption{Diagram of Spearman correlation in terms of coherence between NLG metrics on XSUM dataset from the view of uncertainty estimation methods used in Fig.~\ref{fig:spear_ue_xsum_gpt35}. The generated summaries are from GPT-3.5. For the GPT-3.5-based NLG metrics, we only draw wi-ingt-GPT-3.5 results to save space.}
\label{fig:spear_nlg_xsum_coherence_gpt35}
\end{figure*}

\begin{figure*}[!htbp]
\centering
\includegraphics[width=\textwidth]{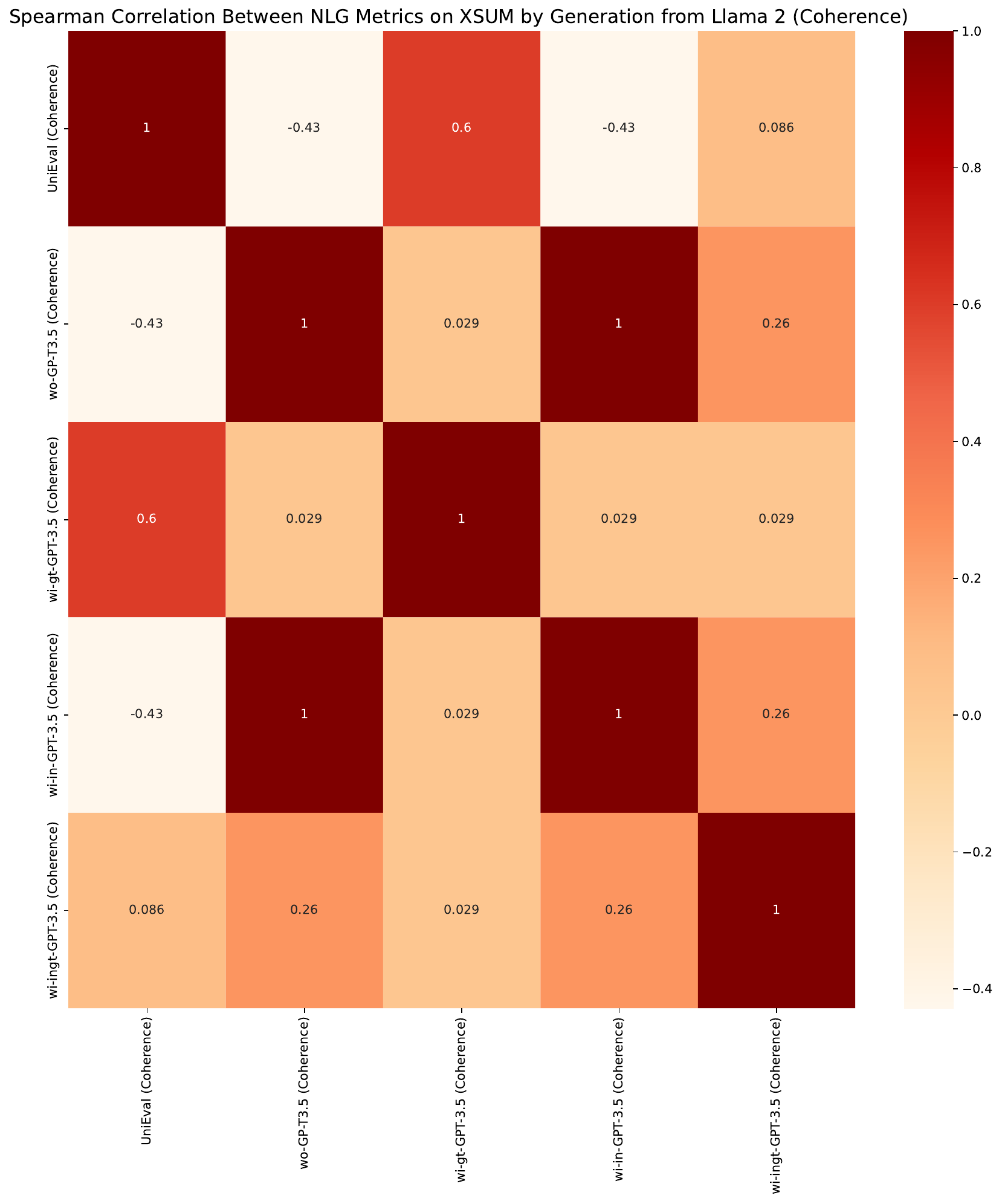}
\caption{Diagram of Spearman correlation in terms of coherence between NLG metrics on XSUM dataset from the view of uncertainty estimation methods used in Fig.~\ref{fig:spear_ue_xsum_llama}. The generated summaries are from Llama 2. For the GPT-3.5-based NLG metrics, we only draw wi-ingt-GPT-3.5 results to save space.}
\label{fig:spear_nlg_xsum_coherence_llama}
\end{figure*}

Figures~\ref{fig:spear_nlg_aes_coherence_gpt35},~\ref{fig:spear_nlg_aes_coherence_llama},~\ref{fig:spear_nlg_xsum_coherence_gpt35} and~\ref{fig:spear_nlg_xsum_coherence_llama} display the evaluation results of UE-TS models based on coherence-dimension NLG metrics. We do not illustrate correlations for BART generation models in the coherence dimension. This is because we only utilize UniEval (coherence) and wo-GPT-3.5 (coherence) for BART generations, and these two metrics may not adequately represent the correlation.

Based on GPT-3.5 generation, Figure~\ref{fig:spear_nlg_aes_coherence_gpt35} indicates that all GPT-3.5-based NLG metrics exhibit a strongly positive correlation with each other in terms of evaluating UE-TS models. However, UniEval (Coherence) demonstrates weak or even negative correlation with these GPT-3.5-based NLG metrics.

Also, based on GPT-3.5, Figure~\ref{fig:spear_nlg_xsum_coherence_gpt35} demonstrates that a group comprising wo-GPT-3.5 (Coherence) and wi-ingt-GPT-3.5 (Coherence) exhibits a strongly positive correlation. Another strong positive correlation is observed in a group consisting of wi-gt-GPT-3.5 (Coherence) and wi-in-GPT-3.5 (Coherence). However, these two groups show no or even negative correlation with each other. Additionally, UniEval (Coherence) demonstrates weak or even negative correlation with these GPT-3.5-based NLG metrics.

Based on Llama 2 generation, Figure~\ref{fig:spear_nlg_aes_coherence_llama} illustrates that all GPT-3.5-based NLG metrics display positive correlations. Among them, wi-gt-GPT-3.5 shows a strongly positive correlation with other GPT-3.5-based NLG metrics except for wi-in-GPT-3.5 (Coherence). However, UniEval (Coherence) demonstrates weak or even negative correlation with these GPT-3.5-based NLG metrics.

Also, based on Llama 2 generation, Figure~\ref{fig:spear_nlg_xsum_coherence_llama} indicates that all GPT-3.5-based NLG metrics exhibit positive correlations. wo-GPT-3.5 (Coherence) and wi-in-GPT-3.5 (Coherence) demonstrate a strongly positive correlation of 1. However, wi-gt-GPT-3.5 shows a relatively weak positive correlation with the other metrics. Regarding UniEval (Coherence), it displays a strongly positive correlation with wi-gt-GPT-3.5 (Coherence) but shows no or even negative correlation with the other three GPT-3.5-based NLG metrics.

According to the above analysis, we can obtain the following conclusions regarding the evaluation of UE-TS models using coherence-dimension NLG metrics:

\noindent 1. It is difficult to determine which GPT-3.5-based metric is better in the coherence dimension. However, based on the strongly positive correlation, either wi-gt-GPT-3.5 or wi-in-GPT-3.5 could be a good choice.

\noindent 2. UniEval (Coherence) exhibits weak or negative correlation with most of the GPT-3.5-based metrics. Therefore, UniEval (Coherence) could serve as a supplement to either wi-gt-GPT-3.5 or wi-in-GPT-3.5.

\noindent 3. Based on Figures~\ref{fig:spear_nlg_xsum_coherence_gpt35} and~\ref{fig:spear_nlg_xsum_coherence_llama} on XSUM datasets, GPT-3.5 exhibits significant divergence between wo-GPT-3.5 and wi-gt-GPT-3.5. This suggests that GPT-3.5 itself might not fully grasp the concept of coherence, and providing this concept could improve evaluation.

\subsubsection{Analysis on NLG Fluency Dimension}
\label{sec:app_nlg_fluency_dimension}

\begin{figure*}[!htbp]
\centering
\includegraphics[width=\textwidth]{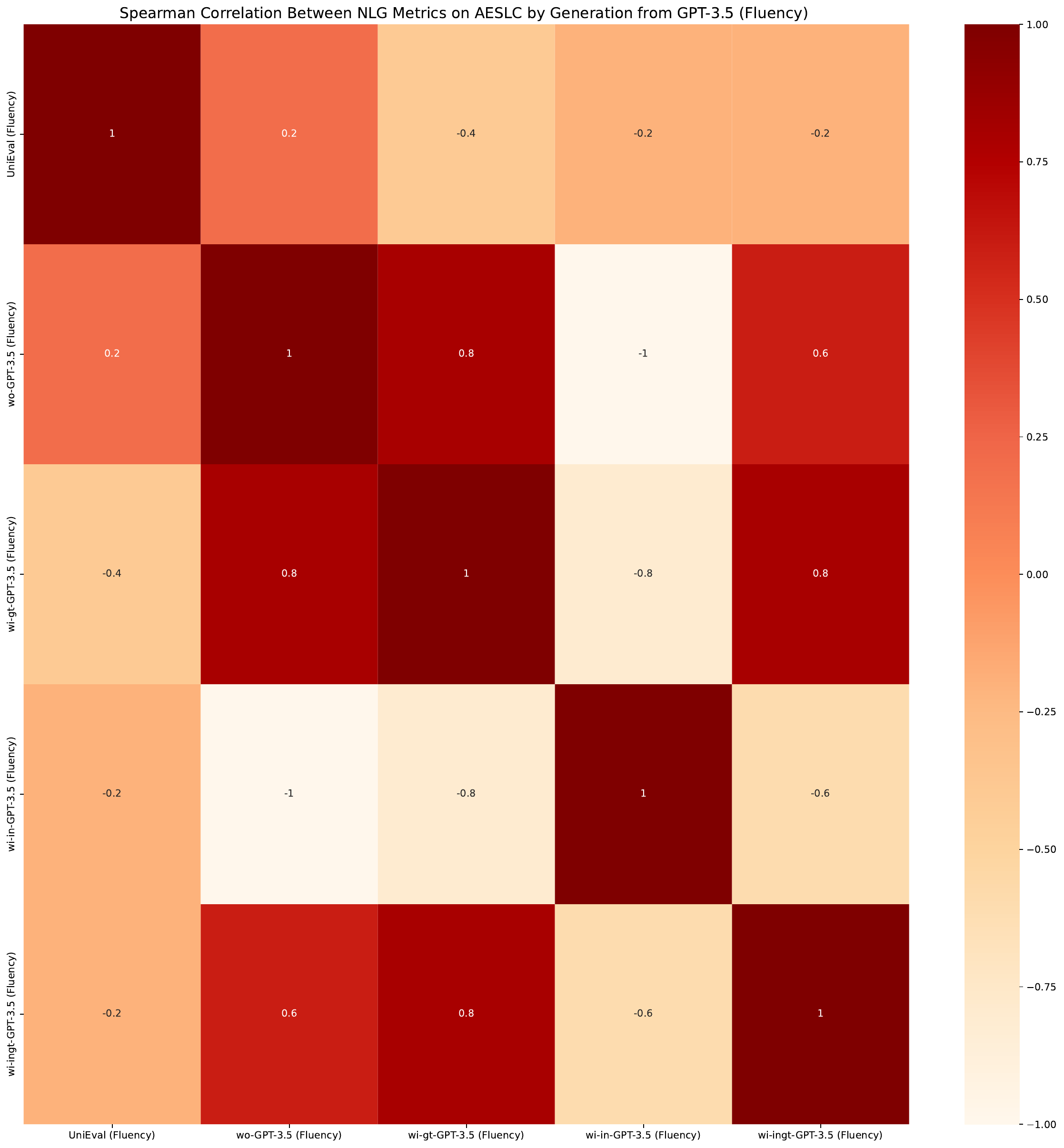}
\caption{Diagram of Spearman correlation in terms of fluency between NLG metrics on AESLC dataset from the view of uncertainty estimation methods used in Fig.~\ref{fig:spear_ue_aes_gpt35}. The generated summaries are from GPT-3.5. For the GPT-3.5-based NLG metrics, we only draw wi-ingt-GPT-3.5 results to save space.}
\label{fig:spear_nlg_aes_fluency_gpt35}
\end{figure*}

\begin{figure*}[!htbp]
\centering
\includegraphics[width=\textwidth]{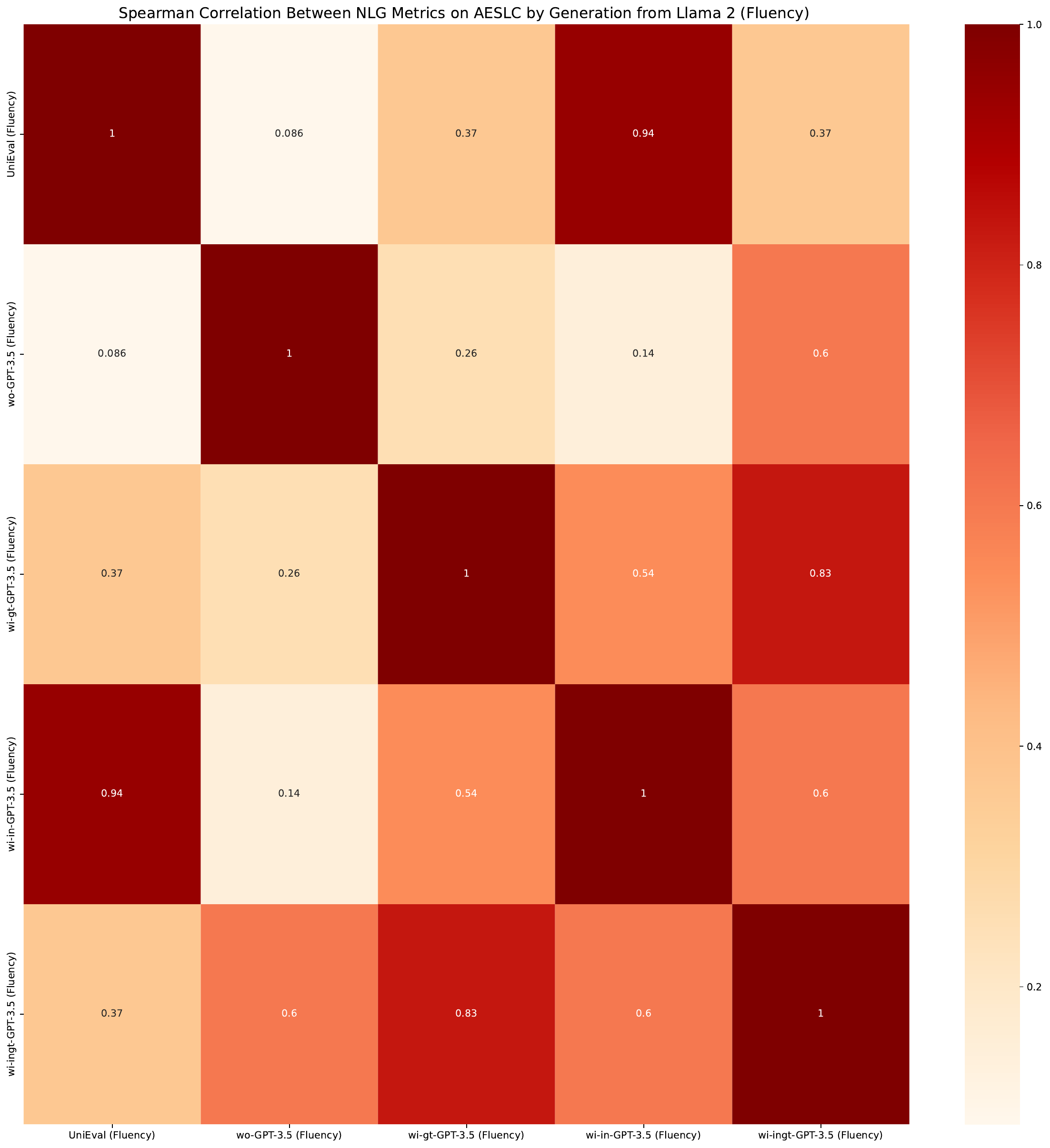}
\caption{Diagram of Spearman correlation in terms of fluency between NLG metrics on AESLC dataset from the view of uncertainty estimation methods used in Fig.~\ref{fig:spear_ue_aes_llama}. The generated summaries are from Llama 2. For the GPT-3.5-based NLG metrics, we only draw wi-ingt-GPT-3.5 results to save space.}
\label{fig:spear_nlg_aes_fluency_llama}
\end{figure*}

\begin{figure*}[!htbp]
\centering
\includegraphics[width=\textwidth]{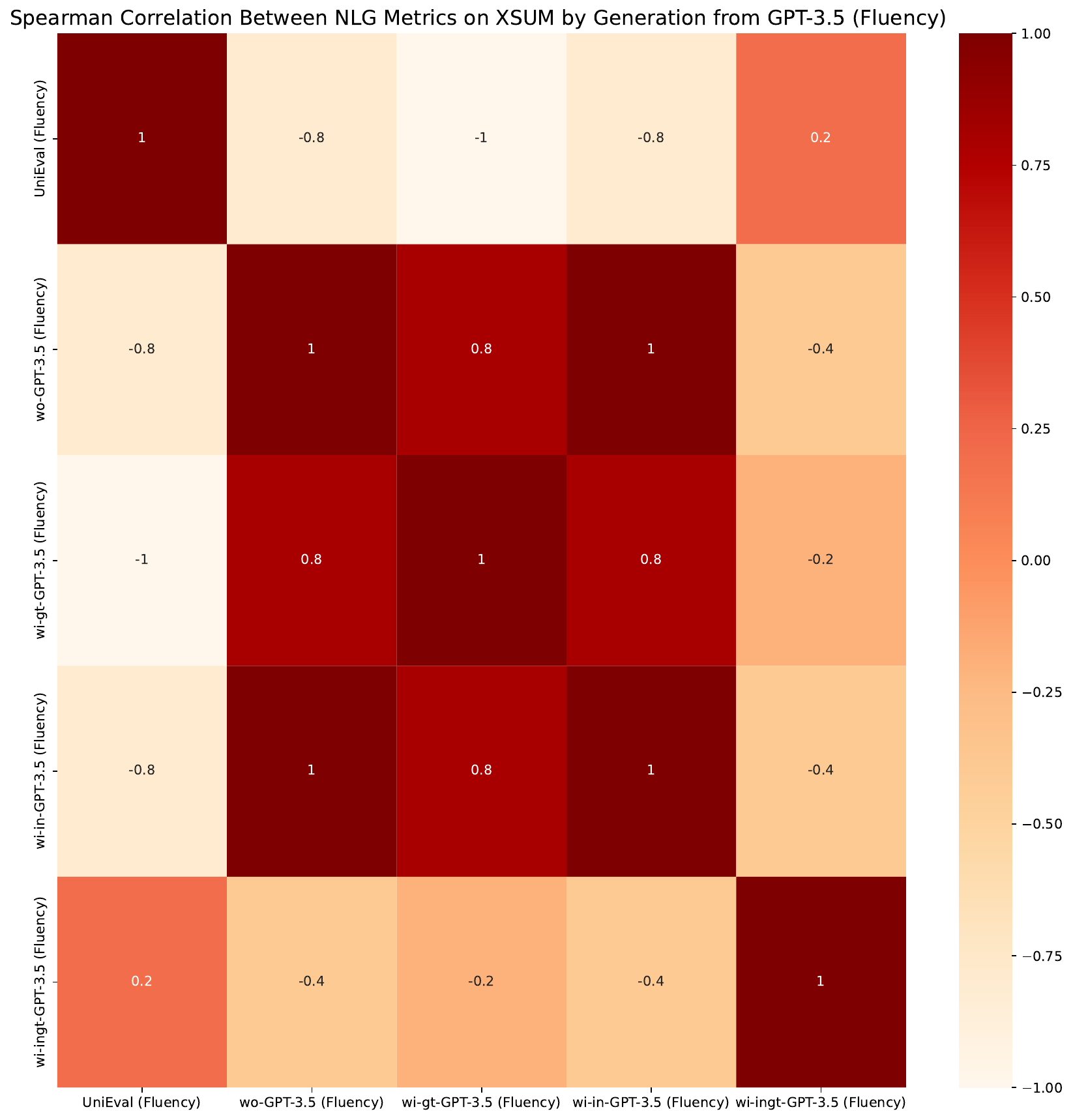}
\caption{Diagram of Spearman correlation in terms of fluency between NLG metrics on XSUM dataset from the view of uncertainty estimation methods used in Fig.~\ref{fig:spear_ue_xsum_gpt35}. The generated summaries are from GPT-3.5. For the GPT-3.5-based NLG metrics, we only draw wi-ingt-GPT-3.5 results to save space.}
\label{fig:spear_nlg_xsum_fluency_gpt35}
\end{figure*}

\begin{figure*}[!htbp]
\centering
\includegraphics[width=\textwidth]{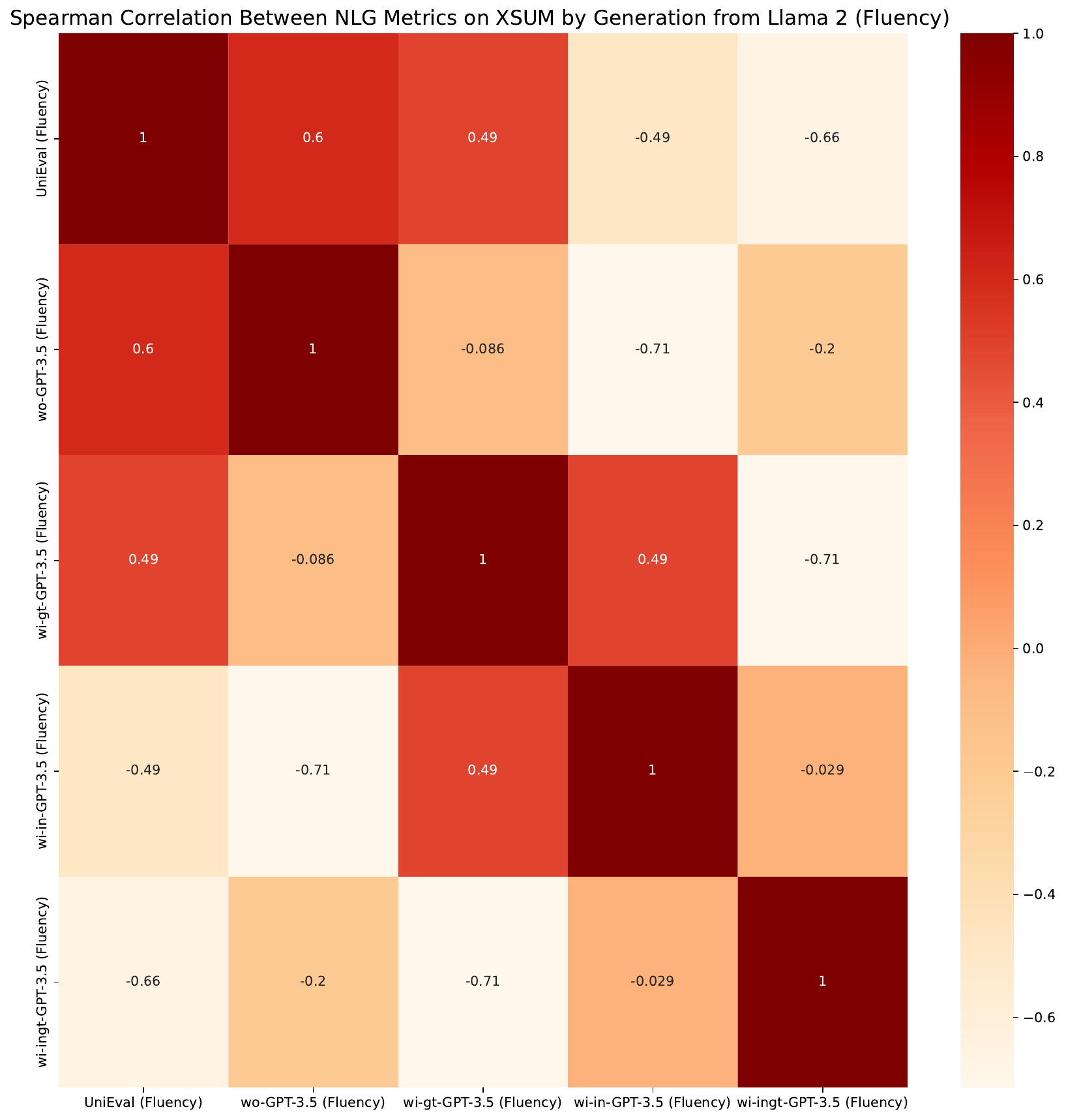}
\caption{Diagram of Spearman correlation in terms of fluency between NLG metrics on XSUM dataset from the view of uncertainty estimation methods used in Fig.~\ref{fig:spear_ue_xsum_llama}. The generated summaries are from Llama 2. For the GPT-3.5-based NLG metrics, we only draw wi-ingt-GPT-3.5 results to save space.}
\label{fig:spear_nlg_xsum_fluency_llama}
\end{figure*}

Figures~\ref{fig:spear_nlg_aes_fluency_gpt35},~\ref{fig:spear_nlg_aes_fluency_llama},~\ref{fig:spear_nlg_xsum_fluency_gpt35}, and~\ref{fig:spear_nlg_xsum_fluency_llama} display the evaluation results of UE-TS models based on fluency-dimension NLG metrics. We do not illustrate correlations for BART generation models in the fluency dimension. This is because we only utilize UniEval (fluency) and wo-GPT-3.5 (fluency) for BART generations, and these two metrics may not adequately represent correlation.

Using GPT-3.5 as a generation model, Figure~\ref{fig:spear_nlg_aes_fluency_gpt35} indicates that wi-in-GPT-3.5 (Fluency) has a negative correlation with wo-GPT-3.5, wi-gt-GPT-3.5, and wi-ingt-GPT-3.5 (Fluency). UniEval (Fluency) demonstrates weak or negative correlation with these GPT-3.5-based NLG metrics.

Also, using GPT-3.5 for generation, Figure~\ref{fig:spear_nlg_xsum_fluency_gpt35} illustrates that all the GPT-3.5-based NLG metrics except wi-ingt-GPT-3.5 (Fluency) exhibit strongly positive correlations. In contrast, these three GPT-3.5-based NLG metrics show negative correlation to UniEval (Fluency) and wi-ingt-GPT-3.5 (Fluency).

Using Llama 2 as a generation model, Figure~\ref{fig:spear_nlg_aes_fluency_llama} demonstrates that all five metrics have positive correlations. Among them, wi-gt-GPT-3.5, wi-in-GPT-3.5, and wi-ingt-GPT-3.5 (Fluency) exhibit strongly positive correlations.

Figure~\ref{fig:spear_nlg_xsum_fluency_llama} illustrates that UniEval (Fluency) has strongly positive correlations with wo-GPT-3.5 (Fluency) and wi-gt-GPT-3.5 (Fluency). However, it exhibits strongly negative correlation with wi-in-GPT-3.5 (Fluency) and wi-ingt-GPT-3.5 (Fluency). Among the GPT-3.5-based NLG metrics, only wi-gt-GPT-3.5 (Fluency) and wi-in-GPT-3.5 (Fluency) show positive correlations, while the other correlations are weak or negative.

Based on the above findings, we can conclude the following observations regarding the fluency dimension.

\noindent 1. It is challenging to determine which NLG metrics consistently exhibit positive correlations with others. However, based on the positive correlations, we recommend wi-gt-GPT-3.5 (Fluency) and wi-ingt-GPT-3.5 (Fluency) due to their correlation patterns.

\noindent 2. Because of the negative correlations between UniEval (Fluency) and GPT-3.5-based NLG metrics, we also recommend using UniEval (Fluency) as a supplement to a GPT-3.5-based NLG metric.

\noindent 3. In the context of GPT-3.5-based NLG metrics, the choice of target text source influences the performance ranking of UE-TS models. Furthermore, employing ground-truth summaries as the target text source exerts a more significant impact compared to using input text.

\noindent 4. The same generation methods cannot achieve similar correlations for fluency NLG dimension. Consequently, correlations within fluency dimensions are more closely tied to the dataset.

\noindent 5. Due to the positive or negative correlations between wo-GPT-3.5 (Fluency) and wi-GPT-3.5 (Fluency), it is difficult to determine whether GPT-3.5 itself understands the concept of fluency.

\subsubsection{Analysis on NLG Overall Dimension}
\label{sec:app_nlg_overall_dimension}

\begin{figure*}[!htbp]
\centering
\includegraphics[width=\textwidth]{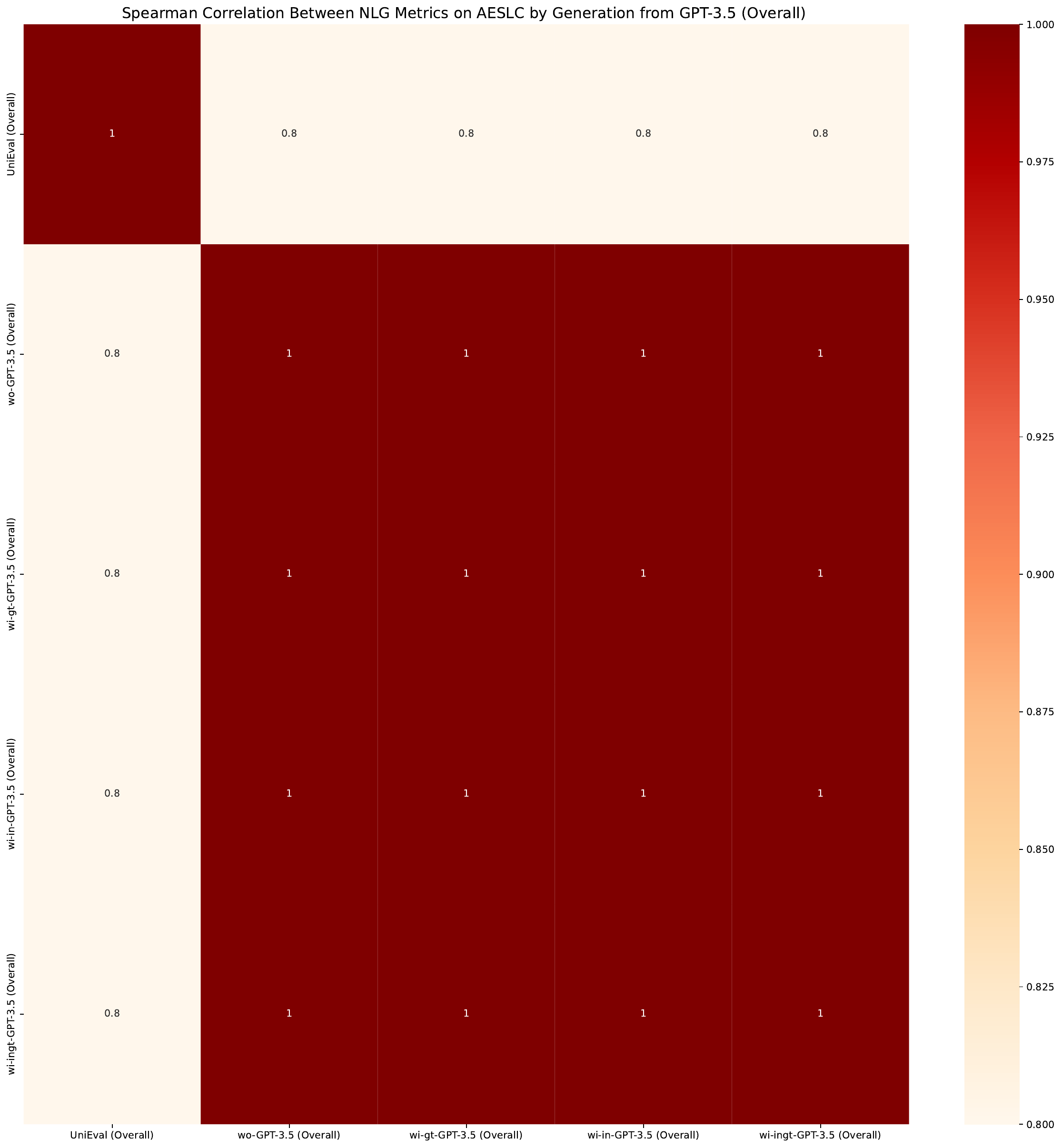}
\caption{Diagram of Spearman correlation in terms of overall between NLG metrics on AESLC dataset from the view of uncertainty estimation methods used in Fig.~\ref{fig:spear_ue_aes_gpt35}. The generated summaries are from GPT-3.5. For the GPT-3.5-based NLG metrics, we only draw wi-ingt-GPT-3.5 results to save space.}
\label{fig:spear_nlg_aes_overall_gpt35}
\end{figure*}

\begin{figure*}[!htbp]
\centering
\includegraphics[width=\textwidth]{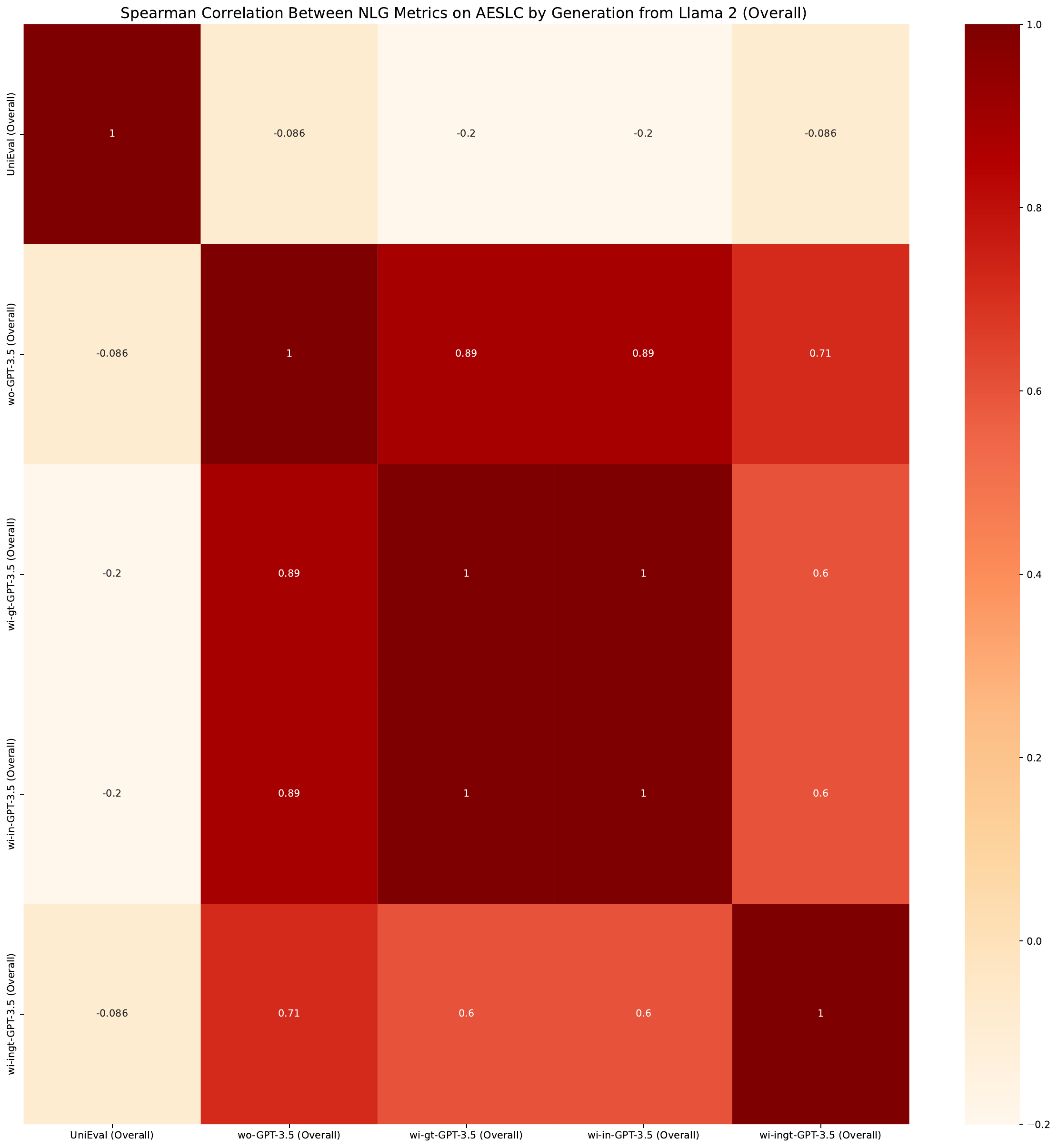}
\caption{Diagram of Spearman correlation in terms of overall between NLG metrics on AESLC dataset from the view of uncertainty estimation methods used in Fig.~\ref{fig:spear_ue_aes_llama}. The generated summaries are from Llama 2. For the GPT-3.5-based NLG metrics, we only draw wi-ingt-GPT-3.5 results to save space.}
\label{fig:spear_nlg_aes_overall_llama}
\end{figure*}

\begin{figure*}[!htbp]
\centering
\includegraphics[width=\textwidth]{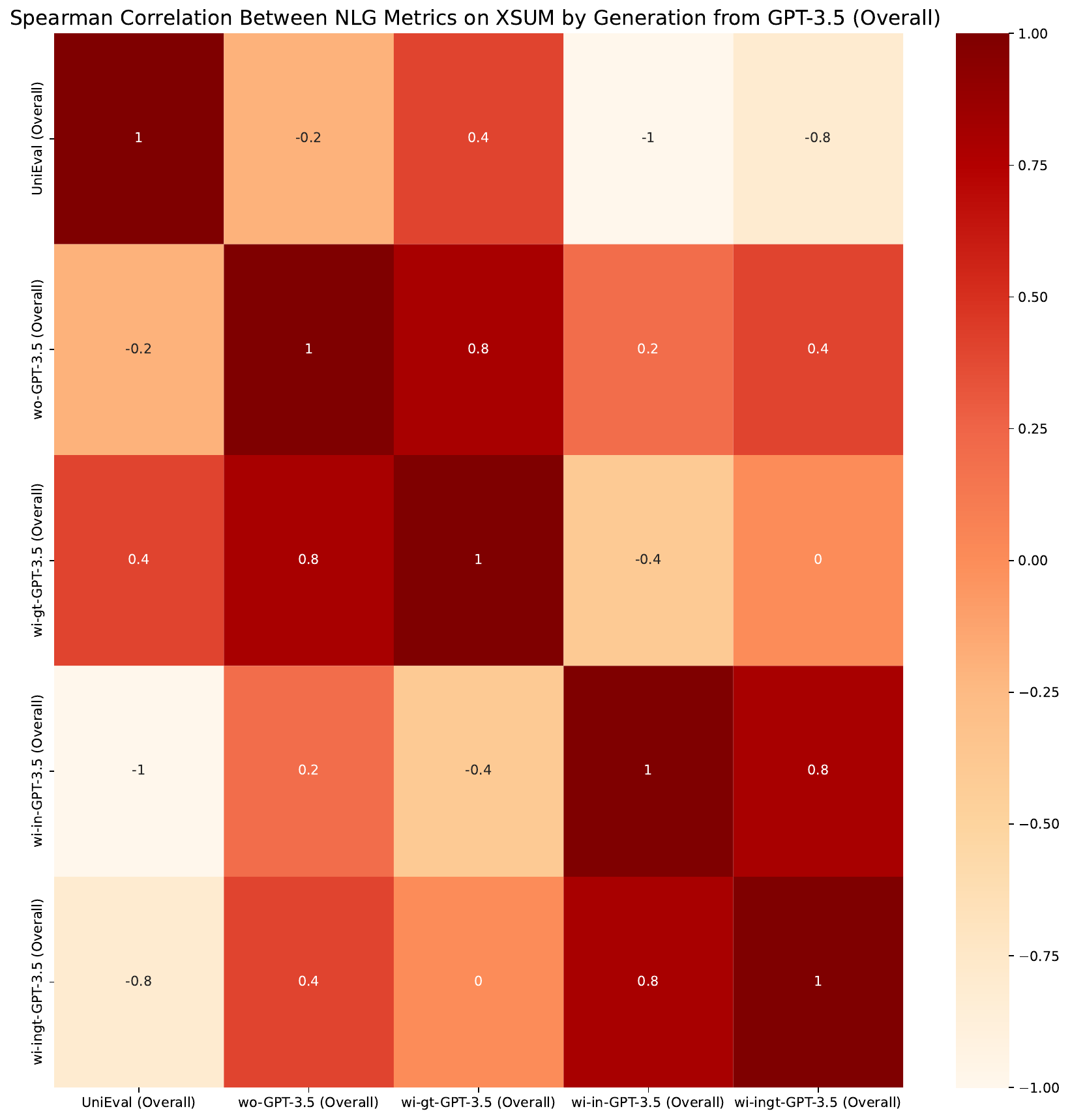}
\caption{Diagram of Spearman correlation in terms of overall between NLG metrics on XSUM dataset from the view of uncertainty estimation methods used in Fig.~\ref{fig:spear_ue_xsum_gpt35}. The generated summaries are from GPT-3.5. For the GPT-3.5-based NLG metrics, we only draw wi-ingt-GPT-3.5 results to save space.}
\label{fig:spear_nlg_xsum_overall_gpt35}
\end{figure*}

\begin{figure*}[!htbp]
\centering
\includegraphics[width=\textwidth]{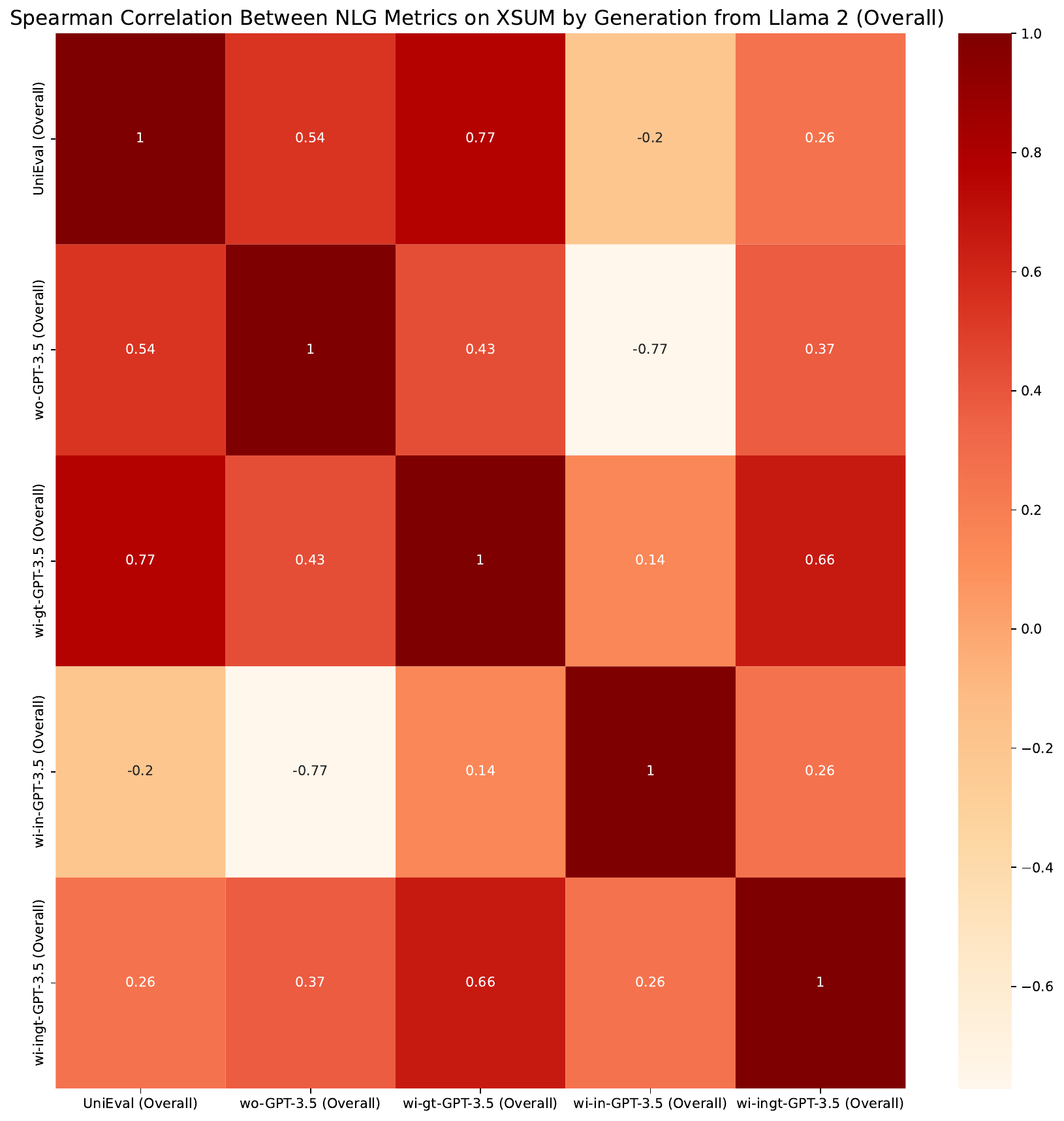}
\caption{Diagram of Spearman correlation in terms of overall between NLG metrics on XSUM dataset from the view of uncertainty estimation methods used in Fig.~\ref{fig:spear_ue_xsum_llama}. The generated summaries are from Llama 2. For the GPT-3.5-based NLG metrics, we only draw wi-ingt-GPT-3.5 results to save space.}
\label{fig:spear_nlg_xsum_overall_llama}
\end{figure*}

Figures \ref{fig:spear_nlg_aes_overall_gpt35},~\ref{fig:spear_nlg_aes_overall_llama},~\ref{fig:spear_nlg_xsum_overall_gpt35}, and~\ref{fig:spear_nlg_xsum_overall_llama} display the evaluation results of UE-TS models based on overall-dimension NLG metrics. Correlations for BART generation models in the overall dimension are not illustrated. This is because we only utilize UniEval (overall) and wo-GPT-3.5 (overall) for BART generations, and these two metrics may not adequately represent correlation.

Utilizing GPT-3.5 as the generation model, Figure~\ref{fig:spear_nlg_aes_overall_gpt35} shows that all GPT-3.5-based NLG metrics have identical performance rankings for UE-TS models. Additionally, these GPT-3.5-based NLG metrics exhibit a strongly positive correlation with UniEval (Overall).

Also, utilizing GPT-3.5 as the generation model, Figure~\ref{fig:spear_nlg_aes_overall_gpt35} illustrates that wo-GPT-3.5 (Overall) has positive correlations with the other three GPT-3.5-based NLG metrics. Furthermore, strongly positive correlations are found between wo-GPT-3.5 (Overall) and wi-gt-GPT-3.5 (Overall). Another strongly positive correlation is observed between wi-in-GPT-3.5 (Overall) and wi-ingt-GPT-3.5 (Overall).

Utilizing Llama 2 as the generation model, Figure~\ref{fig:spear_nlg_aes_overall_llama} indicates that all GPT-3.5-based NLG metrics have strongly positive correlations. However, UniEval (Overall) exhibits negative correlations with these GPT-3.5-based NLG metrics.

Also, utilizing Llama 2 as the generation model, Figure~\ref{fig:spear_nlg_xsum_overall_llama} shows that almost all five metrics have positive correlations, with the exceptions being that wi-in-GPT-3.5 (Overall) has a negative correlation with UniEval (Overall) and wo-GPT-3.5 (Overall).

Based on the above findings, we can conclude the following regarding the overall dimension.

\noindent 1. When it comes to the GPT-3.5-based NLG metrics, given their positive correlations with other metrics, it is recommended to use either wi-in-GPT-3.5 (Overall) or wi-gt-GPT-3.5 (Overall).

\noindent 2. The UniEval (Overall) exhibits high correlations with other GPT-3.5-based methods in most cases and negative correlations in a few instances. It could serve as an optional supplementary tool for GPT-3.5-based NLG metrics.

\noindent 3. Due to the positive correlations between wo-GPT-3.5 and wi-GPT-3.5 NLG metrics in most cases, it suggests that GPT-3.5 itself grasps the concept of overall.

\noindent 4. Unlike previous NLG metric dimensions, it is difficult to determine which one dominates the evaluated performance ranks between using ground-truth summaries and using input text.


\subsection{Experimental Results Involving Human Annotations}
\label{sec:human_seven_dimension}
Besides the aforementioned experiments on AESLC and XSUM, we are also interested in conducting human-related experiments. We utilize the TofuEval dataset~\cite{tang2024tofueval}, which provides human annotations across seven dimensions as will be introduced later.

These experiments consist primarily of three types:
\begin{itemize}
    \item UE-HUM experiments: These experiments involve using the human annotations as the $PR_{oracle}$ and employing uncertainty estimation metric scores as the $PR_{uncertainty}$ in Eq.~\ref{eq:prr}.

    \item NLG-HUM experiments: In these experiments, the human annotations serve as the $PR_{oracle}$ while the negative NLG metric scores are utilized as the $PR_{uncertainty}$ in Eq.~\ref{eq:prr}. The negative NLG metric score is obtained by multiplying the original NLG metric scores by -1. We use a negative NLG metric score because $PR_{uncertainty}$ expects a larger value for lower quality, and the NLG metric scores used in our work are positively correlated with quality.

    \item UE-NLG experiments: These experiments utilize NLG metric scores as the $PR_{oracle}$ and also employ uncertainty estimation metric scores as the $PR_{uncertainty}$ in Eq.~\ref{eq:prr}.
\end{itemize}

The first two types are specifically tailored to the human annotations provided in the TofuEval dataset. The third type follows the same setup as the experiments conducted on the AESLC and XSUM datasets.

\subsubsection{Seven Dimensions of Human Annotations.} The seven dimensions of human annotation used in TofuEval correspond to seven error types.

\begin{itemize}

\item Extrinsic Information (EI): The summary sentence contains new information not grounded in the source document.

\item Mis-Referencing (MR): A property or an event in the summary sentence can be found in the source material, but are associated with the wrong entity.

\item Stating Opinion As Fact (SOAF): The summary sentence entails a proposition that's mentioned in the source material not as a fact, but as someone’s opinion.

\item Reasoning Error (RE): The summary sentence makes one or more
wrong inferences from information in the source document.

\item Tense/Modality Error (TME): The tense or modal (e.g. can, may, must)
used in the summary sentence does not match the tense/modality of the source document.

\item Contradiction (CO): The summary sentence contradicts the source material.

\item Nuanced Meaning Shift (NMS): The summary sentence twists information from the source material in a subtle way.

\end{itemize}

For the human annotation score for an error type $T$, we calculate the $PR_{oracle}$ as $1 - Ratio(T)$, where $Ratio(T)$ represents the ratio of the incorrect word count for error type $T$ to the total word count in a summary.

\begin{table*}[]
\centering
\small
\begin{tabular}{l|cccc}
\hline
\textbf{NLG Metrics} & \textbf{NumSets} & \textbf{ECC} & \textbf{LexSim} & \textbf{EigV} \\
\hline
ROUGE-L                                  & -0.1869          & 0.0045       & 0.3268          & 0.0035        \\
SummaC                                   & 0.1602           & -0.1263      & -0.1231         & -0.0938       \\
CTC                                      & -0.2152          & 0.095        & 0.419           & 0.0952        \\
Spearman                                 & 0.1283           & -0.0595      & 0.1782          & -0.0114       \\
Kendall-Tau                              & 0.1287           & -0.055       & 0.1783          & -0.0116       \\
UniEval (Relevance)                      & -0.0173          & -0.2415      & -0.2378         & -0.3188       \\
UniEval (Consistency)                    & 0.2134           & -0.2464      & -0.2447         & -0.0808       \\
UniEval (Coherence)                      & -0.0798          & -0.1396      & -0.2953         & -0.0963       \\
UniEval (Fluency)                        & 0.2574           & -0.7312      & -0.6689         & -0.6715       \\
UniEval (Overall)                        & 0.0921           & -0.3211      & -0.3598         & -0.2563       \\
wi-ingt-GPT-3.5 (Relevance)              & 0.2014           & -0.0067      & 0.2394          & 0.1356        \\
wi-ingt-GPT-3.5 (Consistency)            & -0.1509          & 0.3037       & 0.4858          & 0.4395        \\
wi-ingt-GPT-3.5 (Coherence)              & 0.0309           & 0.1237       & 0.3326          & 0.2971        \\
wi-ingt-GPT-3.5 (Fluency)                & -0.1544          & 0.2456       & 0.4483          & 0.4712        \\
wi-ingt-GPT-3.5 (Overall)                & 0.1455           & 0.152        & 0.3106          & 0.3046        \\
\hline
Col Mean                                 & 0.0369           & -0.0669      & \bf0.066           & 0.0137

\\
\hline
\end{tabular}
\caption{Main results of the relationship between the uncertainty estimation methods and NLG metrics on TofuEval dataset using generation from GPT-3.5.}
\label{tab:main_tofu_gpt35_unc_nlg}
\end{table*}

\begin{table*}[]
\centering
\small
\begin{tabular}{l|ccccccc|c}
\hline
\textbf{Uncertainty Estimation Methods} & \textbf{EI} & \textbf{MR} & \textbf{SOAF} & \textbf{RE} & \textbf{TME} & \textbf{CO} & \textbf{NMS} & \textbf{Mean} \\
\hline
NumSets              & 0.1929      & 0.0414      & -0.1774       & -0.3319     & -0.2410      & 0.0287      & -0.0487      & -0.0766           \\
ECC                  & 0.1619      & -0.4640     & -0.7079       & 0.2717      & 0.0072       & -0.1679     & -0.2993      & -0.1712           \\
LexSim               & 0.3119      & 0.1317      & 0.5493        & 0.3579      & 0.2672       & -0.1781     & 0.0264       & \bf0.2095            \\
EigV                 & 0.2887      & -0.2198     & -1.1169       & 0.1539      & 0.1343       & -0.1724     & -0.3462      & -0.1826      
\\
\hline
\end{tabular}
\caption{Main results of the relationship between the uncertainty estimation methods and human annotations on TofuEval dataset using generation from GPT-3.5.}
\label{tab:main_tofu_gpt35_unc_hum}
\end{table*}

\subsubsection{UE-HUM Experimental Results}
\label{sec:app_ue_hum_dim}

\begin{figure*}[!htbp]
\centering
\includegraphics[width=\textwidth]{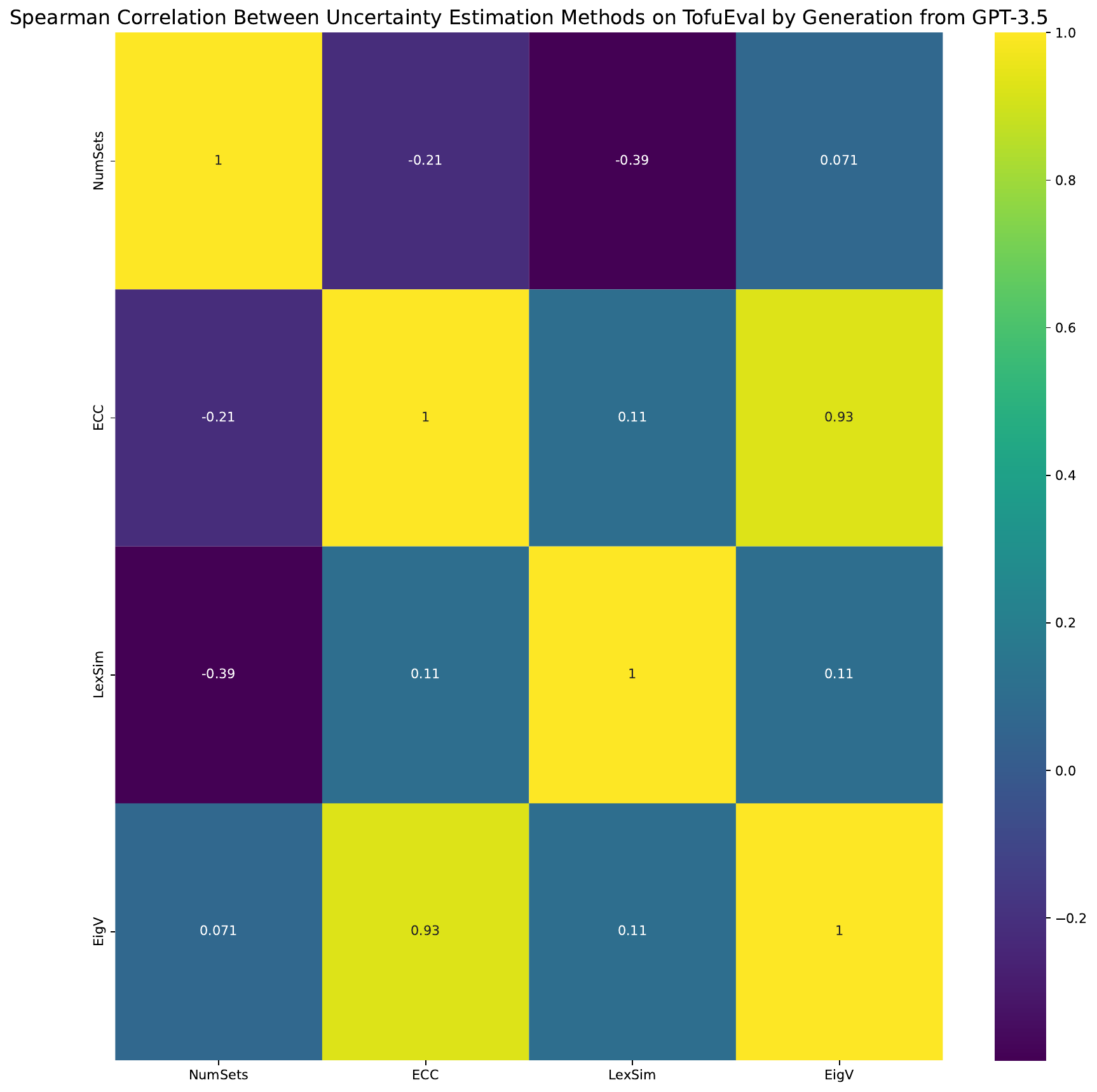}
\caption{Diagram of Spearman correlation between uncertainty estimation on TofuEval dataset from the view of human annotation.  The generated summaries are from GPT-3.5. }
\label{fig:spear_ue_hum_tofu_gpt35}
\end{figure*}

Table~\ref{tab:main_tofu_gpt35_unc_hum} and Figure~\ref{fig:spear_ue_hum_tofu_gpt35} show the UE-HUM experimental results. From Table~\ref{tab:main_tofu_gpt35_unc_hum}, we can draw the following conclusions.

\noindent 1. All uncertainty estimation methods used in our work have positive PRRs for EI.

\noindent 2. There is no uncertainty estimation method that has positive PRRs in all dimensions.

\noindent 3. The LexSim dimension achieves the highest performance among the seven human annotation dimensions, exhibiting the largest mean value across rows.

From Figure~\ref{fig:spear_ue_hum_tofu_gpt35}, we observe that ECC and EigV perform similarly from the perspective of human annotation.

\subsubsection{NLG-HUM Experimental Results}
\label{sec:app_nlg_hum_dim}

\begin{figure*}[!htbp]
\centering
\includegraphics[width=\textwidth]{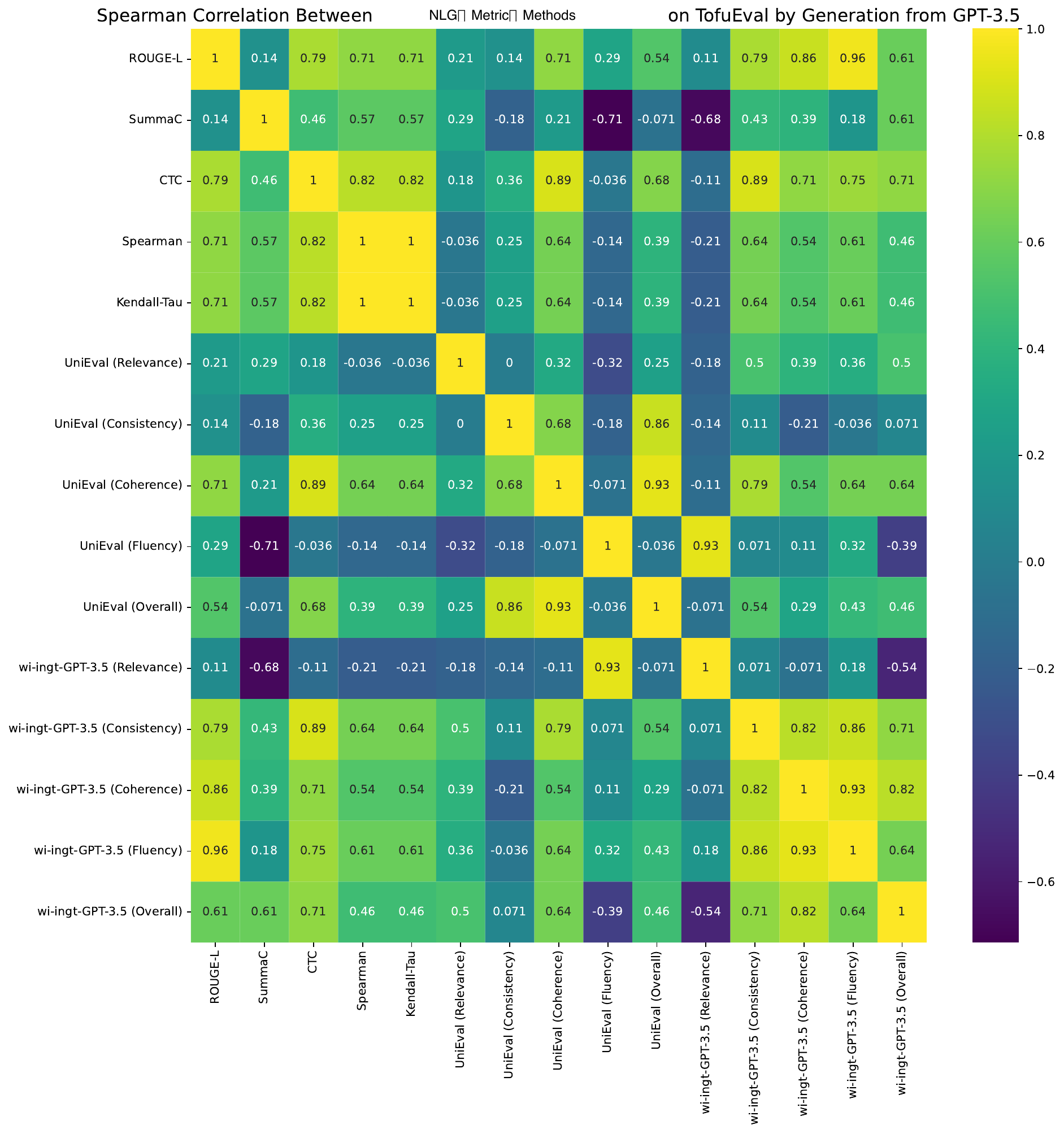}
\caption{Diagram of Spearman correlation between NLG metrics on TofuEval dataset from the view of human annotation.  The generated summaries are from GPT-3.5. }
\label{fig:spear_nlg_hum_tofu_gpt35}
\end{figure*}

Table~\ref{tab:main_tofu_gpt35_nlg_hum} and Figure~\ref{fig:spear_nlg_hum_tofu_gpt35} show the NLG-HUM experimental results. 

Based on Table~\ref{tab:main_tofu_gpt35_nlg_hum}, the following conclusions can be drawn.

\noindent 1. It is challenging for any NLG metric to achieve a positive PRR for all human annotation dimensions because no row consistently shows positivity across all dimensions.

\noindent 2. UniEval (Coherence) shows a large positive PRR (0.6794) with SOAF. However, it also shows a negative PRR (-0.4737) with RE. 

\noindent 3. All NLG metrics show negative PRR values in terms of row mean. This means that the NLG metrics do not show good consistency with the human annotation. 

From Table~\ref{tab:main_tofu_gpt35_unc_hum}, we can obtain below conclusions.

\noindent 1. The ROUGE-L and CTC exhibit similar performance based on human annotation, while SummaC demonstrates dissimilar performance compared to these two.

\noindent 2. The Spearman and Kendall-Tau perform the same from the perspective of human annotation.

\noindent 3. The five dimensions (relevance, consistency, coherence, fluency, overall) of UniEval perform weak correlations among these five dimensions from the perspective of human annotation. However, the four dimensions (consistency, coherence, fluency, overall) of wi-ingt-GPT-3.5 show positive correlations from the perspective of human annotation.

\subsubsection{UE-NLG Experimental Results}
\label{sec:app_ue_nlg_dim}

\begin{figure*}[!htbp]
\centering
\includegraphics[width=\textwidth]{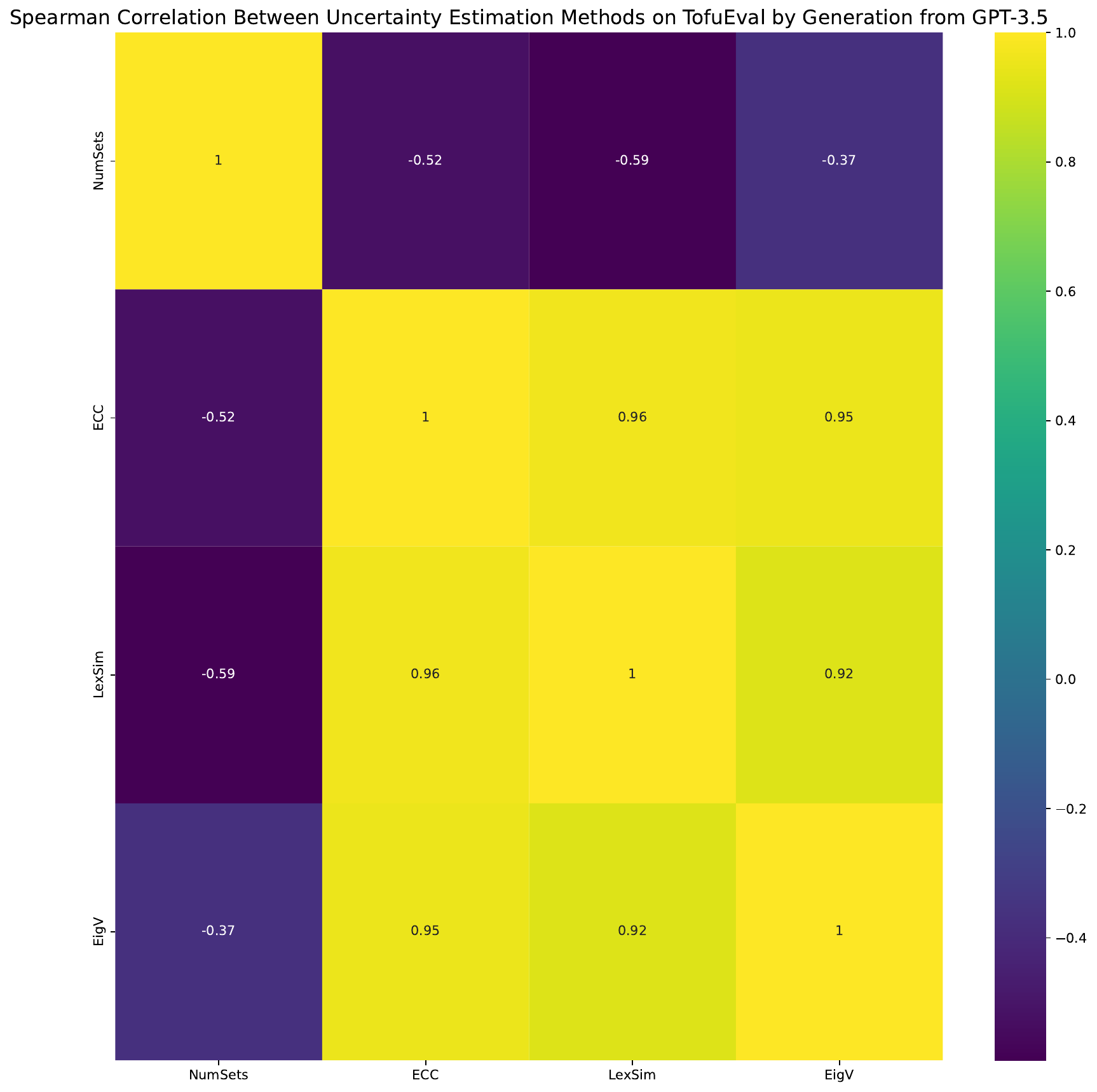}
\caption{Diagram of Spearman correlation between uncertainty estimation on TofuEval dataset from the view of NLG metrics.  The generated summaries are from GPT-3.5. }
\label{fig:spear_ue_nlg_tofu_gpt35}
\end{figure*}

Table~\ref{tab:main_tofu_gpt35_unc_nlg} and Figure~\ref{fig:spear_ue_nlg_tofu_gpt35} show the UE-NLG experimental results. 

From Table~\ref{tab:main_tofu_gpt35_nlg_hum}, we can conclude below.

\noindent 1. It is hard for any uncertainty estimation method to achieve positive PRR for all the NLG metric methods.

\noindent 2. LexSim performs the best among the NLG metrics, consistent with the UE-HUM experimental results. Thus, NLG metrics remain useful for identifying the optimal uncertainty estimation method, which may align with human annotation.

Based on Figures~\ref{tab:main_tofu_gpt35_unc_hum} and~\ref{fig:spear_ue_nlg_tofu_gpt35}, we found that rankings of certain uncertainty estimation methods differ between NLG metrics and human annotation. For instance, NumSets shows a negative correlation with EigV according to NLG metrics, while a positive correlation is observed based on human annotation. However, utilizing NLG metrics as an oracle remains meaningful, as it sometimes yields similar results to using human annotation as the oracle.


\end{document}